%% file: example_paper.tex
\documentclass{article}

\usepackage{microtype}
\usepackage{graphicx}
\usepackage{booktabs} 

\usepackage{hyperref}

\UseRawInputEncoding

\usepackage[accepted]{icml2025}

\usepackage{amsmath}
\usepackage{amssymb}
\usepackage{mathtools}
\usepackage{amsthm}

\usepackage{graphicx}
\usepackage{subcaption}
\usepackage{amsmath}
\usepackage{xspace}

\usepackage{courier}
\usepackage{tcolorbox}

\usepackage[capitalize,noabbrev]{cleveref}
\usepackage{multirow}
\usepackage{makecell}
\usepackage{multicol}
\usepackage{tabularx}
\usepackage{pdfpages}
\usepackage{longtable} 
\usepackage{listings}
\usepackage{xcolor}
\usepackage{colortbl}
\definecolor{LightCyan}{rgb}{0.88,1,1}
\definecolor{LightPink}{rgb}{1, 0.88, 0.88}
\definecolor{LightOrange}{rgb}{1, 0.8, 0.6}
\definecolor{VeryLightPink}{rgb}{1, 0.94, 0.94}

\theoremstyle{plain}

\theoremstyle{definition}

\theoremstyle{remark}

\usepackage[textsize=tiny]{todonotes}

\icmltitlerunning{MAP}

\begin{document}

\twocolumn[
\icmltitle{Aligning Instruction Tuning with Pre-training}



\icmlsetsymbol{equal}{*}


\begin{icmlauthorlist}
\icmlauthor{Yiming Liang}{equal,cas,ia,baai}\hspace{5mm}
\icmlauthor{Tianyu Zheng}{equal,map,01}\hspace{5mm}
\icmlauthor{Xinrun Du}{equal,map,01}\hspace{5mm}
\icmlauthor{Ge Zhang}{equal,map}\\
\vspace{0.5mm}
\icmlauthor{Jiaheng Liu}{map}\hspace{2mm}
\icmlauthor{Xingwei Qu}{map}\hspace{2mm}
\icmlauthor{Wenqiang Zu}{cas,ia}\hspace{2mm}
\icmlauthor{Xingrun Xing}{cas,ia}\hspace{2mm}
\icmlauthor{Chujie Zheng}{01} \\
\vspace{0.5mm}
\icmlauthor{Lei Ma}{baai,pku}
\icmlauthor{Guoyin Wang}{01}
\icmlauthor{Zhaoxiang Zhang}{ia}
\icmlauthor{Wenhao Huang}{map}
\icmlauthor{Xiang Yue}{map}
\icmlauthor{Jiajun Zhang}{cas,ia}
\end{icmlauthorlist}

\icmlaffiliation{cas}{School of Artificial Intelligence, University of Chinese Academy of Sciences}
\icmlaffiliation{ia}{Institute of Automation, Chinese Academy of Sciences}
\icmlaffiliation{map}{M-A-P}
\icmlaffiliation{01}{01.ai}
\icmlaffiliation{baai}{BAAI}
\icmlaffiliation{pku}{Peking University}

\icmlcorrespondingauthor{Jiaheng Liu}{buaaljiaheng@gmail.com}
\icmlcorrespondingauthor{wenhaohuang}{rubio8741@gmail.com}
\icmlcorrespondingauthor{JiaJun Zhang}{jjzhang@nlpr.ia.ac.cn}

\icmlkeywords{Machine Learning, ICML}

\vskip 0.3in
]



\printAffiliationsAndNotice{\icmlEqualContribution} 

\newcommand{\ABMN}{AITP\xspace}

\begin{abstract}


Instruction tuning enhances large language models (LLMs) to follow human instructions across diverse tasks, relying on high-quality datasets to guide behavior. However, these datasets, whether manually curated or synthetically generated, are often narrowly focused and misaligned with the broad distributions captured during pre-training, limiting LLM generalization and effective use of pre-trained knowledge. We propose \textit{Aligning Instruction Tuning with Pre-training} (\ABMN), a method that bridges this gap by identifying coverage shortfalls in instruction-tuning datasets and rewriting underrepresented pre-training data into high-quality instruction-response pairs. This approach enriches dataset diversity while preserving task-specific objectives. Evaluations on three fully open LLMs across eight benchmarks demonstrate consistent performance improvements with \ABMN. Ablations highlight the benefits of adaptive data selection, controlled rewriting, and balanced integration, emphasizing the importance of aligning instruction tuning with pre-training distributions to unlock the full potential of LLMs.
\end{abstract}
\vspace{-0.5cm}

\input{intro_v2}


\section{Methods}\label{methods}
\subsection{Difference Set Generation}
In this section, we define the process of difference set generation, isolating data points from the pre-training corpora that differ from those in the SFT dataset. The goal is to identify regions in the pre-training data distribution that are absent from or sparsely populated in the supervised fine-tuning (SFT) data.
This can be formalized as follows:
\begin{equation}
D_{\text{diff}} = \{d_i | d_i \in D_{\text{pretrain}}, \Delta(d_i, D_{\text{SFT}}) < \tau \}
\end{equation}
Where $D_{\text{pretrain}}$, $D_{\text{SFT}}$, $D_{\text{diff}}$ represent the pre-training dataset, the SFT dataset and the resulting difference set, respectively. 
$\Delta(d_i, D_{\text{SFT}})$ represents the density estimate of the data point $d_i$ in the SFT dataset, and $\tau$ is the threshold that determines whether a data point should be included in the difference set.
To achieve this, we outline the procedure in three main stages: data representation, density estimation, and identification of the difference set.

\subsubsection{Data Representation} 
Each data point is represented as a vector derived from the final-layer embedding of the model.
We then apply dimensionality reduction (DR) to project these high-dimensional embeddings into two-dimensional coordinates, facilitating visualization and density comparison across datasets.
This process can be formalized as follows:
\begin{equation}
(x_i, y_i) = \text{DR}(\text{Model}(d_i))
\end{equation}
Applying the same dimension reduction to both pre-training and SFT embeddings results in two sets of two-dimensional vectors:
\begin{align}
&Z_{\text{pretrain}} = \{(x_i, y_i) \mid d_i \in D_{\text{pretrain}}\} \\
&Z_{\text{SFT}} = \{(x_i, y_i) \mid d_i \in D_{\text{SFT}}\}
\end{align}

\subsubsection{Density Estimation}
To compare data distributions between the pre-training and SFT datasets, we use Kernel Density Estimation (KDE) to visualize the density of points for each dataset. The KDE function \( \hat{f}(x, y) \) estimates the density at any location \( (x, y) \) based on neighboring points:
\begin{equation}
\hat{f}(x, y) = \frac{1}{n h_x h_y} \sum_{i=1}^{n} K\left( \frac{x - x_i}{h_x}, \frac{y - y_i}{h_y} \right)
\end{equation}
\( K(\cdot, \cdot) \) is the kernel function, typically Gaussian:
\begin{equation}
\resizebox{0.9\hsize}{!}{$
K((x, y), (x{\prime}, y{\prime})) = \exp\left(-\frac{(x - x{\prime})^2 + (y - y{\prime})^2}{2\sigma^2}\right)
$}
\end{equation}
Where \((x, y)\) and \((x{\prime}, y{\prime}) \) are two two-dimensional data points, \( h_x \), \( h_y \) and \(\sigma\) are bandwidth parameters that control the smoothness in the x direction, y direction and kernel respectively. The KDE visualization highlights distribution differences, identifying regions of divergence between the pretraining and SFT datasets.

\subsubsection{Finding Difference Set}
The difference set is identified based on the density estimates from the SFT dataset. 
Specifically, if a point  $d_i$  in the pre-training dataset has a low-density estimate within the SFT dataset, we classify this point as absent or sparsely populated in the SFT data. Such points contribute to the observed distributional differences between the two datasets, and we define them formally as:
\begin{equation}
\label{eq:density threshold}
D_{\text{diff}} = \{d_i | d_i \in D_{\text{pretrain}}, \hat{f}_{\text{SFT}}(x_i,y_i)<\tau \}
\end{equation}
\(\hat{f}_{\text{SFT}}(x_i,y_i)\) represents the density estimate of the data point \(d_i\) from the pretrain corpus within the SFT dataset.
\begin{equation}
\hat{f}_{\text{SFT}}(x_i,y_i) = \frac{1}{n h_x h_y} \sum_{j=1}^{n} K\left( \frac{x_i - x_j}{h_x}, \frac{y_i - y_j}{h_y} \right)
\label{eq:density estimation}
\end{equation}
Where \((x_i, y_i) \in Z_{\text{pretrain}}\), \((x_j, y_j) \in Z_{\text{SFT}}\). \(n\) is the total number of points in the SFT dataset.

\subsection{Data Transformation of Difference Set}
The data transformation phase is designed to convert raw text from the pre-training data within the difference set into instruction-pair data formatted for SFT. This process consists of three key steps.
First, we develop a \textbf{query generation prompt} to guide the model in generating relevant questions from the raw text. Next, we implement a \textbf{query scoring prompt} to assess the quality of each generated query. Low-quality queries are filtered out based on these scores, enabling us to eliminate unsuitable questions before answer generation, thus conserving computational resources. Finally, an \textbf{answer generation prompt} is applied to instruct the model in generating responses to the remaining high-quality queries.
These three processes can be formally modeled as follows:
\begin{equation}
\hat{y}_t^i = \arg\max_{y_t} P(y_t \mid p_{\text{generate}}, t, y_{<t}; \theta)
\end{equation}
\begin{equation}
\hat{y}_t^s = \arg\max_{y_t} P(y_t \mid p_{\text{score}}, i, y_{<t}; \theta)
\end{equation}
\begin{equation}
\hat{y}_t^a = \arg\max_{y_t} P(y_t \mid p_{\text{answer}}, i, y_{<t}; \theta)
\end{equation}
where \( p_{\text{generate}} \), \( p_{\text{score}} \), and \( p_{\text{answer}} \) represent the prompts used for query generation, query scoring, and answer generation, respectively. Here, \( t \) denotes the raw text, \( i \) represents the instruction, and \( \theta \) denotes the model parameters. The \( \hat{y}_t^i \), \( \hat{y}_t^s \), and \( \hat{y}_t^a \) represent the most probable tokens generated at time step \( t \) for the instruction, score, and answer, respectively.
The detailed prompts utilized in this process can be found in \autoref{appendix:prompts}.

\subsection{Training}

In this phase, the model is trained on a combined dataset that includes both the rewritten data derived from the difference set and the original SFT dataset. Notably, the model trained on the combined dataset is the same as the one trained on the pre-training corpus. This serves two main purposes: first, it ensures consistency between the supplemented knowledge distribution and the model's internal knowledge. Second, high-quality instruction-pair data helps correct semantic inaccuracies that may arise from formatting errors in the pre-training corpus. 
The loss function for training is defined as follows:
\begin{equation}
    \mathcal{L}_{\text{avg}} = -\frac{1}{N} \sum_{t=1}^{N} \log P(a_t \mid i, a_{<t}; \theta)
\end{equation}
where \( N \) denotes the sequence length, \( i \) and \( a \) denote the instruction and response sequence, respectively.

\section{Experiment Settings}\label{experiments}
\subsection{Evaluation}
We evaluate the model's instruction-following ability using the IFEval benchmark \cite{zhou2023instructionfollowing}, which is unbiased because it does not rely on LLM-generated evaluation scores. It provides four types of accuracy scores: Prompt-level Strict-accuracy (P-S), Instruction-level Strict-accuracy (I-S), Prompt-level Loose-accuracy (P-L), and Instruction-level Loose-accuracy (I-L). We use the OpenCompass, a comprehensive, one-stop platform for LLM evaluation \cite{2023opencompass}. We evaluate the effectiveness of \ABMN across seven standard benchmarks. These benchmarks provide a comprehensive evaluation of the diverse capabilities of language models across various tasks and domains. MMLU \cite{hendryckstest2021} offers a broad assessment of multitask reasoning and knowledge retrieval, while ARC-c \cite{clark2018think} and GPQA-diamond \cite{rein2023gpqa} focus on complex scientific reasoning and physics-specific understanding, respectively. For code generation and problem-solving, HumanEval \cite{chen2021evaluating} and MBPP \cite{austin2021program} measure a model’s ability to write correct and multi-step logical solutions. Additionally, HellaSwag \cite{zellers2019hellaswag} tests commonsense reasoning by predicting contextually appropriate continuations, and GSM8K \cite{cobbe2021training} challenges models with elementary-level math problems, combining natural language understanding with mathematical reasoning.

\subsection{Main Setting}
Our experiments utilize three fully open-source models: OLMo \cite{groeneveld2024olmo}, MAP-Neo \cite{zhang2024mapneo} and Pythia \cite{biderman2023pythia}. These models not only release model weights but also training datasets and intermediate checkpoints, aiming to facilitate reproduction and advance scientific research in LLMs. In this paper, the  \href{https://huggingface.co/allenai/OLMo-7B}{OLMo-7B-base},  \href{https://huggingface.co/m-a-p/neo\_7b}{MAP-Neo-7B-base}, and \href{https://huggingface.co/EleutherAI/pythia-12b}{Pythia-12B} models, along with their corresponding pre-training corpora, are chosen as the foundational setup for \ABMN. The \href{https://huggingface.co/allenai/OLMo-7B-SFT}{OLMo-7B-SFT} and \href{https://huggingface.co/m-a-p/neo\_7b\_sft\_v0.1}{MAP-Neo-7B-SFT-v0.1} models are used as baselines to validate the effectiveness of \ABMN. Since the SFT dataset for Pythia has not been released, we use Tulu-v2 for fine-tuning as the baseline for Pythia.

Due to the substantial storage and computational resources required for the data embedding and shift phase, we don't use the full pre-training corpus given resource constraints. Instead, we apply reservoir sampling \cite{10.1145/3147.3165}, an algorithm that enables uniform sampling from streaming data, ensuring that the sampled subset maintains a distribution consistent with the full pre-training corpus. The reservoir sampling algorithm is described in the \autoref{appendix:algorithm}.

We conduct experiments on the NVIDIA A800-SXM4-80GB, with the difference set generation phase taking approximately 56 GPU hours. The Data Transformation Setting phase utilizes the vLLM~\cite{kwon2023efficient} framework to accelerate inference, requiring approximately 640 GPU hours, while the Training Setting phase, involving full-parameter fine-tuning, takes approximately 256 GPU hours.

\subsection{Difference Set Generation Setting}

We obtain the text embeddings using two encoding models: \href{https://huggingface.co/BAAI/bge-m3}{BAAI/bge-m3}~\cite{bge-m3} and \href{https://huggingface.co/sentence-transformers/all-MiniLM-L6-v2}{sentence-transformers/all-MiniLM-L6-v2}~\cite{reimers-2019-sentence-bert}. 
We choose the all-MiniLM-L6-v2 model for its simplicity and ease of use, while bge-m3 can handle multilingual input and varying input lengths, from short sentences to long documents up to 8192 tokens. 
For the pre-training corpus, we directly use the text field as input for encoding. For the SFT dataset, we concatenate the instruction and response fields to form a complete input text for encoding. After obtaining the text embeddings, we apply principal component analysis (PCA) to reduce the high-dimensional data to two dimensions, thus simplifying the visualization and analysis. 
For visualization, we employ kernel density estimation (KDE), which effectively represents data density by smoothing distributions and avoids the issue of point overlap in dense regions that can occur in scatter plots.

To identify the difference set, we use two settings: density estimation and density comparison. 
The density estimation setting is presented in \autoref{eq:density threshold} and \autoref{eq:density estimation}.  In this paper, the density comparison setting compares the density estimation of each data point in the pre-training and SFT datasets, selecting difference points based on their density ratio. The density comparison setting is formalized as follows:
\begin{equation}
\resizebox{0.85\hsize}{!}{$\hat{f}_{\text{Pre}}(x_i,y_i) = \frac{1}{m h_x h_y} \displaystyle \sum_{\substack{k=1, k\neq i}}^{m} K\left( \frac{x_i - x_k}{h_x}, \frac{y_i - y_k}{h_y} \right)$}
\end{equation}

\begin{equation}
D_{\text{diff}} = \{d_i | d_i \in D_{\text{pretrain}}, \frac{\hat{f}_{\text{Pre}}(x_i,y_i)}{\hat{f}_{\text{SFT}}(x_i,y_i)}>\tau \}
\label{eq:density comparison}
\end{equation}
Where \((x_i, y_i), (x_k, y_k) \in Z_{\text{pretrain}}\). \(m\) is the total number of points in the pre-training dataset. In this paper, we set $\tau$ to 0.7 and 1.0 for equations (\ref{eq:density threshold}) and (\ref{eq:density comparison}), respectively.


\subsection{Data Transformation Setting}
We employ the \href{https://huggingface.co/Qwen/Qwen2.5-72B-Instruct}{Qwen2.5-72B-Instruct} \cite{qwen2.5} model for data transformation. In the instruction generation phase, we ensure that generated instructions are contextually relevant and self-contained, meaning they should not require the raw text as background for understanding. During the instruction scoring phase, each instruction is assessed based on three criteria: quality, difficulty, and the additional information required. We rate the quality of each instruction on a scale from 1 to 10 based on its clarity, assess its difficulty depending on whether specialized knowledge is required, and mark the additional information required field \textit{true} or \textit{false}, based on whether extra information is needed to fully answer the query. In the answer generation phase, the model is prompted to produce comprehensive and accurate responses informed by both the instruction and text content, ensuring that the responses are detailed and well-aligned with the question context. 

\subsection{Ablation Setting}
We conduct two ablation studies to evaluate the impact of dataset size and distillation during the data transformation process on \ABMN. To determine whether the improvement arises from the increased size of the SFT dataset after adding the rewritten difference set, we sample a subset from the combined dataset (original SFT and rewritten difference set) that is equal in size to the original SFT dataset and use it for training. To test whether the improvement is due to distillation in the data transformation phase, we replace the original SFT dataset with a subset sampled from the pre-training corpus that shares a similar distribution and train the model on the combined dataset  (the rewritten same set and the rewritten difference set). This setup aligns with the approach used in LongForm \cite{kksal2023longform}, which trains models on fully rewritten pre-training datasets but overlooks leveraging existing high-quality datasets.

\begin{figure*}[!ht]
\begin{center}
\begin{subfigure}{0.245\linewidth}
    \centering
    \includegraphics[width=\linewidth]{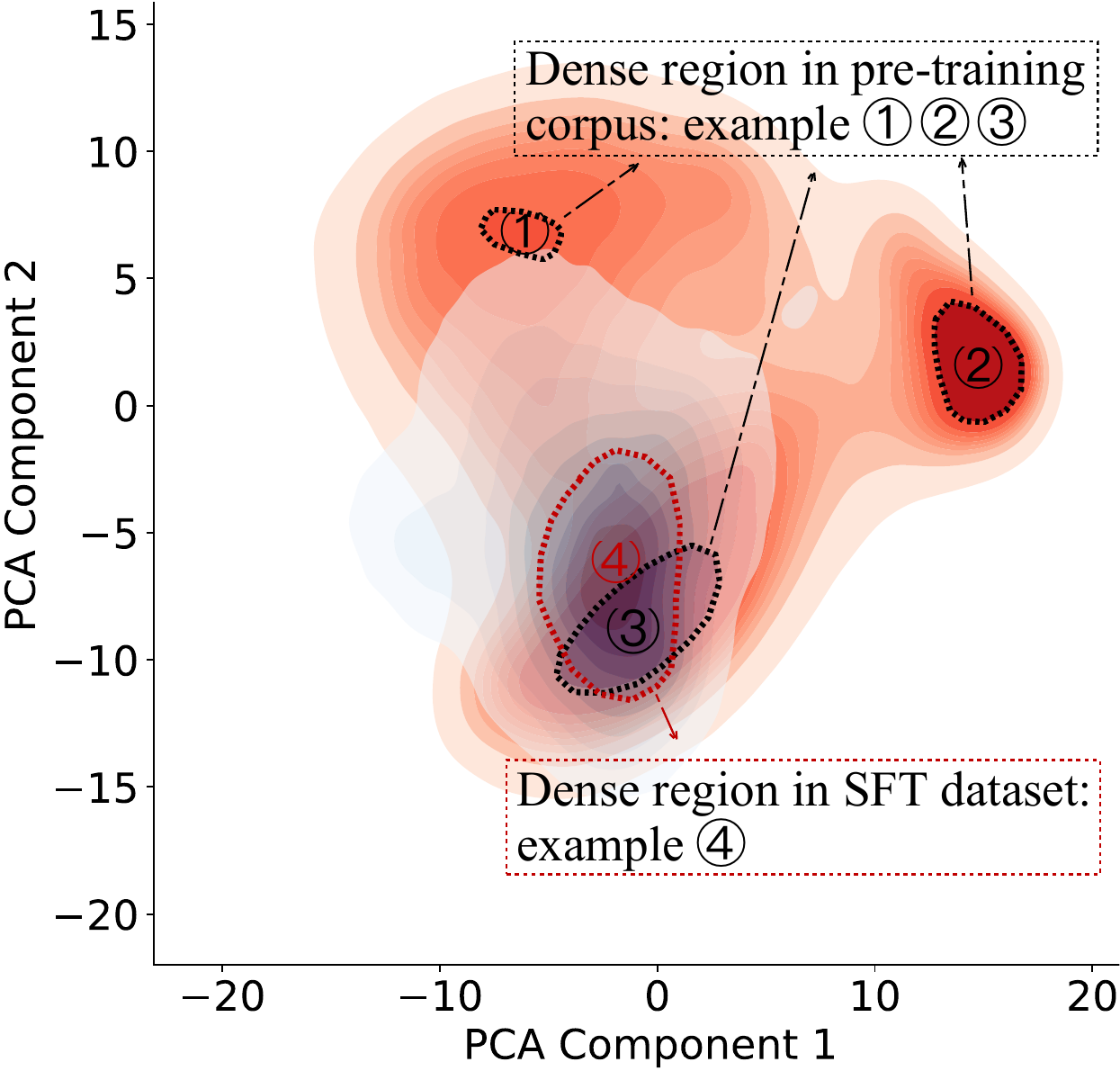}
    \subcaption{Original dataset TuluV2}
    \label{fig:sub1}
\end{subfigure}%
\begin{subfigure}{0.245\linewidth}
    \centering
    \includegraphics[width=\linewidth]{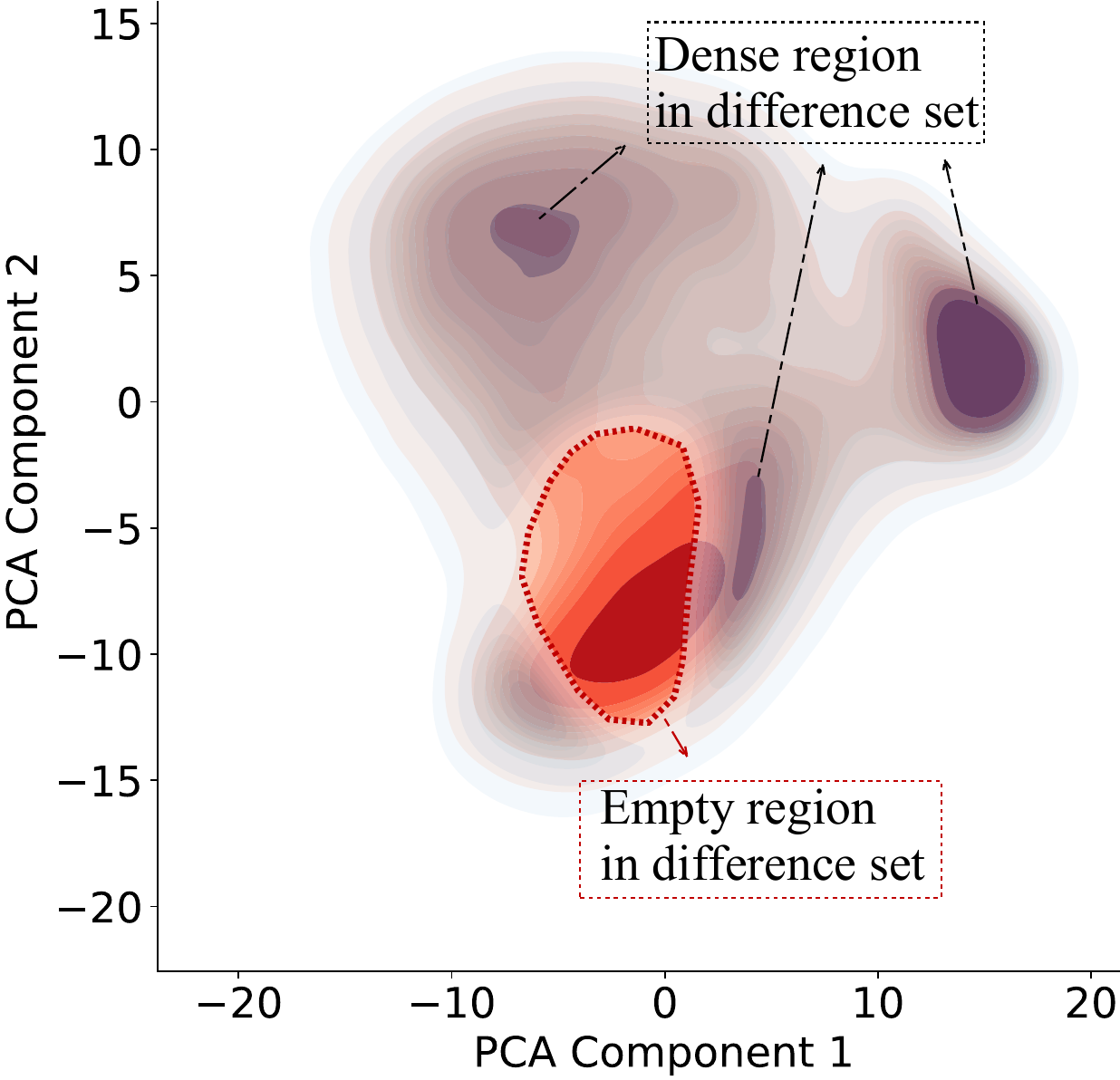}
    \subcaption{The difference set}
    \label{fig:sub2}
\end{subfigure}%
\begin{subfigure}{0.245\linewidth}
    \centering
    \includegraphics[width=\linewidth]{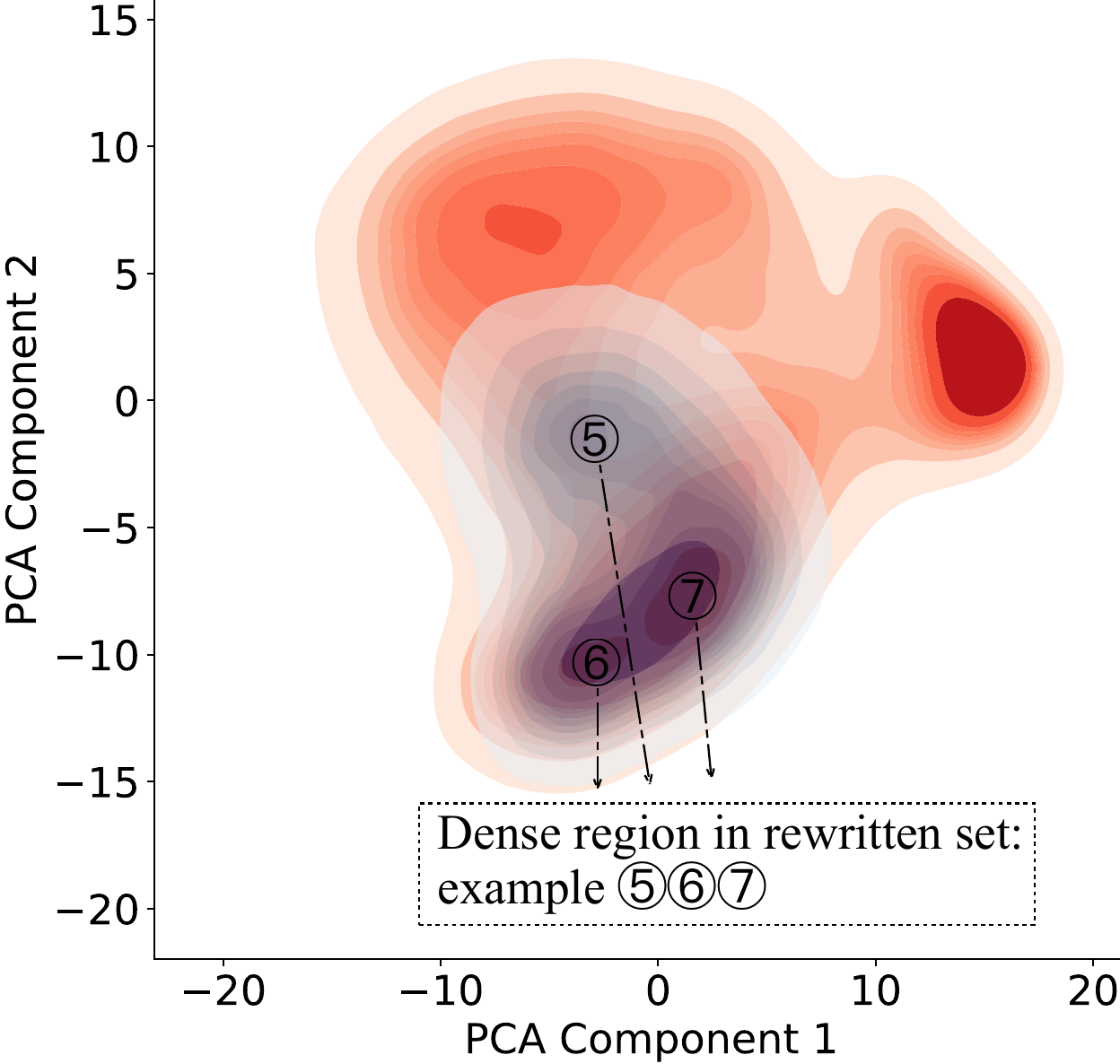}
    \subcaption{The rewritten set}
    \label{fig:sub3}
\end{subfigure}%
\begin{subfigure}{0.245\linewidth}
    \centering
    \includegraphics[width=\linewidth]{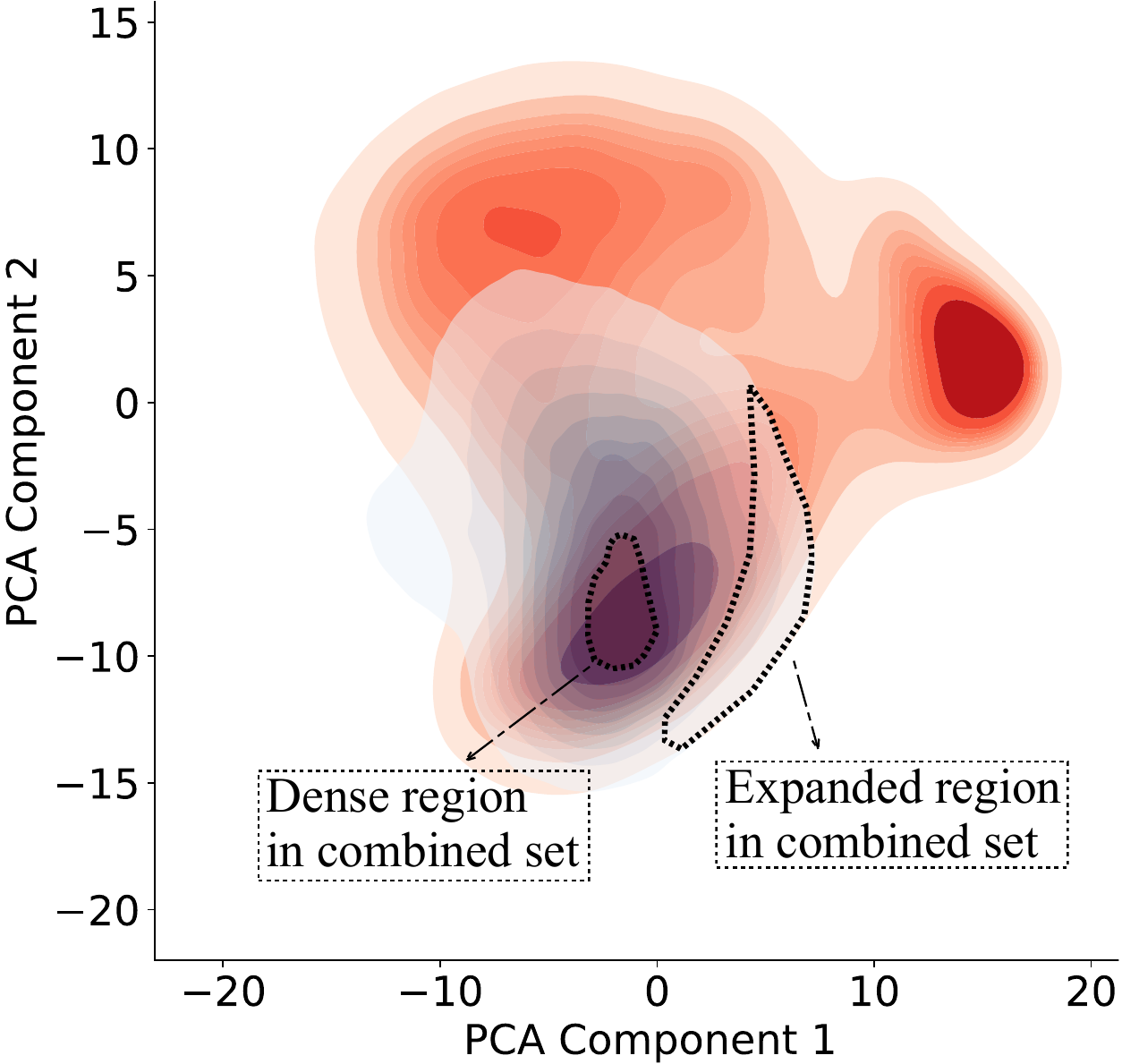}
    \subcaption{The combined set}
    \label{fig:sub4}
\end{subfigure}%

\vspace{1em} 

\begin{subfigure}{0.245\linewidth}
    \centering
    \includegraphics[width=\linewidth]{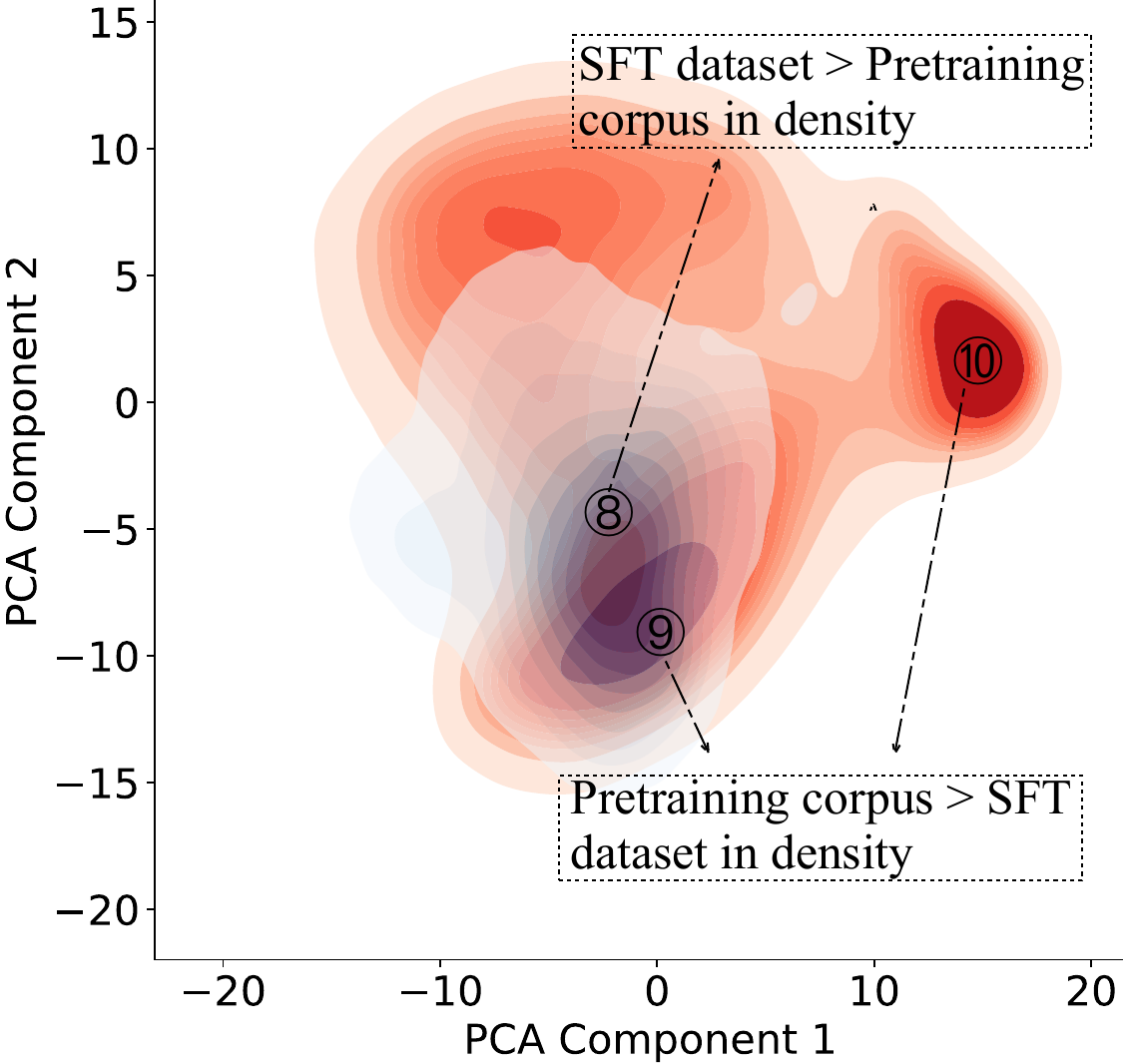}
    \subcaption{Original dataset TuluV2}
    \label{fig:sub5}
\end{subfigure}%
\begin{subfigure}{0.245\linewidth}
    \centering
    \includegraphics[width=\linewidth]{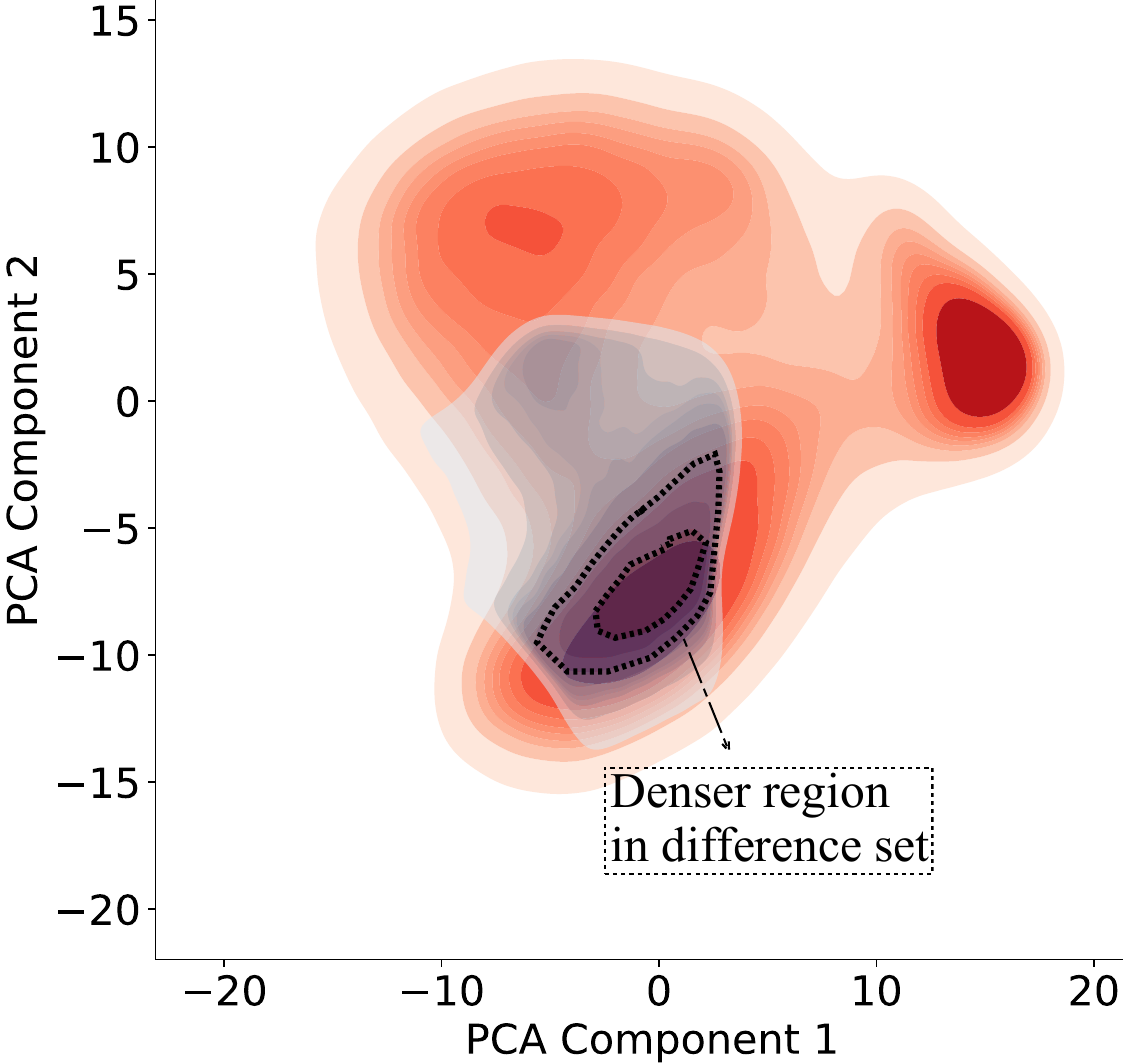}
    \subcaption{The difference set}
    \label{fig:sub6}
\end{subfigure}%
\begin{subfigure}{0.245\linewidth}
    \centering
    \includegraphics[width=\linewidth]{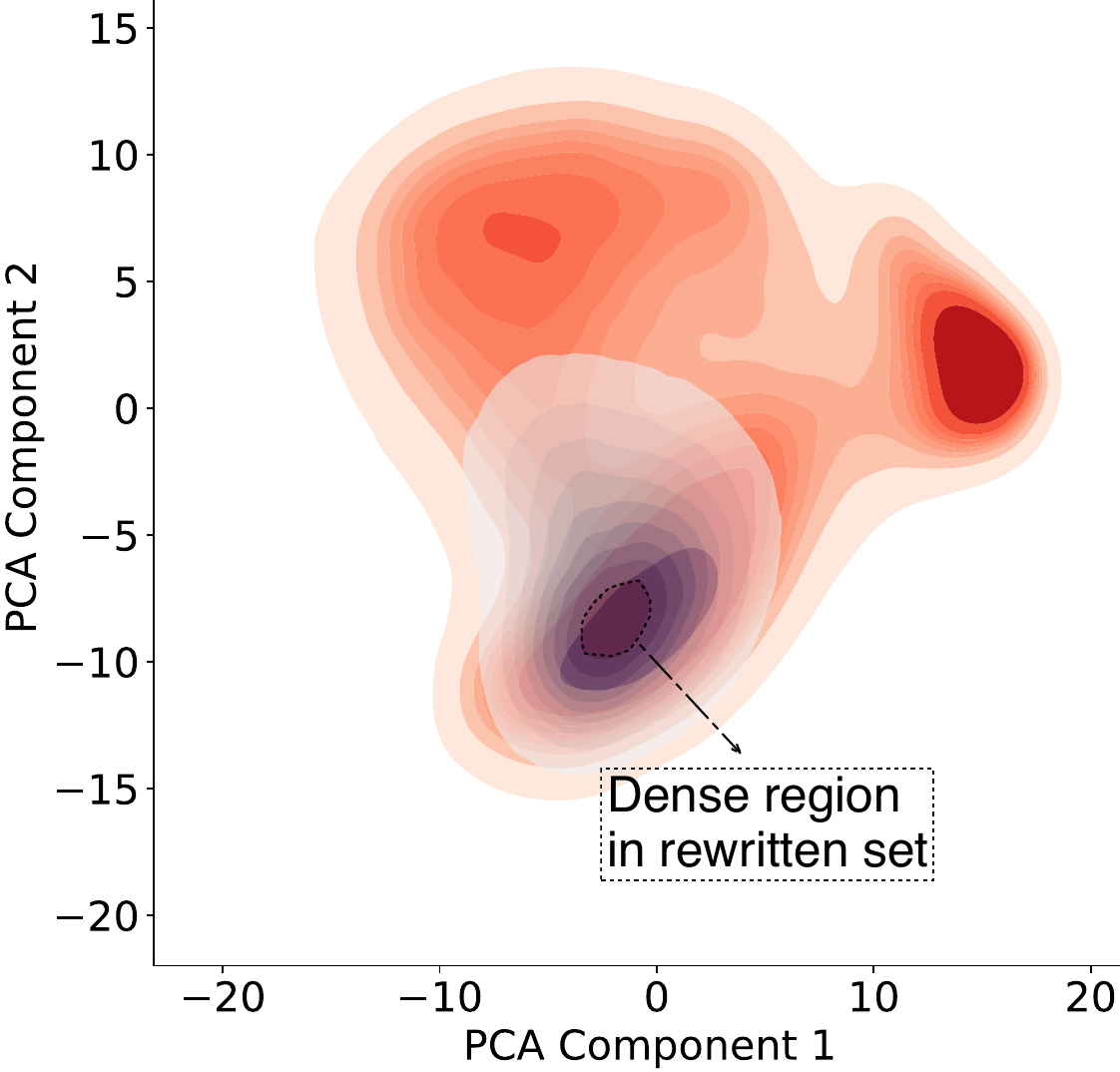}
    \subcaption{The rewritten set}
    \label{fig:sub7}
\end{subfigure}%
\begin{subfigure}{0.245\linewidth}
    \centering
    \includegraphics[width=\linewidth]{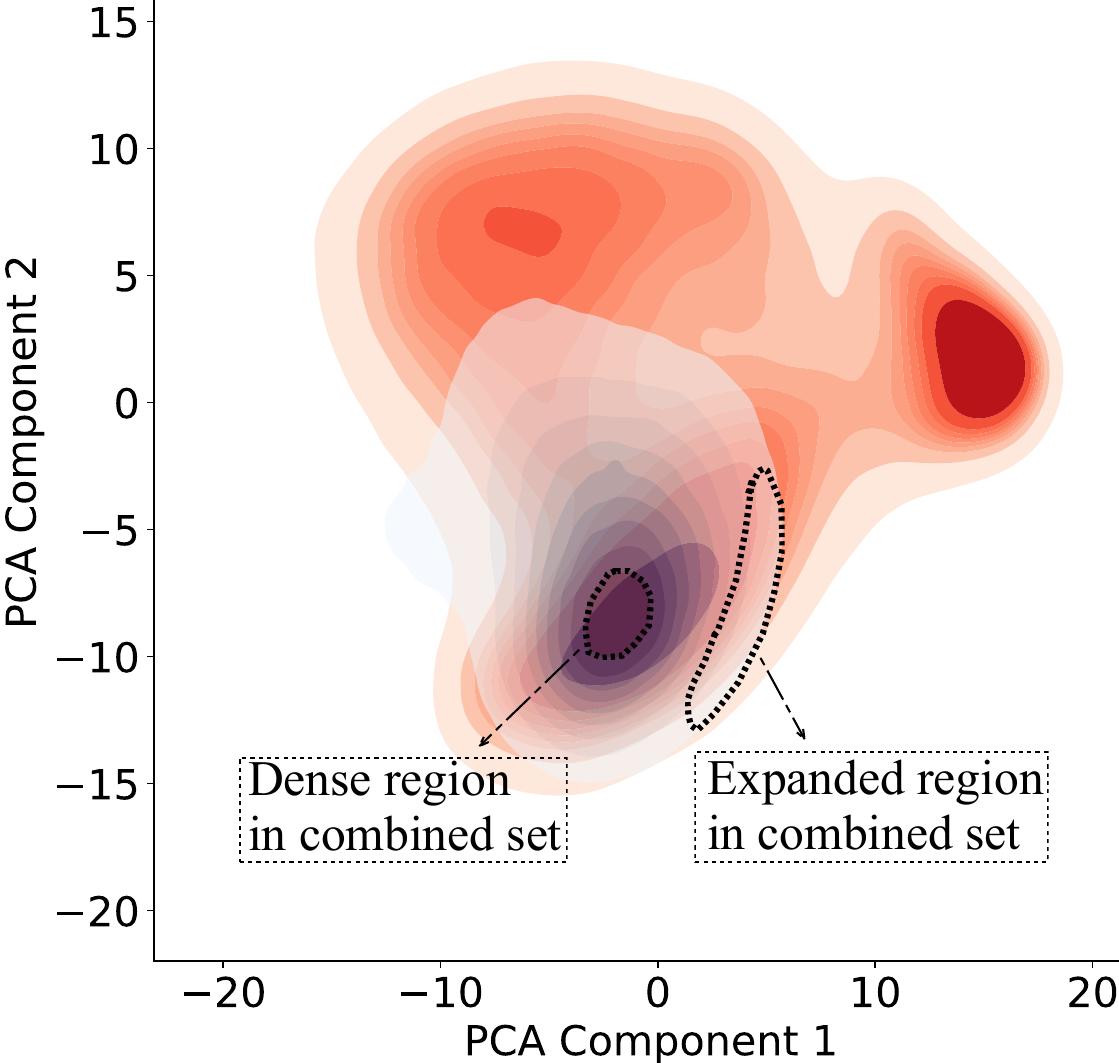}
    \subcaption{The combined set}
    \label{fig:sub8}
\end{subfigure}%
\end{center}

\caption{\textbf{Data Distribution Changes in \ABMN}. Subfigures (a)-(d) and (e)-(h) illustrate the distribution changes of the datasets under density estimation and the density comparison settings. The red region at the bottom represents the pre-training corpus, Dolma, while the blue regions in the subfigures represent the projections of Tulu V2, the difference set, the rewritten set, and the combined set, respectively. Darker areas indicate a higher concentration of data points, whereas lighter areas signify sparser distributions. The examples in the subfigures can be found in \autoref{appendix:examples}.}

\label{fig:distribution}
\end{figure*}

\subsection{Training Setting}
We use combined datasets in \ABMN to train three open-source models: OLMo, MAP-Neo, and Pythia. The rewritten difference set in the combined datasets is obtained by subtracting the corresponding SFT datasets (TuluV2, Neo-SFT, TuluV2) from the respective pre-training corpora (Dolma, Matrix, and Pile). Since the SFT dataset for Pythia has not been released, we use TuluV2 as a substitute. Full-parameter fine-tuning is applied, with the detailed training parameters provided in \autoref{appendix:parameters}.

\section{Results}

\begin{table*}[!htb]
\caption{\textbf{Main Results: Experiment performance of different models across various benchmarks}. $\Delta$ represents the change in performance when using \ABMN compared to the corresponding baseline. P-S, I-S, P-L, and I-L denote prompt-level strict accuracy, instance-level strict accuracy, prompt-level loose accuracy, and instance-level loose accuracy, respectively.}
\label{results:main}
\centering
\resizebox{0.95\textwidth}{!}{
    \begin{tabular}{@{}lccccccccccccccc@{}}
    \toprule
        \multicolumn{2}{l}{\multirow{4}{*}{\textbf{\makecell{Experiment \\Setting}}}} &\multicolumn{4}{c}{\textbf{Chat Benchmark}}&\multicolumn{7}{c}{\textbf{Standard Benchmark}} & \multirow{4}{*}{\textbf{Average}}\\ \cmidrule(r){3-6} \cmidrule(r){7-13}
        & & \multicolumn{4}{c}{IFEval} & \multicolumn{3}{c}{Exam} & \multicolumn{2}{c}{Coding} & \multicolumn{2}{c}{Reasoning} \\ \cmidrule(r){3-6} \cmidrule(r){7-9} \cmidrule(r){10-11} \cmidrule(r){12-13}
        & & \makecell{P-S} & \makecell{I-S} & \makecell{P-L} & \makecell{I-L} & MMLU & ARC & GPQA &\makecell{Human\\Eval} & MBPP & \makecell{Hella\\Swag} & \makecell{GSM\\8K} \\
    \midrule
        \multicolumn{2}{l}{OLMo-SFT}& 35.3 & 46.5 & 38.6 & 50.2 & 52.9 & 63.7 & 17.7 & 26.8 & 43.9 & 60.4 & 26.8 & 42.1\\
        \rowcolor{LightCyan}
        \multicolumn{2}{l}{$\Delta$ over OLMo}& +3.3 & +2.4 & +1.7 & +1.2 & +2.8 & +12.9 & +7.1 & +1.3 & +0.3 & +4.6 & +4.1 & +3.8\\
        \multicolumn{2}{l}{Neo-SFT}& 37.9 & 49.2 & 41.2 & 52.3 & 57.6 & 77.6 & 12.1 & 44.5 & 45.0 & 72.1 & 70.1 & 50.9 \\
        \rowcolor{LightCyan}
        \multicolumn{2}{l}{$\Delta$ over Neo}& +0.6 & +0.8 & +0.4 & +1.1 & +0.8 & +3.4 & +7.1 & -6.1 & +6.3 & -3.9 & +1.9 & +1.1 \\
        \multicolumn{2}{l}{Pythia-SFT}& 20.2 & 32.3 & 22.2 & 34.7 & 24.2 & 27.8 & 20.2 & 13.4 & 19.6 & 26.0 & 7.7 & 22.5 \\
        \rowcolor{LightCyan}
        \multicolumn{2}{l}{$\Delta$ over Pythia}& +1.8 & +1.6 & +2.4 & +1.6 & +1.0 & -4.7 & +4.6 & +1.8 & +0.2 & +0.1 & -0.1 & +0.9 \\
        
    \bottomrule
    \end{tabular}
}
\end{table*}

\subsection{Distribution Change Analysis}\label{Distributionchanges}
In the density estimation setting, \ABMN focuses on the dense regions of the SFT dataset and the pre-training corpus to identify points in the pre-training corpus that are underrepresented in the SFT dataset. \autoref{fig:sub1} highlights the dense regions of Tulu and Dolma (examples are provided in \autoref{appendix:examples}). Dense regions 1 and 2 correspond to code and scientific literature data, respectively. \autoref{fig:sub2} demonstrates that the difference set avoids the dense regions in the SFT dataset and aligns with dense regions of Dolma. \autoref{fig:sub3} shows the narrowing of the distribution during rewriting (examples are provided in \autoref{appendix:examples}), while \autoref{fig:sub4} indicates that the combined dataset expands the original SFT distribution and highly overlaps with the dense regions of the pre-training corpus. In the density comparison setting (\autoref{fig:sub5}-\ref{fig:sub8}), \ABMN focuses on points where the pre-training corpus has a higher density than the SFT dataset. Similarly, \ABMN with the density comparison setting can also expand the coverage of the existing dataset and optimize the data distribution.


\subsection{Main Results}

\begin{table*}[!htb]
\caption{\textbf{The Results of Various Difference Set Generation setting}. bge and MiniLM represent the embedding model, and estimation and comparison represent the setting of choosing difference sets. P-S, I-S, P-L, and I-L denote prompt-level strict accuracy, instance-level strict accuracy, prompt-level loose accuracy, and instance-level loose accuracy, respectively.}
\label{results:different set generation}
\centering
\resizebox{0.95\textwidth}{!}{
    \begin{tabular}{@{}lccccccccccccccc@{}}
    \toprule
        \multicolumn{2}{l}{\multirow{4}{*}{\textbf{\makecell{Experiment \\Setting}}}} &\multicolumn{4}{c}{\textbf{Chat Benchmark}}&\multicolumn{7}{c}{\textbf{Standard Benchmark}} & \multirow{4}{*}{\textbf{Average}}\\ \cmidrule(r){3-6} \cmidrule(r){7-13}
        & & \multicolumn{4}{c}{IFEval} & \multicolumn{3}{c}{Exam} & \multicolumn{2}{c}{Coding} & \multicolumn{2}{c}{Reasoning} \\ \cmidrule(r){3-6} \cmidrule(r){7-9} \cmidrule(r){10-11} \cmidrule(r){12-13}
        & & \makecell{P-S} & \makecell{I-S} & \makecell{P-L} & \makecell{I-L} & MMLU & ARC & GPQA &\makecell{Human\\Eval} & MBPP & \makecell{Hella\\Swag} & \makecell{GSM\\8K} \\
    \midrule
        \multicolumn{2}{l}{OLMo-SFT}& 35.3 & 46.5 & 38.6 & 50.2 & 52.9 & 63.7 & 17.7 & 26.8 & 43.9 & 60.4 & 26.8  & 42.1 \\
        \rowcolor{LightCyan}
        \multicolumn{2}{l}{bge-estimation ($\Delta$)}& +3.3 & +2.4 & +1.7 & +1.2 & +2.8 & +12.9 & +7.1 & +1.3 & +0.3 & +4.6 & +4.1 & +3.8 \\
        \rowcolor{LightCyan}
        \multicolumn{2}{l}{bge-comparison ($\Delta$)}& +2.6 & +2.5 & +1.1 & +1.1 & +2.8 & +10.5 & +10.6 & +4.9 & +3.7 & +2.9 & +4.7 & +4.3 \\
        \rowcolor{VeryLightPink}
        \multicolumn{2}{l}{ MiniLM-estimation ($\Delta$)}& -0.7 & +0.5 & -2.0 & -0.9 & +2.6 & +10.5 & +9.1 & +3.1 & +2.7 & +4.1 & +3.5 & +3.0 \\
        \rowcolor{VeryLightPink}
        \multicolumn{2}{l}{MiniLM-comparison ($\Delta$)}& +0.9 & +1.0 & -0.1 & +0.2 & +2.6 & +10.9 & +8.6 & +1.9 & +0.5 & +3.1 & +4.8 & +3.1 \\
    \bottomrule
    \end{tabular}
}
\end{table*}

\begin{table*}[!htb]
\caption{\textbf{The Ablation Results on Data Size and Distillation}. P-S, I-S, P-L, and I-L denote prompt-level strict accuracy, instance-level strict accuracy, prompt-level loose accuracy, and instance-level loose accuracy, respectively.}
\label{results:ablation}
\centering
\resizebox{0.95\textwidth}{!}{
    \begin{tabular}{@{}lccccccccccccccc@{}}
    \toprule
        \multicolumn{2}{l}{\multirow{4}{*}{\textbf{\makecell{Experiment \\Setting}}}} &\multicolumn{4}{c}{\textbf{Chat Benchmark}}&\multicolumn{7}{c}{\textbf{Standard Benchmark}} & \multirow{4}{*}{\textbf{Average}}\\ \cmidrule(r){3-6} \cmidrule(r){7-13}
        & & \multicolumn{4}{c}{IFEval} & \multicolumn{3}{c}{Exam} & \multicolumn{2}{c}{Coding} & \multicolumn{2}{c}{Reasoning} \\ \cmidrule(r){3-6} \cmidrule(r){7-9} \cmidrule(r){10-11} \cmidrule(r){12-13}
        & & \makecell{P-S} & \makecell{I-S} & \makecell{P-L} & \makecell{I-L} & MMLU & ARC & GPQA &\makecell{Human\\Eval} & MBPP & \makecell{Hella\\Swag} & \makecell{GSM\\8K} \\
    \midrule
        \multicolumn{2}{l}{OLMo-SFT} & 35.3 & 46.5 & 38.6 & 50.2 & 52.9 & 63.7 & 17.7 & 26.8 & 43.9 & 60.4 & 26.8  & 42.1 \\
        \rowcolor{LightCyan}
        \multicolumn{2}{l}{Distillation ($\Delta$)} & -4.1 & -3.8 & -4.6 & -4.5 & +0.9 & +4.1 & +4.0 & -3.6 & -2.6 & -6.9 & +14.1 & -0.6 \\
        \rowcolor{LightCyan}
        \multicolumn{2}{l}{Same Size ($\Delta$)} & +0.4 & +0.1 & -0.5 & -1.0 & +2.6 & +10.5 & +12.6 & -2.4 & -0.3 & -0.9 & +1.9 & +2.1 \\
        \rowcolor{LightCyan}
        \multicolumn{2}{l}{OLMo ($\Delta$)} & +3.3 & +2.4 & +1.7 & +1.2 & +2.8 & +12.9 & +7.1 & +1.3 & +0.3 & +4.6 & +4.1 & +3.8 \\
    \bottomrule
    \end{tabular}
}
\end{table*}

As shown in Table \ref{results:main}, compared to the SFT model of OLMo, MAP-Neo, and Pythia baselines, the counterparts trained with \ABMN achieve \textbf{average performance improvements} of 3.8, 1.1 and 0.9 across eight benchmarks. This illustrates the effectiveness of \ABMN. \textbf{We suppose that this improvement results from \ABMN supplementing the original SFT dataset with lacking data, expanding its coverage, and optimizing its distribution.}

Based on the analysis in \autoref{Distributionchanges}, we can summarize two points supporting the above supposition: 
(1) A comparison of~\autoref{fig:sub1} and~\ref{fig:sub2} reveals that the difference set includes data from the pre-training corpus that is lacking in SFT datasets, such as code and scientific literature data. 
(2) Although the distribution narrows during the rewriting process (as shown in \autoref{fig:sub2} and \ref{fig:sub3}), the final combined dataset expands the coverage of the original SFT dataset, and the dense regions of the combined data align closely those of the pre-training corpus (as shown in~\autoref{fig:sub4}).

\subsection{Difference Set Generation Setting Results}
Table \ref{results:different set generation} presents the experimental results for various embedding models and different set generation settings. As shown in Table \ref{results:different set generation}, the four \ABMN variants show improvements over the baseline model OLMo-SFT across various settings: using the bge model with density estimation to identify the difference set achieves an average absolute improvement of 3.8; using bge with density comparison yields an improvement of 4.3; using MiniLM with density comparison results in an improvement of 3.0; and using bge with density comparison achieves an improvement of 3.1. These results suggest that \ABMN is robust across various choices of embedding model and difference set generation method.

\begin{figure*}[!htb]
\centering
\resizebox{0.95\textwidth}{!}{
    \begin{subfigure}{0.33\linewidth}
        \centering
        \includegraphics[width=\linewidth]{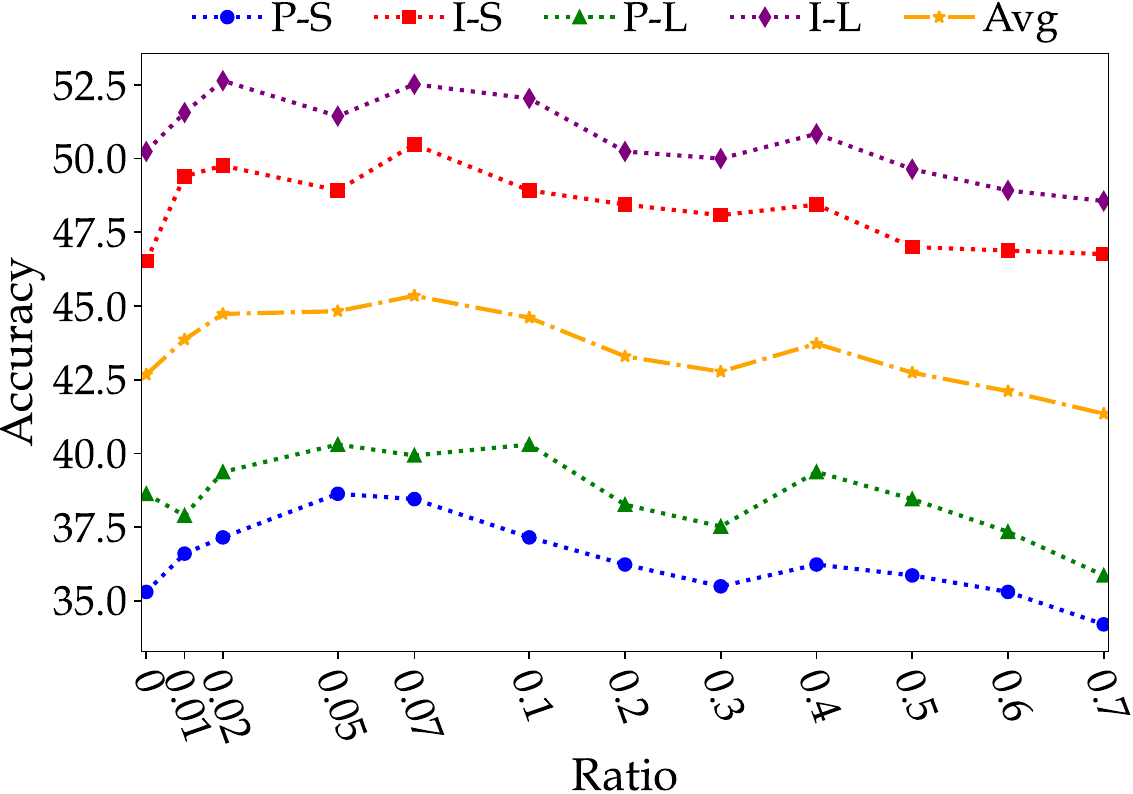}
        \subcaption{IFEval across different ratios}
        \label{fig:subIFEval}
    \end{subfigure}%
    \begin{subfigure}{0.33\linewidth}
        \centering
        \includegraphics[width=\linewidth]{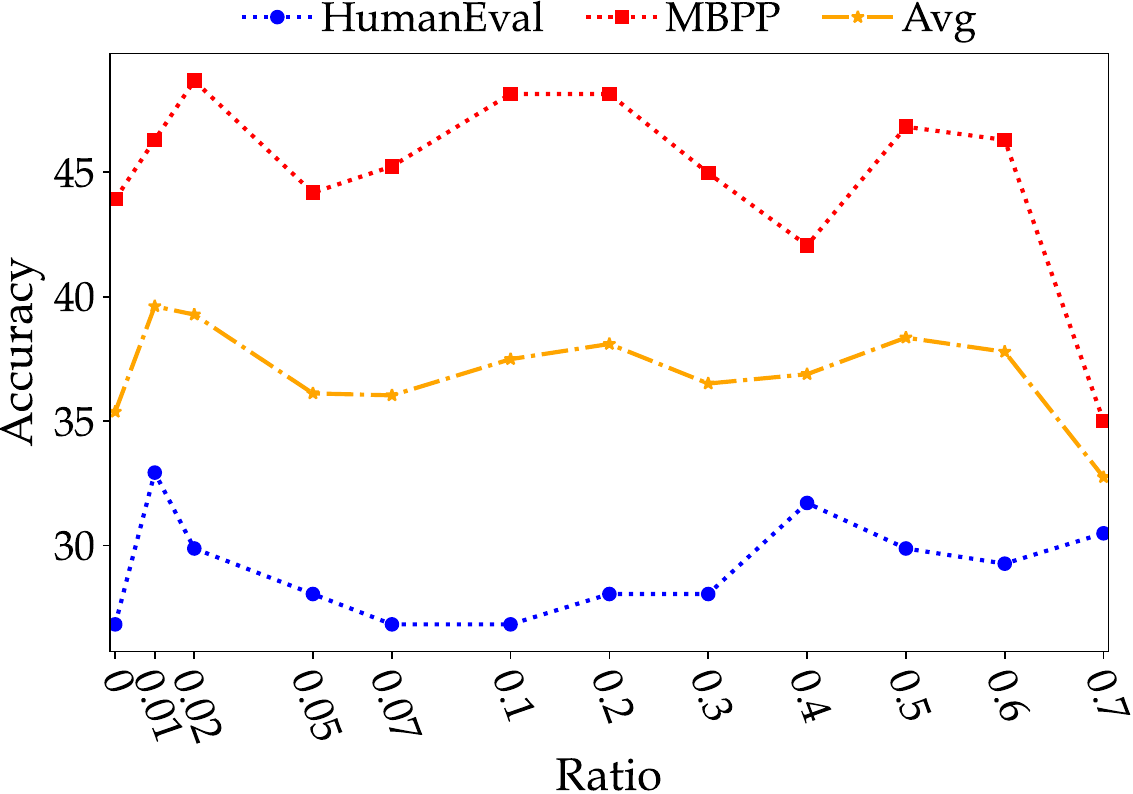}
        \subcaption{Coding across different ratios}
        \label{fig:subcoding}
    \end{subfigure}%
    \begin{subfigure}{0.33\linewidth}
        \centering
        \includegraphics[width=\linewidth]{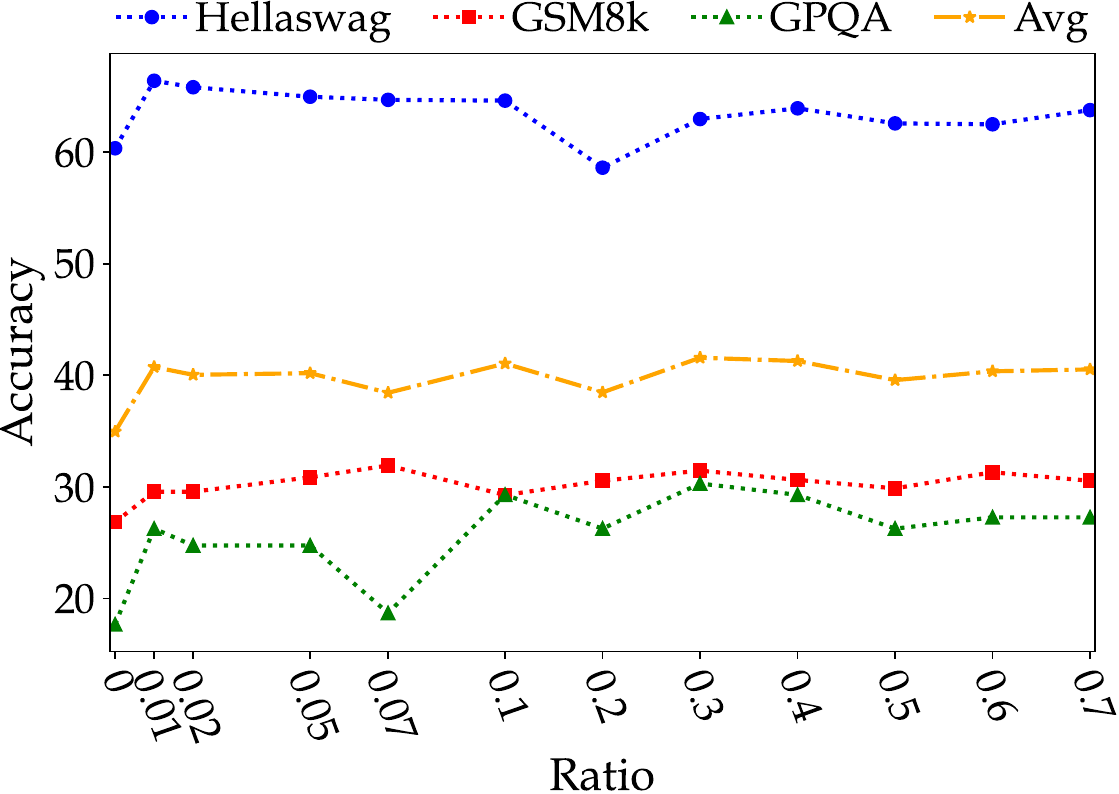}
        \subcaption{Reasoning across different ratios}
        \label{fig:subreasoning}
    \end{subfigure}%
}
\caption{\textbf{Line graph across different ratios}. The x-label represents the ratio of the rewritten set to the original SFT dataset, while the y-label shows accuracy across different benchmarks. More results can be found in \autoref{appendix:ratios}.}
\label{fig:results_ratios}

\end{figure*}

\begin{figure}[!tb]
\centering
\begin{subfigure}{\linewidth}
    \centering
    \includegraphics[width=\linewidth]{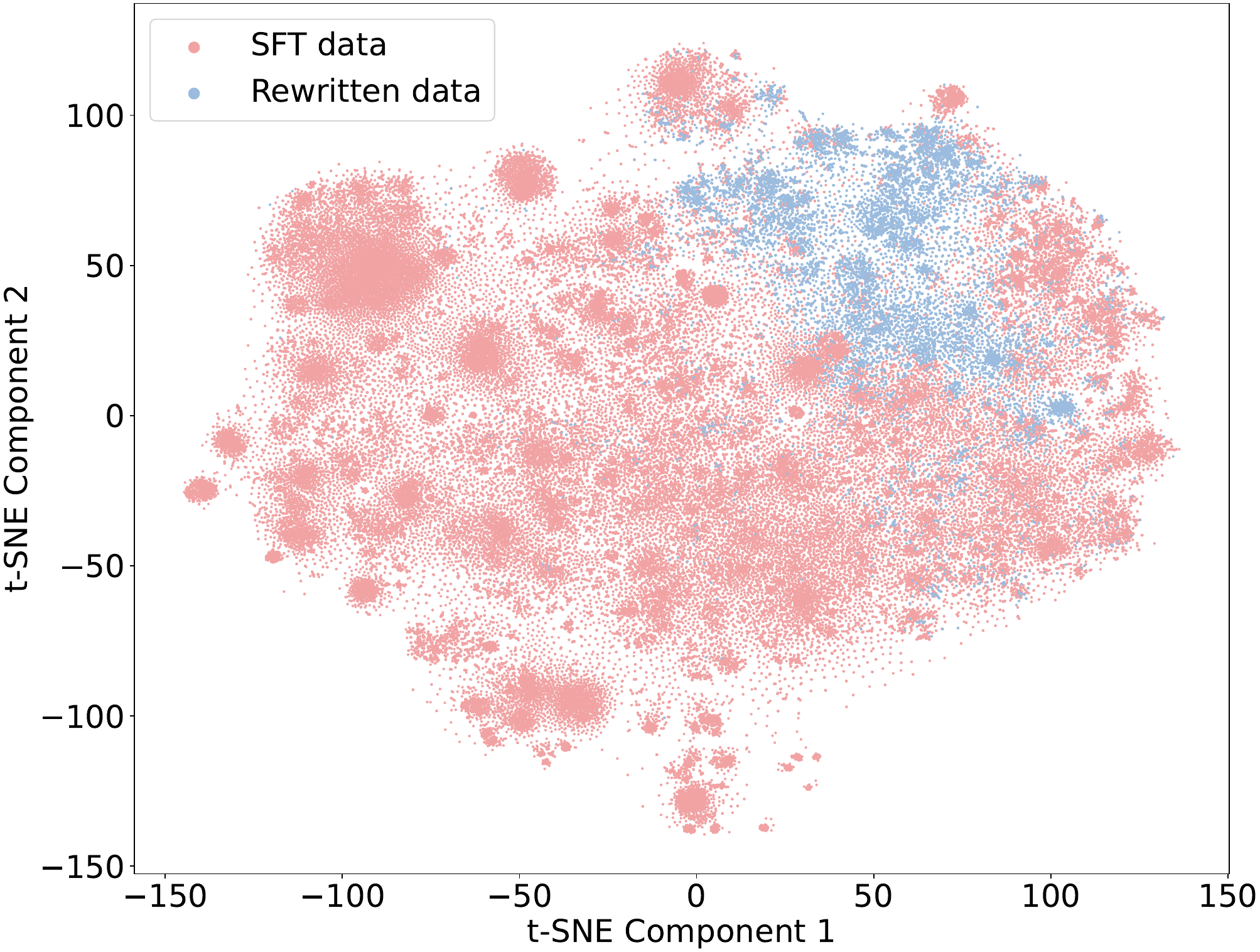}
\end{subfigure}%
\caption{\textbf{The t-SNE Visualization of SFT and Rewritten Data.} The red points and blue points represent the original SFT data and the rewritten data, respectively.}
\label{fig:tsne}
\end{figure}

\subsection{Ablation Results}\label{results:datasize}
To verify whether the gains of \ABMN result from the increased size of the SFT dataset after adding the rewritten difference set, we sample a subset from the combined dataset (original SFT and rewritten difference set) that is equal in size to the original SFT dataset and use it for training. Comparing the first and third rows in Table 3, the \ABMN method achieves an average absolute improvement of 2.1, even with the same dataset size. Comparing the third and fourth rows, the improvement for the same dataset size setting is smaller than the final \ABMN improvement. 

Additionally, to test whether the improvement arises from distillation by a stronger model during the rewriting phase, we replace the original SFT dataset with the rewritten dataset from the same distribution and train the model on a combined dataset (rewritten same distribution set and rewritten difference set). Comparing the first and second rows in Table~\ref{results:ablation}, the distillation setting does not outperform the OLMo-SFT baseline, likely because the quality of the rewritten data is lower than that of the original SFT dataset. This indicates that the improvement does not result from distillation by an aligned model.

\subsection{Ratio Results}\label{result:ratio}
We further investigate the effect of incorporating various ratios of rewritten difference data on \ABMN. As shown in Figure \ref{fig:results_ratios}, the \ABMN achieves excellent performance with a rewritten data set comprising less than 10 \% of the original SFT dataset. However, performance declines as the size of the rewritten set increases. We hypothesize that incorporating a small amount of rewritten data improves model performance significantly by filling gaps in the original SFT data. On the other hand, the quality of the rewritten data might be low, which could degrade the overall data quality when the rewritten ratio is increased. This is consistent with the ablation study on data size in Section \ref{results:datasize}, which shows that the quality of the rewritten data is lower than that of the original SFT dataset and that the improvement in \ABMN is not due to the increased data size.

\subsection{Visualization}
Figure \ref{fig:tsne} illustrates that the manually combined original SFT dataset (Tulu) forms multiple distinct clusters, indicating a high level of diversity within the original dataset. The rewritten data is densely distributed in areas underrepresented by the original SFT dataset, while intentionally avoiding regions where the original SFT dataset is densely populated. This result clearly demonstrates the effectiveness of the difference set generated by \ABMN in optimizing data coverage.

\section{Related Work}
\subsection{Open-Source Large Language Model}
Current models like GPT-4 \cite{openai2023gpt4}, Gemini \cite{team2023gemini}, and Claude \cite{anthropic2024claude3} have demonstrated impressive performance across various fields. However, their closed-source nature and API-only access limit deployment flexibility. To address this, several open-source models, such as LLaMA \cite{touvron2023llama} , Qwen \cite{yang2024qwen2}, DeepSeek\cite{deepseek-ai2024deepseek}, ChatGLM \cite{glm2024chatglm}, Mixtral \cite{jiang2024mixtral}, and Yi \cite{ai2024yi} have emerged, offering freely accessible model weights. Furthermore, some open-source communities have introduced fully transparent models, such as OLMo \cite{groeneveld2024olmo}, Map-Neo \cite{zhang2024mapneo}, LLM360 \cite{liu2023llm360}, and Pythia \cite{biderman2023pythia}, which go beyond sharing model weights by providing accessible pre-training corpora, SFT datasets, data-cleaning processes, intermediate checkpoints, and reproducible code, fostering a more open and reproducible research ecosystem. In this paper, we primarily conduct experiments on fully transparent open-source models due to their accessible pre-training and SFT datasets. Notably, our method can also be applied to enhance the performance of closed-source models or those that provide open-access weights.

\subsection{Instruction Tuning}

Instruction tuning evolves from relying on human-annotated data to incorporating synthetic data, aiming to enhance the adaptability and generalization of pre-trained language models. Initially, instruction tuning involves training models on diverse instruction-response pairs from manually curated datasets, such as FLAN \cite{wei2021finetuned} and T0 \cite{sanh2021multitask}, which significantly improve zero-shot and few-shot learning performance. To further enhance cross-task generalization, multi-task learning approaches, like UnifiedQA \cite{khashabi2020unifiedqa} and FLAN-T5 \cite{chung2024scaling}, present multiple tasks as instructions, reducing the need for task-specific data and manual prompt engineering. As instruction tuning progresses, the importance of large-scale, diverse datasets becomes evident. Datasets like Super-Natural Instructions \cite{supernaturalinstructions} provide extensive coverage across tasks, domains, and instruction styles, improving model robustness and mitigating biases. Additionally, the exploration of synthetic data generation techniques augments training sets, enabling models to better handle rare or complex instructions \cite{xie2024monte, asai2023selfrag}. These approaches, which leverage language models to generate additional training samples, demonstrate significant improvements in both performance and generalization.

\subsection{Improving LLM Using Synthetic Data}
Some methods enhance model capabilities by synthesizing data using external signals, such as seed data \cite{wang2022selfinstruct,sun2023principledriven,kang-etal-2024-self,liang2024isheep,alpaca}, pre-training data \cite{li2023selfalignment,zheng2024kun}, query data \cite{huang-etal-2023-large,madaan2023selfrefine,yu2023metamath}, feedback data \cite{lu2023self,scheurer2022training}, and retrieval-augmented generation (RAG) \cite{asai2023selfrag}.  These methods can be classified into two types: those that generate synthetic data using the model itself \cite{liang2024isheep,wang2022selfinstruct,sun2023principledriven} and those that use a teacher model for data synthesis \cite{lee2024llm2llm,DBLP:conf/acl/LiCCHGZ24,alpaca}.  While synthetic data approaches effectively mitigate the limitations of supervised dataset sizes, they also introduce challenges such as increased hallucinations, lack of diversity, low quality, and distribution misalignment \cite{liu2024best}. Training models iteratively with this synthetic data can lead to issues like model collapse, increased hallucinations, and reduced generalizability \cite{shumailov2023curse,alemohammad2023selfconsuming,guo-etal-2024-curious}.

Recent studies address these limitations through various methods. Some methods aim to improve the quality of generated instruction pairs using self-consistency\cite{huang-etal-2023-large}, reflection\cite{renze2024selfreflection,DBLP:conf/acl/LiCCHGZ24}, filtering \cite{liang2024isheep,yuan2024selfrewarding}, and Monte Carlo tree search (MCTS) \cite{xie2024monte,gao2024interpretable}. Others focus on enhancing diversity of generated instruction pairs \cite{ge2024scaling,o'neill2023steering}, reducing hallucinations \cite{chung2023increasing,zhang2024sled,jones2023teaching}, or optimizing synthetic data distribution \cite{lupidi2024source2synth,jiang2024importance,yang2024selfdistillation}. Our method mainly focuses on further enhancing the diversity of synthetic data after combining existing datasets manually.



\section{Conclusion}
The existing SFT datasets exhibit significant differences from the pre-training corpus in terms of coverage and distribution. In this paper, we present the \ABMN method, which adaptively fills the gaps in current manually-assembled SFT datasets by identifying the difference set between the pre-training corpus and the SFT dataset. This approach utilizes existing high-quality SFT data and offers guidance for synthesizing lacking data of existing SFT datasets. Our experiments demonstrate the effectiveness of \ABMN, showing that bridging the gap between SFT and pre-training datasets can be achieved by adding a small amount of difference data (less than 10 \%). This feature makes \ABMN a cost-effective and practical solution for real-world applications.

\bibliography{reference}
\bibliographystyle{icml2025}

\newpage
\appendix
\onecolumn
\section{Visualization of SFT Dataset Projections onto the Pre-training Corpus}\label{appendix:visualization}
\begin{figure*}[!hb]
\begin{center}
\begin{subfigure}{0.33\linewidth}
    \centering
    \includegraphics[width=\linewidth]{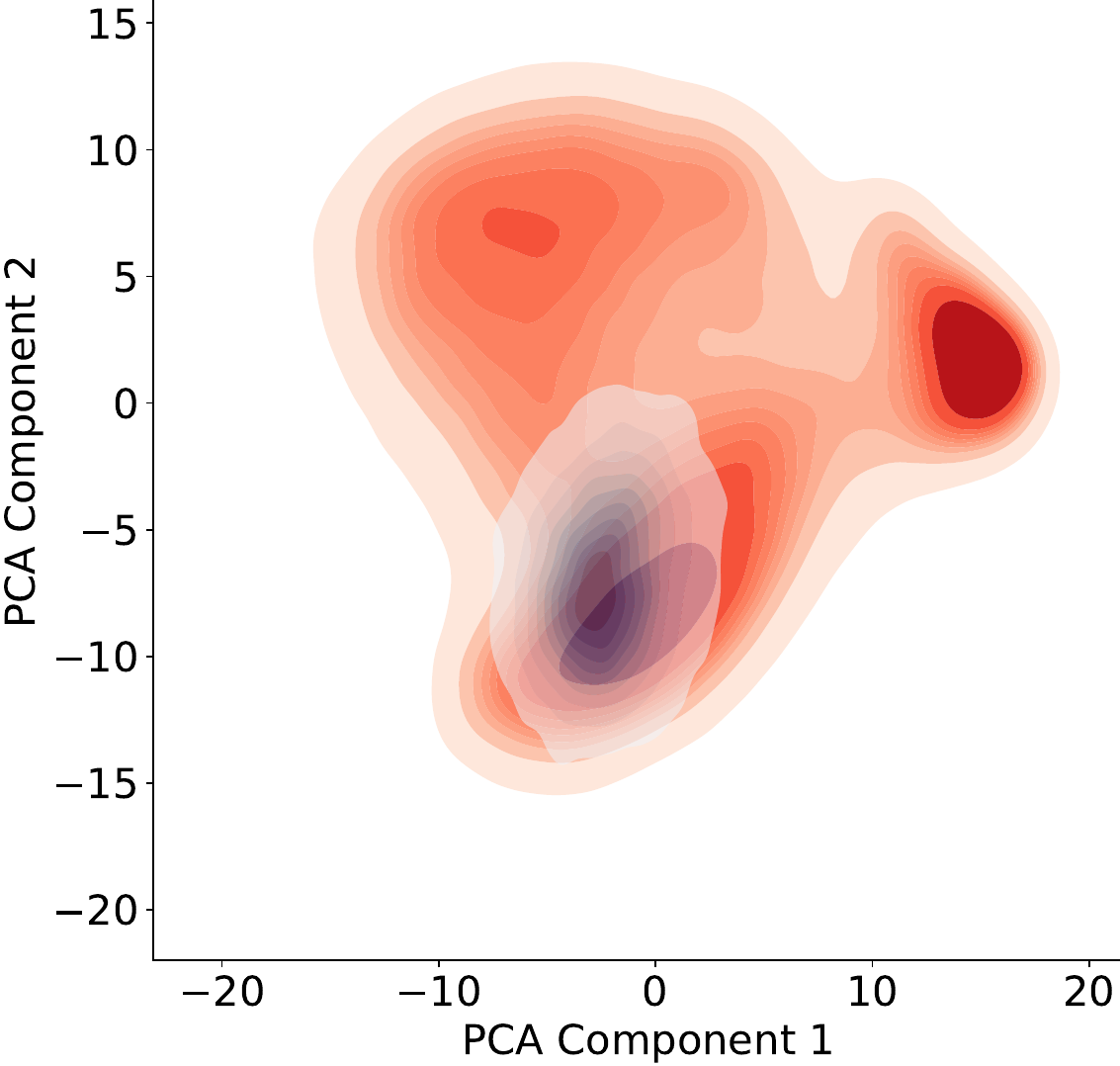}
    \subcaption{\href{https://huggingface.co/datasets/databricks/databricks-dolly-15k}{Dolly} VS \href{https://huggingface.co/datasets/allenai/dolma}{Dolma}}
    \label{fig:appendixsub1}
\end{subfigure}%
\begin{subfigure}{0.33\linewidth}
    \centering
    \includegraphics[width=\linewidth]{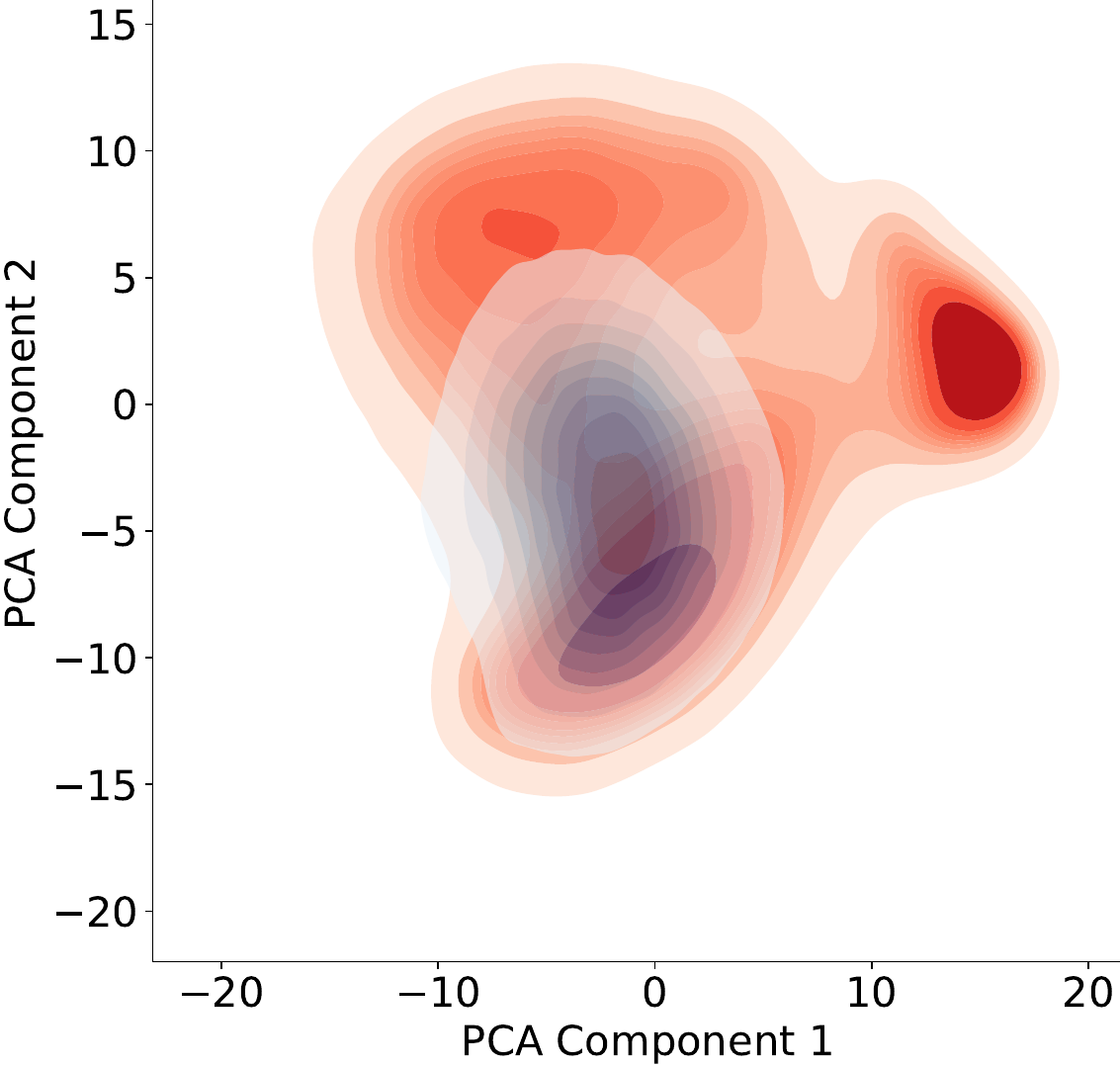}
    \subcaption{\href{https://huggingface.co/datasets/WizardLMTeam/WizardLM_evol_instruct_V2_196k}{EvolInstruct} VS \href{https://huggingface.co/datasets/allenai/dolma}{Dolma}}
    \label{fig:appendixsub2}
\end{subfigure}%
\begin{subfigure}{0.33\linewidth}
    \centering
    \includegraphics[width=\linewidth]{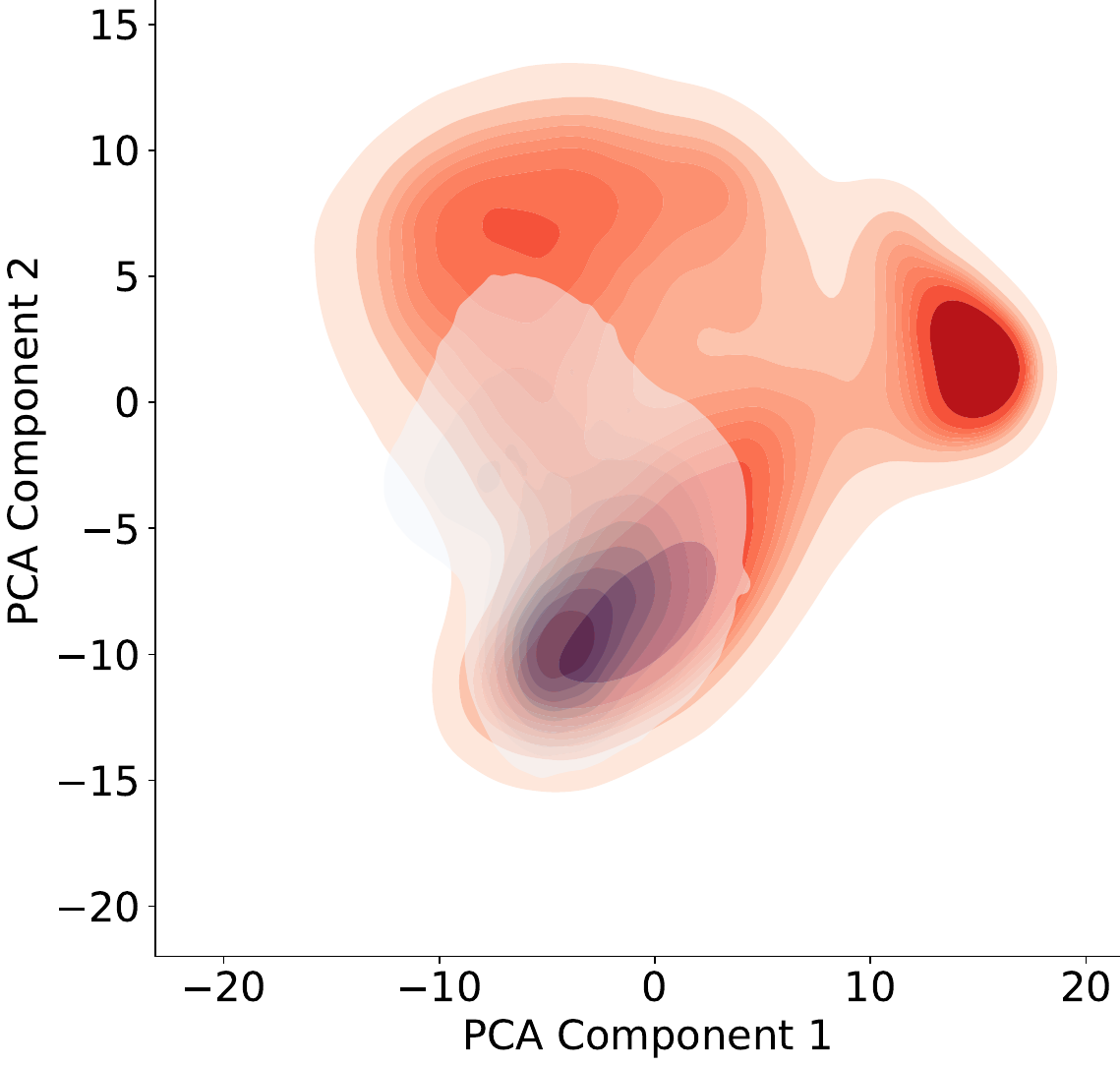}
    \subcaption{\href{https://github.com/multimodal-art-projection/MAP-NEO}{Neo-SFT} VS \href{https://huggingface.co/datasets/allenai/dolma}{Dolma}}
    \label{fig:appendixsub3}
\end{subfigure}%

\begin{subfigure}{0.33\linewidth}
    \centering
    \includegraphics[width=\linewidth]{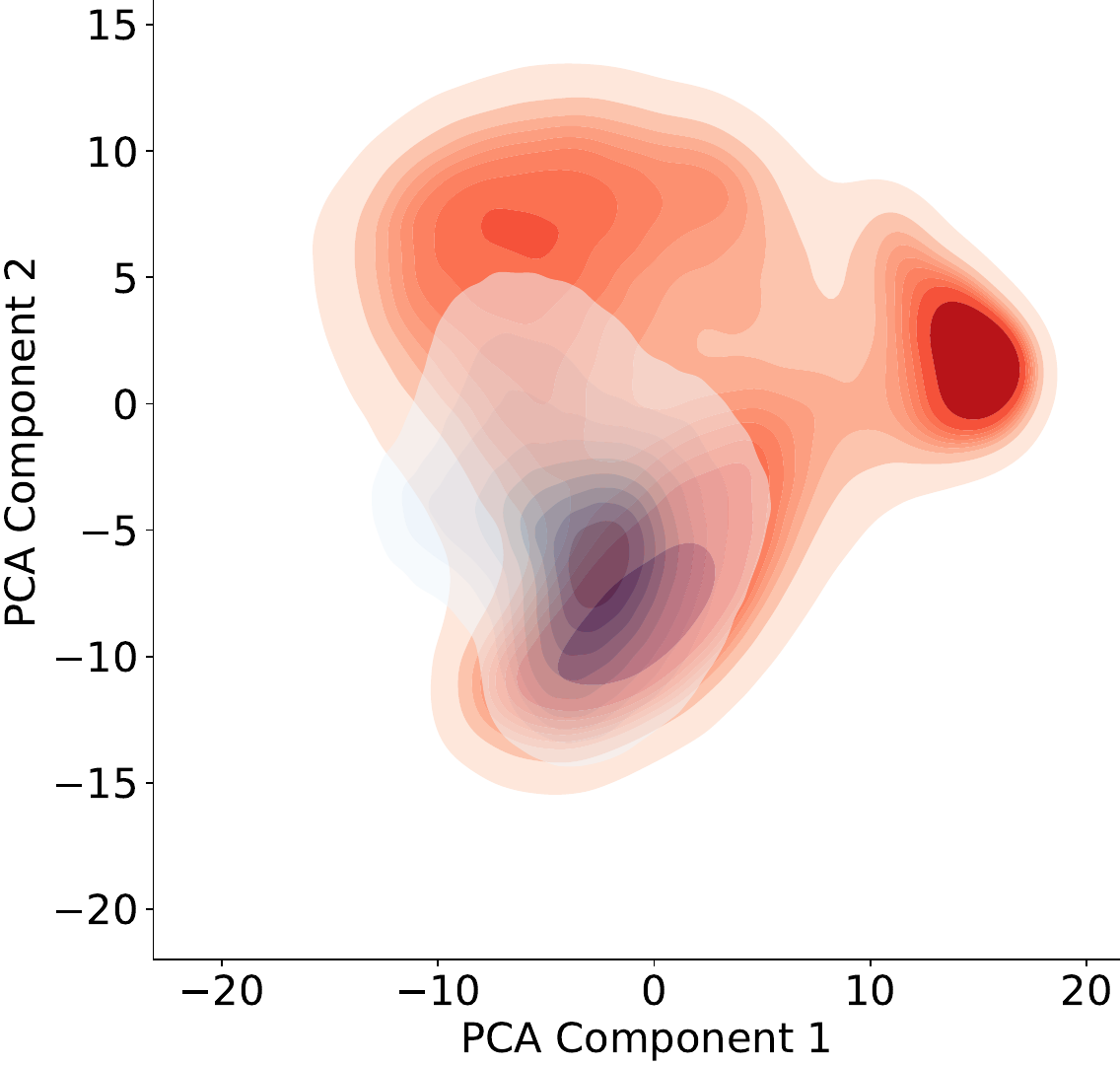}
    \subcaption{\href{https://huggingface.co/datasets/teknium/OpenHermes-2.5}{OpenHermes} VS \href{https://huggingface.co/datasets/allenai/dolma}{Dolma}}
    \label{fig:appendixsub4}
\end{subfigure}%
\begin{subfigure}{0.33\linewidth}
    \centering
    \includegraphics[width=\linewidth]{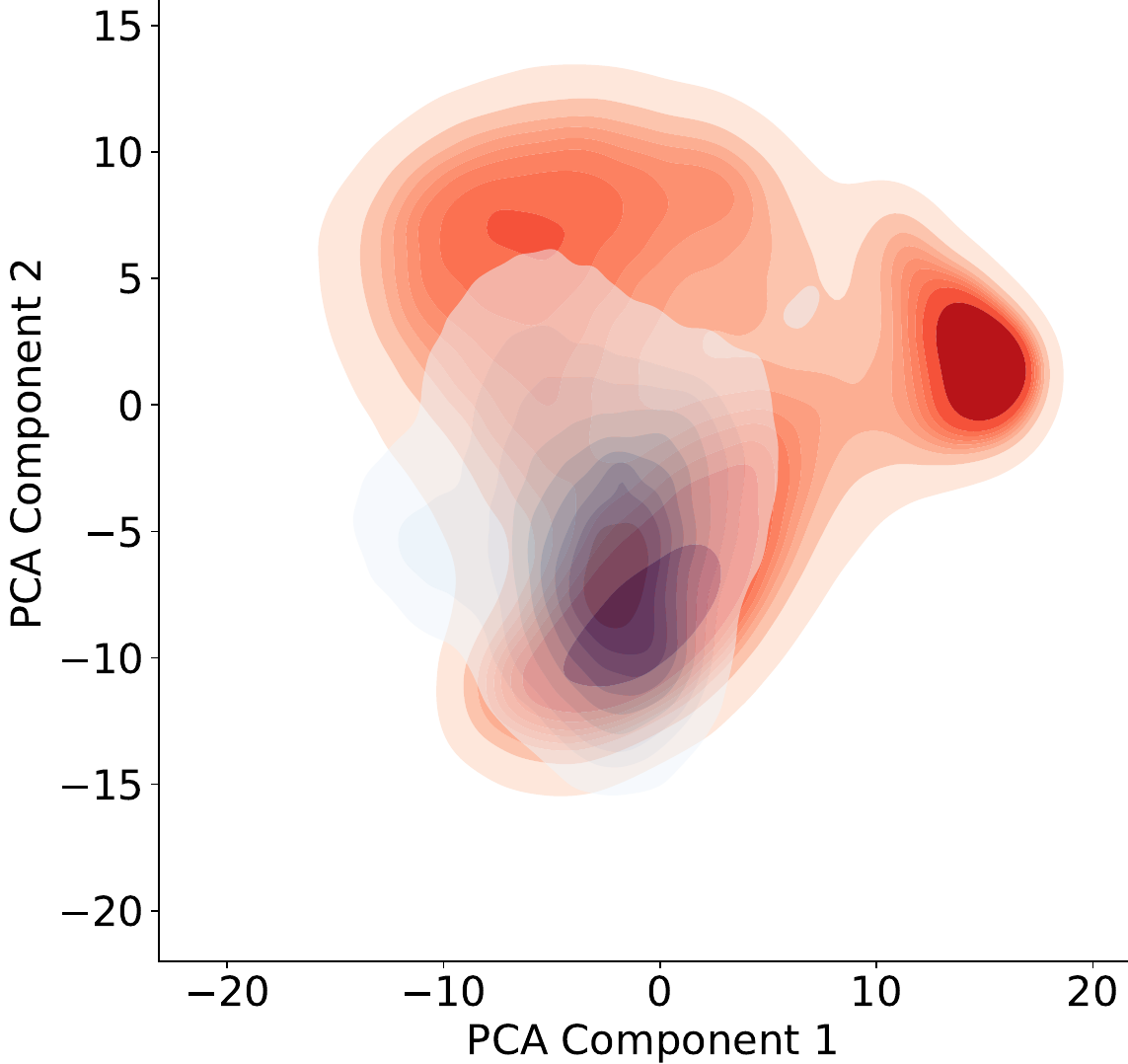}
    \subcaption{\href{https://huggingface.co/datasets/allenai/tulu-v2-sft-mixture}{Tulu} VS \href{https://huggingface.co/datasets/allenai/dolma}{Dolma}}
    \label{fig:appendixsub5}
\end{subfigure}%
\begin{subfigure}{0.33\linewidth}
    \centering
    \includegraphics[width=\linewidth]{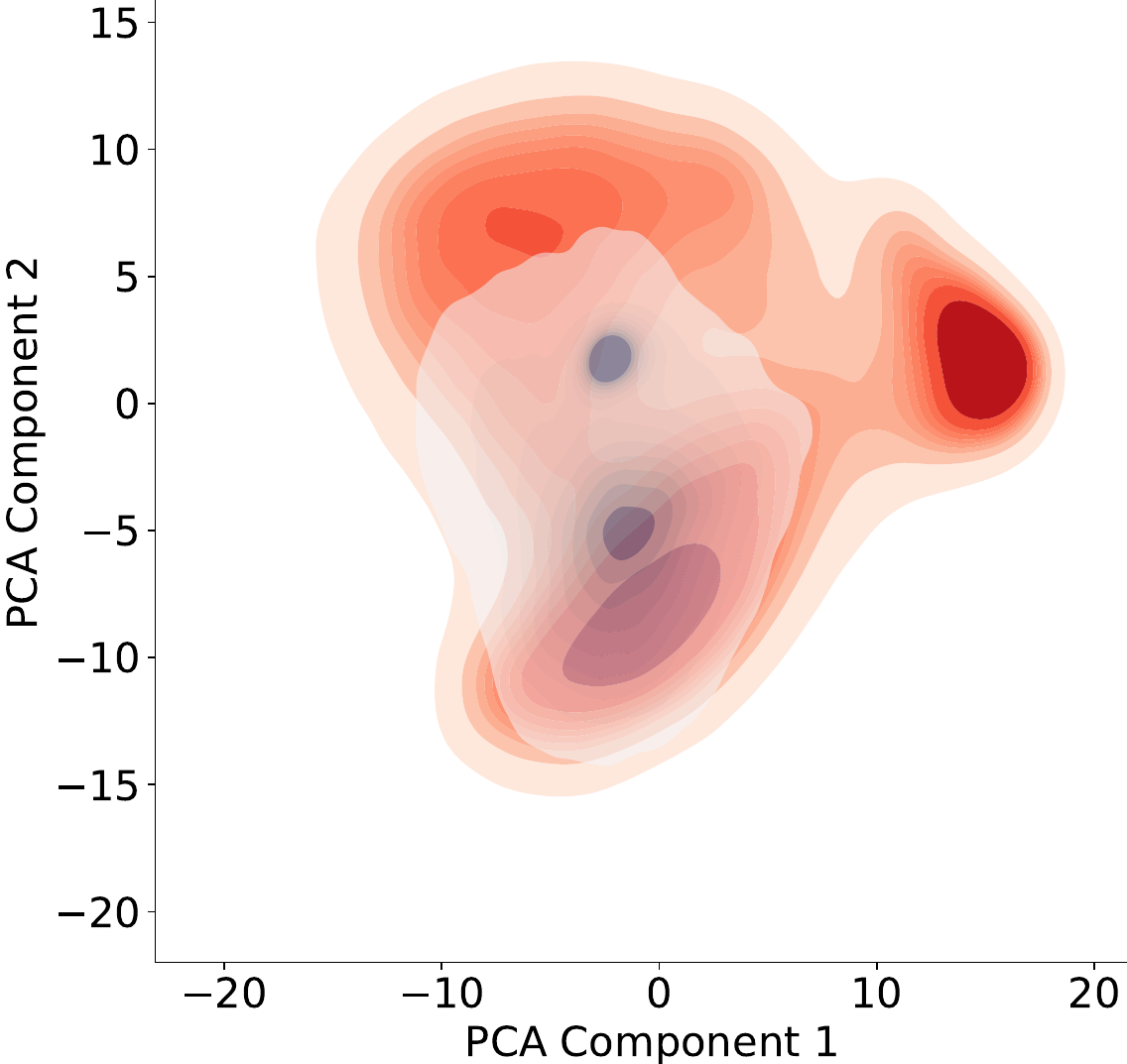}
    \subcaption{\href{https://huggingface.co/datasets/allenai/WildChat}{WildChat} VS \href{https://huggingface.co/datasets/allenai/dolma}{Dolma}}
    \label{fig:appendixsub6}
\end{subfigure}%

\begin{subfigure}{0.33\linewidth}
    \centering
    \includegraphics[width=\linewidth]{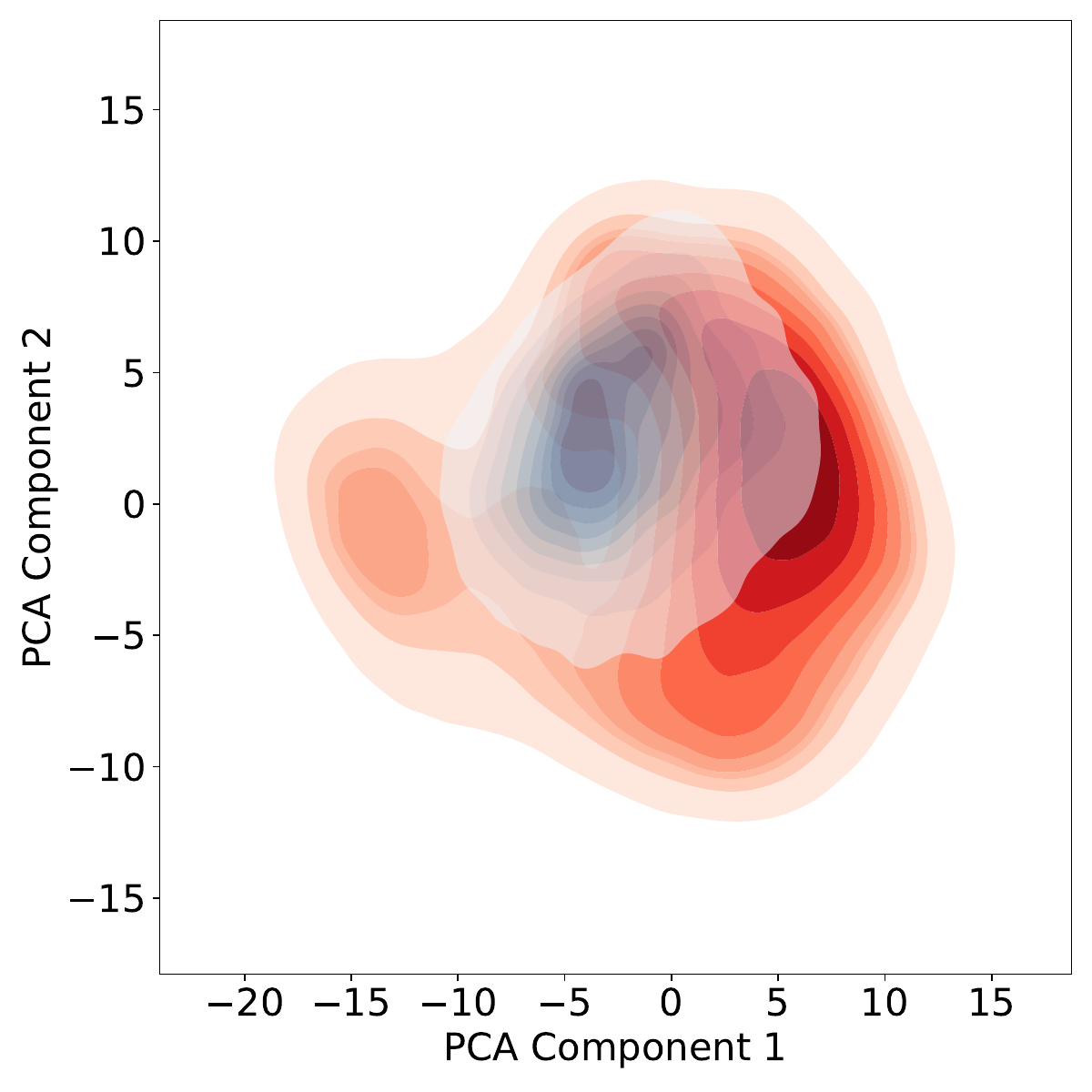}
    \subcaption{\href{https://huggingface.co/datasets/databricks/databricks-dolly-15k}{Dolly} VS \href{https://huggingface.co/datasets/HuggingFaceTB/cosmopedia}{Cosmopedia}}
    \label{fig:appendixsub7}
\end{subfigure}%
\begin{subfigure}{0.33\linewidth}
    \centering
    \includegraphics[width=\linewidth]{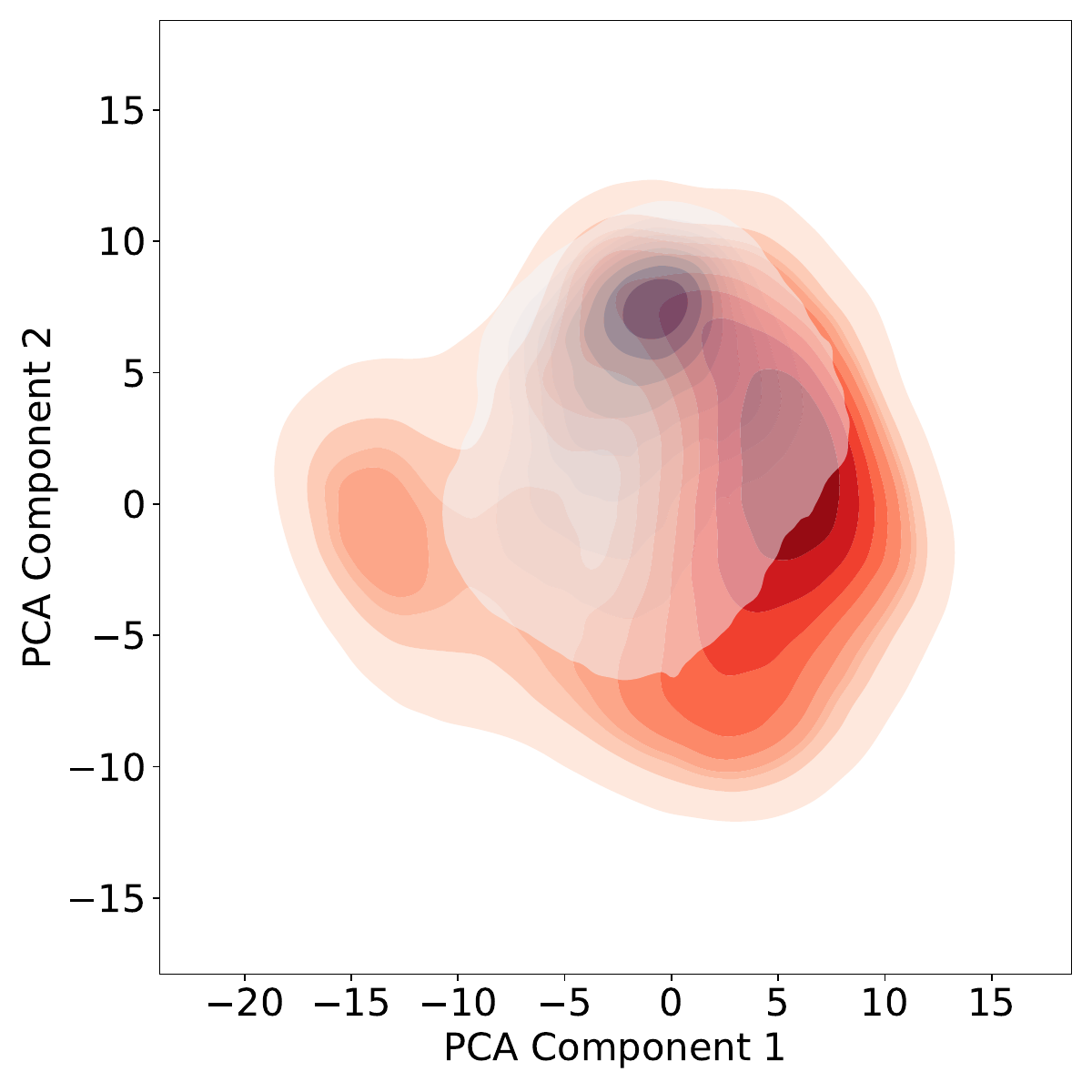}
    \subcaption{\href{https://github.com/multimodal-art-projection/MAP-NEO}{Neo-SFT} VS \href{https://huggingface.co/datasets/HuggingFaceTB/cosmopedia}{Cosmopedia}}
    \label{fig:appendixsub8}
\end{subfigure}%
\begin{subfigure}{0.33\linewidth}
    \centering
    \includegraphics[width=\linewidth]{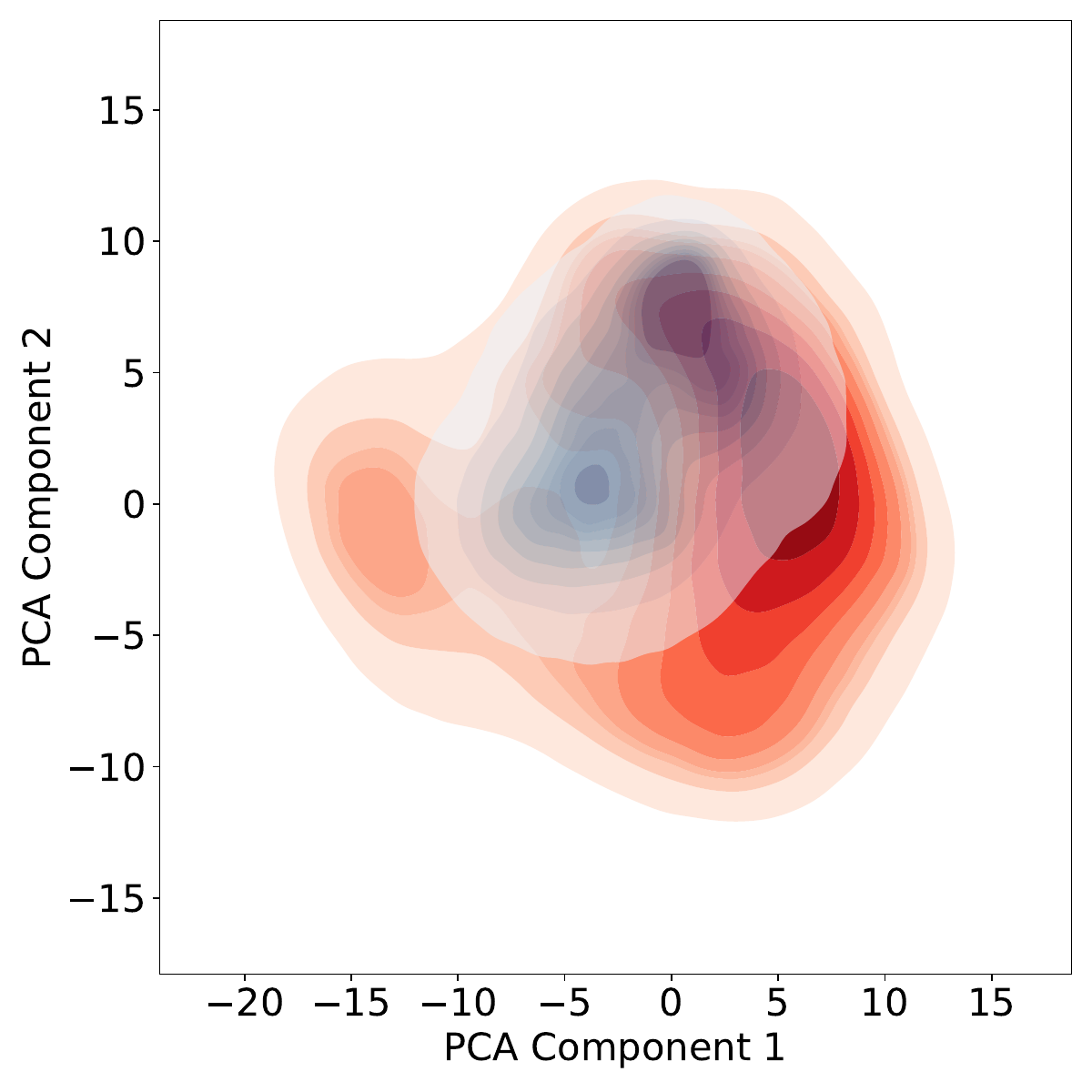}
    \subcaption{\href{https://huggingface.co/datasets/teknium/OpenHermes-2.5}{OpenHermes} VS \href{https://huggingface.co/datasets/HuggingFaceTB/cosmopedia}{Cosmopedia}}
    \label{fig:appendixsub9}
\end{subfigure}%

\end{center}

\caption{\textbf{Visualization of data distribution changes in \ABMN}. The red regions at the bottom denote the pre-training corpus, while the light blue regions above represent the SFT datasets. Darker areas indicate a higher concentration of data points, whereas lighter areas signify sparser distributions.}
\label{fig:appendixprojection}
\end{figure*}

\section{Reservoir sampling algorithm}\label{appendix:algorithm}
Reservoir Sampling is an efficient streaming data sampling method that enables uniform sampling of k items from a data stream without knowing the total size of the stream. It is particularly suited for scenarios with memory constraints or uncertain stream sizes, allowing for equal-probability sampling in a single pass over the data.

\begin{algorithm}[!h]
   \caption{Reservoir Sampling}
   \label{alg:reservoir_sampling}
\begin{algorithmic}
   \STATE {\bfseries Input:} stream of data $x_1, x_2, \dots$, sample size $k$
   \STATE {\bfseries Output:} a random sample of size $k$
   \STATE Initialize an empty reservoir array $R$ of size $k$
   \FOR{$i = 1$ {\bfseries to} $k$}
      \STATE $R[i] \gets x_i$
   \ENDFOR
   \FOR{$i = k+1$ {\bfseries to} $n$}
      \STATE $j \gets \text{random integer from } 1 \text{ to } i$
      \IF{$j \leq k$}
         \STATE $R[j] \gets x_i$
      \ENDIF
   \ENDFOR
   \STATE \textbf{return} $R$
\end{algorithmic}
\end{algorithm}

\section{Prompts for data transformation phase}\label{appendix:prompts}
This section introduces the prompts defined in our data transformation phase, including the question generation prompt, the question evaluation prompt, and the answer generation prompt.
\definecolor{darkorange}{RGB}{255, 140, 0}
\definecolor{darkblue}{RGB}{84, 112, 198}
\definecolor{lightgreen}{RGB}{145, 204, 117}
\definecolor{lightyellow}{RGB}{250, 200, 88}
\definecolor{lightred}{RGB}{238, 102, 102}
\definecolor{lightblue}{RGB}{115, 192, 222}
\newtcolorbox{promptbox}[2][Prompt]{
colback=black!5!white,
arc=5pt, 
boxrule=0.5pt,
fonttitle=\bfseries,
title=#1, 
before upper={\small}, fontupper=\fontfamily{ptm}\selectfont,
colframe=#2, 
}

\begin{promptbox}[Query-to-Questions Generator (Step 1)]{darkblue}
Your task is to generate two questions based on the given text content. Ensure the questions are relevant and directly related to the details provided in the text. Follow these guidelines:
\begin{enumerate}
    \item Question Guidelines:
    \begin{itemize}
        \item Make sure the questions are closely related to the main points or themes mentioned in the text.
        \item Ensure the two questions are as diverse as possible, avoiding homogeneity.
        \item Ensure the questions include all the information needed for the answers. If necessary, add introductory information to the questions.
        \item The questions must be self-contained and should not require the provided text as background to be understood.
    \end{itemize}
\end{enumerate}
Please rewrite the following text into related questions, and output them in JSON format:
Text: \{\}
\\
Output format example:
\begin{verbatim}
{
  "questions": [
    {
      "question": "Generated question content 1"
    },
    {
      "question": "Generated question content 2"
    }
  ]
}
\end{verbatim}
\end{promptbox}

\begin{promptbox}[Query Evaluation Scorer (Step 2)]{darkblue}
Your task is to evaluate the given query based on the following criteria and output the results in JSON format. The output should include three parts: quality, difficulty, and whether additional necessary information is required to answer the query. Please follow the scoring standards below:
\begin{enumerate}
    \item Quality (Score 1-10): Assess the clarity and accuracy of the query. If the query is a simple statement without any question or instruction, score it 1-2.
    \begin{itemize}
        \item 9-10: Very clear, accurate expression, no ambiguity.
        \item 7-8: Clear, accurate expression, but may have minimal ambiguity.
        \item 5-6: Fairly clear, generally accurate expression, but some ambiguity exists.
        \item 3-4: Not very clear, somewhat vague expression, with obvious ambiguity.
        \item 1-2: Unclear, very vague expression, difficult to understand or a simple statement.
    \end{itemize}
    \item Difficulty (Score 1-10): Assess the difficulty of understanding and answering the query.
    \begin{itemize}
        \item 9-10: Very difficult, requires specialized knowledge and complex analysis to answer.
        \item 7-8: Quite difficult, requires some specialized knowledge and analysis.
        \item 5-6: Moderate difficulty, requires general knowledge and analysis.
        \item 3-4: Fairly simple, can be answered with basic knowledge.
        \item 1-2: Very simple, no special knowledge required to answer.
    \end{itemize}
    \item Whether additional necessary information is required to answer: Determine if extra information is needed to fully answer the query.
\end{enumerate}
Please strictly follow the format below for output:
Quality: 1-10\\
Difficulty: 1-10\\
Additional Information Needed: True/False\\
Please evaluate the following query:
Query: \{\}
\\
Output format example:
\begin{verbatim}
{
  "quality": 8,
  "difficulty": 5,
  "additional_info_needed": true
}
\end{verbatim}
\end{promptbox}

\begin{promptbox}[Question-to-Answer Generator (Step 3)]{darkblue}
Your task is to generate an answer based on the given question. Use the background information provided in the text to assist in formulating a relevant and detailed answer. Follow these guidelines:\\
\begin{enumerate}
    \item Question Guidelines:
    \begin{itemize}
      \item Ensure the answer is closely related to the main points or themes mentioned in the question.
      \item Utilize the text content to provide a comprehensive and accurate answer.
      \item Ensure proper formatting and readability, including the correct rendering of any LaTeX or mathematical symbols.
      \item Ensure that the answer provides a complete solution or explanation, with clear and detailed steps.
      \item Use JSON format with the key \texttt{"answer"} for easy extraction and processing.   
    \end{itemize}
\end{enumerate}
Text:\\
\{\}\
\\
Question: \\
\{\}\
\\
Output format example:\\
\begin{verbatim}
{
  "answer": "Generated answer content"
}
\end{verbatim}
\end{promptbox}

\section{Training parameters}\label{appendix:parameters}
Table \ref{tab:training_details} presents the hyperparameters used to train the model in the \ABMN method, which is consistent with those used in the original model's SFT version.
\begin{table}[h]
\centering
\small{
\caption{Hyperparameters in \ABMN.}
\begin{tabular}{lcccccc}
\toprule
\textbf{Base Model}       & Learning Rate & Weight Decay & Warmup Ratio & Batchsize & Epoch & Maximum Sequence Length \\ 
\midrule
 \textbf{OLMo 7B-0724-hf}    & 2e-6     & 0       & 0.03        & 256  & 3  & 4096 \\
 \textbf{Pythia 12b}         & 2e-6     & 0       & 0.03        & 256  & 3  & 4096 \\
 \textbf{Neo 7b}         & 5e-6      & 0       & 0.05        & 512  &    2  & 4096\\

\bottomrule
\end{tabular}
\label{tab:training_details}}
\end{table}

\section{Performances across different ratios}\label{appendix:ratios}

\begin{table*}[!htb]
\caption{\textbf{The results across various ratios}. P-S, I-S, P-L, and I-L denote prompt-level strict
accuracy, instance-level strict accuracy, prompt-level loose accuracy, and instance-level loose accuracy, respectively.}
\label{results:ratio}
\centering
\resizebox{0.95\textwidth}{!}{
    \begin{tabular}{@{}cccccccccccccccc@{}}
    \toprule
        \multicolumn{2}{c}{\multirow{4}{*}{\textbf{\makecell{Experiment \\Setting}}}} &\multicolumn{4}{c}{\textbf{Chat Benchmark}}&\multicolumn{7}{c}{\textbf{Standard Benchmark}} & \multirow{4}{*}{\textbf{Average}}\\ \cmidrule(r){3-6} \cmidrule(r){7-13}
        & & \multicolumn{4}{c}{IFEval} & \multicolumn{3}{c}{Exam} & \multicolumn{2}{c}{Coding} & \multicolumn{2}{c}{Reasoning} \\ \cmidrule(r){3-6} \cmidrule(r){7-9} \cmidrule(r){10-11} \cmidrule(r){12-13}
        & & \makecell{P-S} & \makecell{I-S} & \makecell{P-L} & \makecell{I-L} & MMLU & ARC-c & GPQA-d &\makecell{Human\\Eval} & MBPP & hellaswag & gsm8k \\
    \midrule
        \multicolumn{2}{c}{OLMo-SFT}& 35.30 & 46.52 & 38.63 & 50.24 & 52.93 & 63.73 & 17.68 & 26.83 & 43.92 & 60.35 & 26.84 & 42.09\\
        \multicolumn{2}{c}{0.01}& 36.60 & 49.40 & 37.89 & 51.56 & 55.59 & 71.19 & 26.26 & 32.93 & 46.30 & 66.40 & 29.57 & 45.79\\
        \multicolumn{2}{c}{0.02}& 37.15 & 49.76 & 39.37 & 52.64 & 55.29 & 73.90 & 24.75 & 29.88 & 48.68 & 65.82 & 29.57 & 46.07\\
        \multicolumn{2}{c}{0.05}& 38.63 & 48.92 & 40.30 & 51.44 & 55.71 & 76.61 & 24.75 & 28.05 & 44.18 & 64.97 & 30.86 & 45.86\\
        \multicolumn{2}{c}{0.07}& 38.45 & 50.48 & 39.93 & 52.52 & 54.88 & 71.53 & 18.69 & 26.83 & 45.24 & 64.69 & 31.92 & 45.01\\
        \multicolumn{2}{c}{0.1}& 37.15 & 48.92 & 40.30 & 52.04 & 55.38 & 71.86 & 29.29 & 26.83 & 48.15 & 64.62 & 29.27 & 45.80\\
        \multicolumn{2}{c}{0.2}& 36.23 & 48.44 & 38.26 & 50.24 & 55.29 & 70.85 & 26.26 & 28.05 & 48.15 & 58.61 & 30.55 & 44.63 \\
        \multicolumn{2}{c}{0.3}& 35.49 & 48.08 & 37.52 & 50.00 & 56.04 & 70.51 & 30.30 & 28.05 & 44.97 & 62.97 & 31.46 & 45.04\\
        \multicolumn{2}{c}{0.4}& 36.23 & 48.44 & 39.37 & 50.84 & 55.91 & 73.56 & 29.29 & 31.71 & 42.06 & 63.93 & 30.63 & 45.63 \\
        \multicolumn{2}{c}{0.5}& 35.86 & 47.00 & 38.45 & 49.64 & 55.91 & 72.54 & 26.26 & 29.88 & 46.83 & 62.58 & 29.87 & 44.98\\
        \multicolumn{2}{c}{0.6}& 35.30 & 46.88 & 37.34 & 48.92 & 55.78 & 72.54 & 27.27 & 29.27 & 46.30 & 62.50 & 31.31 & 44.86\\
        \multicolumn{2}{c}{0.7}& 34.20 & 46.76 & 35.86 & 48.56 & 55.49 & 74.24 & 27.27 & 30.49 & 35.00 & 63.77 & 30.55 & 43.84\\  
        
    \bottomrule
    \end{tabular}
}
\end{table*}

\section{Examples} \label{appendix:examples}
Examples 1, 2, and 3 represent three dense regions in the pretraining corpus, corresponding to code, scientific literature, and general text data, respectively. Example 4 represents the dense region of the SFT dataset. Examples 5, 6, and 7 correspond to the three dense regions in the rewritten set. Example 8 indicates points where the SFT data density is higher than that of the pretraining data. Examples 9 and 10 represent points where the pretraining data density exceeds that of the SFT data. Example 10 is as shown in Example 2.

\begin{promptbox}[Example 1 in the original pretraining corpus (density estimation)]{darkblue}
\begin{verbatim}"text":"python
# %load /Users/facai/Study/book_notes/preconfig.py
%matplotlib inline

import matplotlib.pyplot as plt
import seaborn as sns
sns.set(color_codes=True)
#sns.set(font='SimHei')
plt.rcParams['axes.grid'] = False

#from IPython.display import SVG
def show_image(filename, figsize=None, res_dir=True):
    if figsize:
        plt.figure(figsize=figsize)

    if res_dir:
        filename = './res/{}'.format(filename)

    plt.imshow(plt.imread(filename))

Chapter 7 Regularization for Deep Learning
==========================================

the best fitting model is a large model that has been regularized appropriately.

### 7.1 Parameter Norm Penalties

\begin{equation}
    \tilde{J}(\theta; X, y) = J(\theta; X, y) + \alpha \Omega(\theta)
\end{equation}

where $\Omega(\theta)$ is a paramter norm penalty.

typically, penalizes **only the weights** of the affine transformation at each layer
and leaves the biases unregularized.

#### 7.1.1 $L^2$ Parameter Regularization

#### 7.1.2 $L^1$ Regularization

The sparsity property induced by $L^1$ regularization => feature selection

### 7.2 Norm Penalties as Constrained Optimization

constrain $\Omega(\theta)$ to be less than some constant $k$:

\begin{equation}
    \mathcal{L}(\theta, \alpha; X, y) = J(\theta; X, y) + \alpha(\Omega(\theta) - k)
\end{equation}

In practice, column norm limitation is always implemented as an explicit constraint
with reprojection.

### 7.3 Regularization and Under-Constrained Problems

regularized matrix is guarantedd to be invertible."
\end{verbatim}
\end{promptbox}

\begin{promptbox}[Example 2 in the original pretraining corpus (density estimation)]{darkblue}
\begin{verbatim}"text":"\section{\label{intro} Introduction}
The high-precision determination of the machine luminosity at
{\sc lep/slc} is an essential ingredient of the success of
precision tests of the electroweak interactions on top of the $Z$
resonance \cite{review}.
As well known, the Bhabha scattering process at small angle (of the
order of a few degrees) is the reference reaction used for luminosity
monitoring at {\sc lep/slc}, owing to its large cross section (dominated by
$t$-channel photon exchange) and its substantial independence of
purely electroweak effects. Experimental efforts in the development of
efficient, dedicated luminometry detectors, as well as precision
calculations of the small-angle Bhabha (hereafter {\sc sabh}) scattering cross
section both contribute to achieve a measurement of the ``$Z$ factories''
luminosity with a total relative error at the $0.1\%$ level \cite{review,exp,common}.
On the experimental side, the present total uncertainty is smaller than
$0.1\%$ \cite{exp}, close to the $0.05$ level \cite{ward}. As far as the theory
contribution to the luminosity measurement is concerned, the
estimate of the theoretical errors, used by the {\sc lep} collaborations,
is summarized in table \ref{sabs} \cite{common} for centre of mass
energies around and above the $Z$ resonance.
\begin{table}[ht]
\caption[sabs]{\label{sabs}
               Theoretical error in {\sc sabh} scattering according
               to ref.~\cite{common} at typical {\sc lep1} and {\sc lep2}
               energies.}
\medskip
\begin{center}
\begin{tabular}{|l||c|c|} \hline
Type of correction/error & {\sc lep1} ($\%$) & {\sc lep2} ($\%$)\\ \hline \hline
missing photonic $O(\alpha^2L)$   & $0.100 $ & $0.200$ \\
missing photonic $O(\alpha^3L^3)$ & $0.015 $ & $0.030$ \\
vacuum polarization      & $0.040 $ & $0.100$ \\
light pairs              & $0.030 $ & $0.050$ \\
$Z$-exchange             & $0.015 $ & $0.000$ \\ \hline
total                    & $0.110 $ & $0.250$ \\ \hline
\end{tabular}
\end{center}
\end{table}
Some comments on table \ref{sabs} are in order. The components of the theoretical
error refer to the {\sc sabh} scattering cross section, for any typical event
selection of {\sc lep} experiments, as computed by the program {\tt
BHLUMI v4.03} \cite{bhl}.
The largely dominating source of theoretical error 
is due to the missing part of $O(\alpha^2 L)$ subleading photonic corrections,
where $L = \ln(-t/m^2)$ is the collinear logarithm in $t$-channel scattering.
Also the contribution of the missing part of the leading $O(\alpha^3L^3)$
corrections is of photonic nature. The vacuum polarization entry
is the effect of the uncertainty in the hadronic contribution
to the running of $\alpha_{\rm QED}$, when considering the parameterization and
relative error estimate of ref.~\cite{oldvacuum}.
The next contribution is the uncertainty introduced by the corrections due
to the production of light pairs, chiefly $e^+ e^-$ ones. The last
entry refers to the uncertainty associated to the treatment of the
$\gamma$-$Z$ interference.
More details about the strategy adopted in order to estimate the various 
sources of theoretical error can be found in ref.~\cite{common}.
After the analysis of ref. \cite{common}, important theoretical
developments took place. Additional work in the sector of two-loop
photonic corrections \cite{pv,kr} led to the conclusion that the
perturbative contribution due to the uncontrolled part of $O(\alpha^2L)$
corrections does not exceed the $0.03\%$ level. 
..."
\end{verbatim}
\end{promptbox}

\begin{promptbox}[Example 3 in the original pretraining corpus (density estimation)]{darkblue}
\begin{verbatim}"text":"# A marble dropped from a bridge strikes the water in 5.0 s. 
what is the height of the bridge? (Answer in meters) Jun 6, 2018. $\\text{122.5 m}$.
#### Explanation:
Use equation of motion: ${\\text{S" = "ut" + 1/2"at}}^{2}$
Where，$\\text{S =}$ Displacement covered，$\\text{u =}$ Initial velocity
$\\text{a =}$ Acceleration (It’s ${\\text{9.8 m/s}}^{2}$ due to Earths gravity)
$\\text{t =}$ Time taken to cover displacement $\\text{S}$
Marble is dropped. It means initial velocity ($\\text{u}$) of marble is $0$
Therefore the equation becomes ${\\text{S" = 1/2"at}}^{2}$
color(white)("S") = 1/2 × "9.8 m/s"^2 × "(5 s)"^2
color(white)("S") = 122.5\\ "m"
Displacement covered by the marble when thrown from the bridge is 
obviously height of the bridge.
\end{verbatim}
\end{promptbox}

\begin{promptbox}[Example 4 in the original SFT dataset (density estimation)]{darkblue}
\begin{verbatim}"query":"Tweet: enjoyed watching Glee, it brought back some good memories of being 
in musical theatre, and snowflake
Sentiment: positive

Tweet: @dandebuf i'm pretty sure he's every main character we've ever written. 
that or @iamnotmatt 
Sentiment: positive

Tweet: Still no iPhone update 
Sentiment: negative

Tweet: @jimblum must have been quite a ride to loose 44 lbs 
Sentiment:"

"answer":"positive"
\end{verbatim}
\end{promptbox}

\begin{promptbox}[Example 5 in the rewritten dataset (density estimation)]{darkblue}
\begin{verbatim}"query":"What is the effect of L1 regularization on the model parameters, and how 
does it contribute to feature selection within a deep learning model?"

"answer":"L1 regularization, also known as Lasso regularization, adds a penalty to 
the model based on the absolute value of the model's coefficients. Mathematically, 
this can be represented as an addition to the loss function: 

\\(J(\\theta; X, y) + \\alpha \\sum_{i} |\\theta_i|\\), where \\(\\theta_i\\) 
are the model parameters and \\(\\alpha\\) is the regularization strength. 

The key effect of L1 regularization is that it encourages the model to set some of the 
less important feature coefficients to zero. This property is particularly useful for
feature selection because it effectively removes irrelevant or less important features
from the model, leading to a more interpretable and potentially more robust model. 

By pruning out these unimportant features, L1 regularization helps in simplifying the 
model, reducing overfitting, and improving computational efficiency, especially in
high-dimensional datasets where many features might be redundant or irrelevant."
\end{verbatim}
\end{promptbox}

\begin{promptbox}[Example 6 in the rewritten dataset (density estimation)]{darkblue}
\begin{verbatim}"query":"What are the constituent elements of carbon dioxide and how many atoms of 
each are present in a single molecule?,"

"answer":"The constituent elements of carbon dioxide are carbon and oxygen.
In a single molecule of carbon dioxide (CO2), there is one atom of carbon 
and two atoms of oxygen."
\end{verbatim}
\end{promptbox}

\begin{promptbox}[Example 7 in the rewritten dataset (density estimation)]{darkblue}
\begin{verbatim}"query":"How is the signless Laplacian matrix of a graph G defined, and what is the 
relationship between the signless Laplacian spectrum of G and the spectrum of the 
line graph of G?"

"answer":"The signless Laplacian matrix of a graph \( G \) is defined as 
\( Q_G = D_G + A_G \), where \( D_G \) is the diagonal matrix of vertex degrees 
and \( A_G \) is the adjacency matrix of the graph. 

The signless Laplacian spectrum of \( G \) is the multiset of eigenvalues of \( Q_G \). 
The relationship between the signless Laplacian spectrum of \( G \) and
the spectrum of the line graph of \( G \), denoted \( \mathcal{L}_G \), 
is given by the following: 

for an \( (n, m) \)-graph \( G \), the eigenvalues of the signless Laplacian matrix 
and the line graph are related as \( q_i(G) = 2 + \lambda_i(\mathcal{L}_G) \) 
for \( i = 1, 2, \ldots, k \), where \( k = \min\{m, n\} \) and
\( \lambda_i(\mathcal{L}_G) \) is the \( i \)-th largest eigenvalue of the line graph. 
Moreover, if \( m > n \), then \( \lambda_i(\mathcal{L}_G) = -2 \) for \( m \geq i 
\geq n+1 \), and if \( n > m \), then \( q_i = 0 \) for \( n \geq i \geq m+1 \)."
\end{verbatim}
\end{promptbox}

\begin{promptbox}[Example 8 in the original SFT dataset (density estimation)]{darkblue}
\begin{verbatim}"query":"Ques: Given this review: Great app. Fun and cool graphics.
Would you recommend this app to a friend? Not at all, No, Maybe, Yes, or Definitely?
Ans: Definitely

Ques: Given this review: In Zenfone 2 ram 2gb intel processor not very well after
I upgrade to marshmallow..i dont know why..very very lag..before in lollipop version
this psp game works really well..i use default settings no problem..
Would you recommend this app to a friend? Not at all, No, Maybe, Yes, or Definitely?
Ans: No

Ques: Given this review: Did not work at all To much errors
Would you recommend this app to a friend? Not at all, No, Maybe, Yes, or Definitely?
Ans: Not at all

Ques: Given this review: I like this app Open anyyy apps & anyone like this app
Would you recommend this app to a friend? Not at all, No, Maybe, Yes, or Definitely?
Ans:"

"answer":"Definitely"
\end{verbatim}
\end{promptbox}

\begin{promptbox}[Example 9 in the original pretraining corpus (density estimation)]{darkblue}
\begin{verbatim}"text":"# If the length of a 46 cm spring increases to 57 cm when a 8 kg 
weight is hanging from it, what is the spring\'s constant?

$\\text{713 N/m}$\n$F = k \\Delta x$\nk = F/(Deltax) = (mg)/(Deltax) = 
("8 kg × 9.8 m/s"^2)/("0.57 m - 0.46 m") = "713 N/m""
\end{verbatim}
\end{promptbox}

\end{document}

%% file: intro_v2.tex
\section{Introduction}
Instruction tuning is essential for adapting large language models (LLMs) to effectively follow human instructions across diverse tasks. This process relies on high-quality datasets to guide model behavior, yet existing instruction-tuning datasets are often narrowly focused, relying on either manual annotation or synthetic generation. While manual datasets offer precision, they are costly and lack scalability \cite{supernaturalinstructions, zhou2023lima}. Synthetic datasets, on the other hand, frequently depend on expensive APIs of strong models and are tightly coupled with their generation pipelines, limiting flexibility \cite{peng2023instruction, OpenOrca}. Additionally, manually combining open-source datasets, as seen in efforts like OpenHermes-2.5~\cite{OpenHermes25} and Tulu-V2 \cite{ivison2023camels}, often overlooks the underlying data distributions, leading to inefficiencies.

\begin{figure}[t]
\begin{center}
\includegraphics[width=0.495\columnwidth]{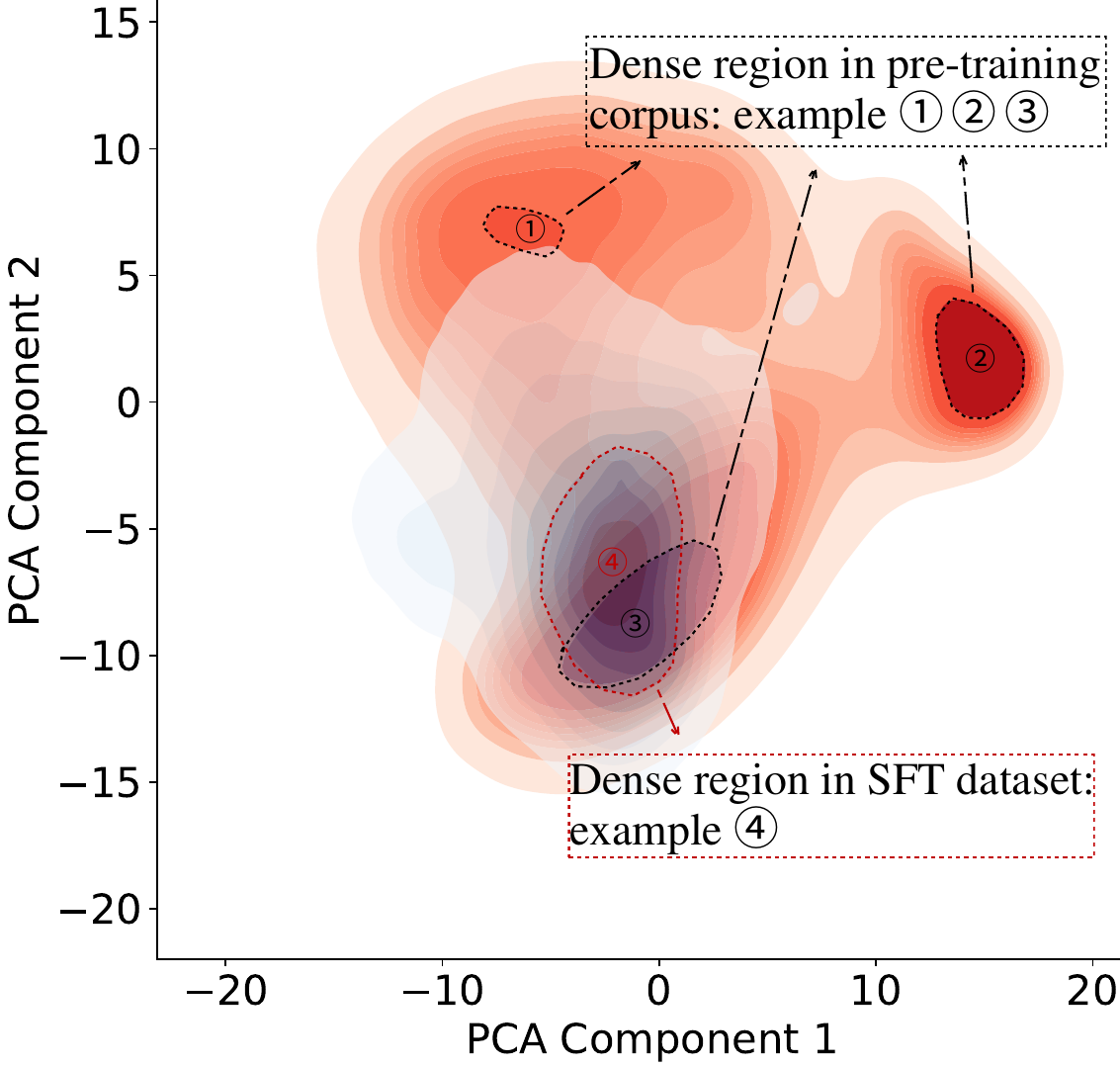}
\includegraphics[width=0.49\columnwidth]{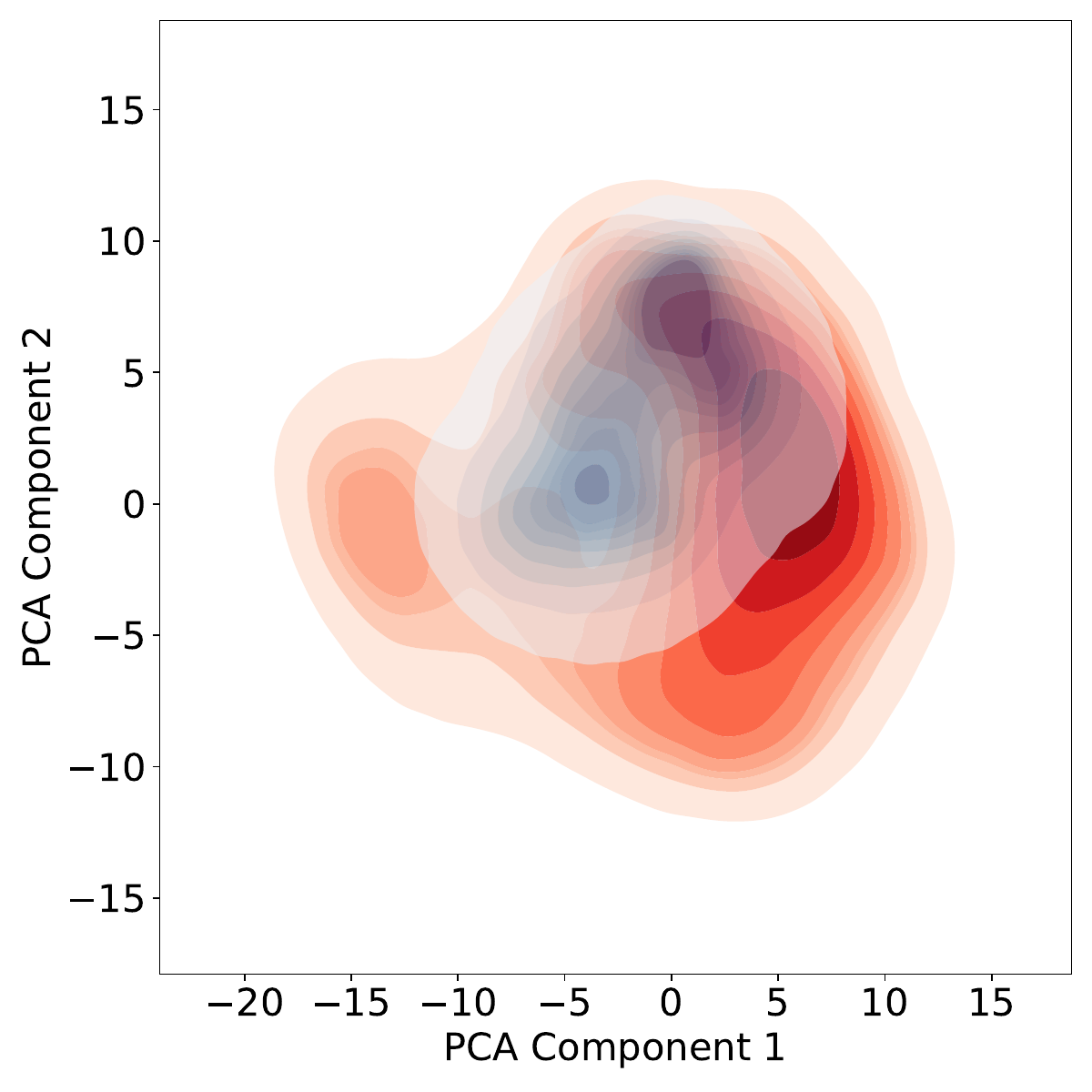}
\end{center}
\caption{\textbf{Visualization of Projections}. The red regions at the bottom represent the pre-training corpus, while the light blue regions above represent the SFT datasets. Darker areas indicate a higher concentration of data points, whereas lighter areas represent sparser distributions. Additional projections are shown in \autoref{appendix:visualization}.}
\label{fig:projection}
\vspace{-0.4cm}
\end{figure}

\begin{figure*}[ht]
\begin{center}
\includegraphics[width=\linewidth]{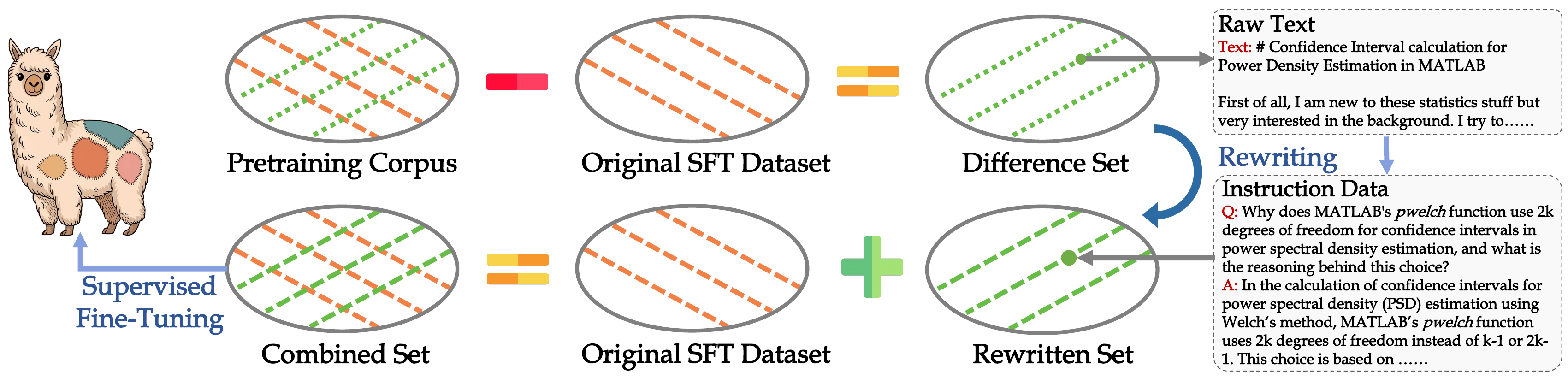}
\end{center}
\caption{\textbf{The pipeline of \ABMN.} \ABMN first generates a difference set, then rewrites the raw text into instruction-response pairs to form a rewritten set, and finally combines the rewritten set with the original SFT dataset for model training.}
\label{fig:pipeline}
\end{figure*}

Pre-training corpora, by contrast, reflect broader real-world distributions and align closely with the internal knowledge of LLMs, making them a rich source of high-quality supervisory signals. However, current instruction-tuning methods fail to leverage this alignment, creating a fundamental gap in optimizing dataset coverage and distribution. Addressing this challenge requires aligning instruction-tuning datasets with pre-training distributions to fully exploit the knowledge embedded in LLMs.

In this paper, we propose Aligning Instruction Tuning with Pre-training (\ABMN), a method that systematically bridges this gap. Rather than generating instruction-response pairs from scratch, \ABMN identifies gaps in existing datasets by comparing their distribution to that of the pre-training corpus. Underrepresented data is then rewritten into high-quality instruction-response pairs, enhancing dataset coverage and alignment. As shown in Figure \ref{fig:pipeline}, \ABMN involves three stages: (1) generating a difference set based on density comparisons, (2) rewriting raw text into instruction-response pairs, and (3) integrating these pairs into the original dataset for fine-tuning.

Figure \ref{fig:projection} visualizes the significant distributional differences between instruction-tuning datasets and the pre-training corpus, underscoring the need for such alignment. Through experiments on three open-source LLMs across eight benchmarks, we demonstrate that \ABMN consistently improves model performance. Detailed ablation studies highlight the effectiveness of adaptive data selection and integration, showing how \ABMN guides instruction tuning toward more effective and generalizable fine-tuned models.

Our contributions include: 1) Demonstrating the distributional gaps between instruction-tuning datasets and pre-training corpora through visualization. 2) Proposing the \ABMN method to adaptively optimize instruction-tuning datasets by leveraging pre-training corpora as a reference. 3) Validating the effectiveness of \ABMN with extensive experiments and ablation studies.

%% file: example_paper.bbl
\begin{thebibliography}{69}
\providecommand{\natexlab}[1]{#1}
\providecommand{\url}[1]{\texttt{#1}}
\expandafter\ifx\csname urlstyle\endcsname\relax
  \providecommand{\doi}[1]{doi: #1}\else
  \providecommand{\doi}{doi: \begingroup \urlstyle{rm}\Url}\fi

\bibitem[AI et~al.(2024)AI, :, Young, Chen, Li, Huang, Zhang, Zhang, Li, Zhu, Chen, Chang, Yu, Liu, Liu, Yue, Yang, Yang, Yu, Xie, Huang, Hu, Ren, Niu, Nie, Xu, Liu, Wang, Cai, Gu, Liu, and Dai]{ai2024yi}
AI, ., :, Young, A., Chen, B., Li, C., Huang, C., Zhang, G., Zhang, G., Li, H., Zhu, J., Chen, J., Chang, J., Yu, K., Liu, P., Liu, Q., Yue, S., Yang, S., Yang, S., Yu, T., Xie, W., Huang, W., Hu, X., Ren, X., Niu, X., Nie, P., Xu, Y., Liu, Y., Wang, Y., Cai, Y., Gu, Z., Liu, Z., and Dai, Z.
\newblock Yi: Open foundation models by 01.ai.
\newblock \emph{arXiv preprint arXiv: 2403.04652}, 2024.

\bibitem[Alemohammad et~al.(2023)Alemohammad, Casco-Rodriguez, Luzi, Humayun, Babaei, LeJeune, Siahkoohi, and Baraniuk]{alemohammad2023selfconsuming}
Alemohammad, S., Casco-Rodriguez, J., Luzi, L., Humayun, A.~I., Babaei, H., LeJeune, D., Siahkoohi, A., and Baraniuk, R.~G.
\newblock Self-consuming generative models go mad.
\newblock \emph{arXiv preprint arXiv: 2307.01850}, 2023.

\bibitem[Anthropic(2024)]{anthropic2024claude3}
Anthropic.
\newblock Claude 3 haiku: Our fastest model yet, 2024.
\newblock Available at: \url{https://www.anthropic.com/news/claude-3-haiku}.

\bibitem[Asai et~al.(2023)Asai, Wu, Wang, Sil, and Hajishirzi]{asai2023selfrag}
Asai, A., Wu, Z., Wang, Y., Sil, A., and Hajishirzi, H.
\newblock Self-rag: Learning to retrieve, generate, and critique through self-reflection.
\newblock \emph{arXiv preprint arXiv: 2310.11511}, 2023.

\bibitem[Austin et~al.(2021)Austin, Odena, Nye, Bosma, Michalewski, Dohan, Jiang, Cai, Terry, Le, and Sutton]{austin2021program}
Austin, J., Odena, A., Nye, M., Bosma, M., Michalewski, H., Dohan, D., Jiang, E., Cai, C., Terry, M., Le, Q., and Sutton, C.
\newblock Program synthesis with large language models.
\newblock \emph{arXiv preprint arXiv: 2108.07732}, 2021.

\bibitem[Biderman et~al.(2023)Biderman, Schoelkopf, Anthony, Bradley, O'Brien, Hallahan, Khan, Purohit, Prashanth, Raff, Skowron, Sutawika, and van~der Wal]{biderman2023pythia}
Biderman, S., Schoelkopf, H., Anthony, Q., Bradley, H., O'Brien, K., Hallahan, E., Khan, M.~A., Purohit, S., Prashanth, U.~S., Raff, E., Skowron, A., Sutawika, L., and van~der Wal, O.
\newblock Pythia: A suite for analyzing large language models across training and scaling.
\newblock \emph{arXiv preprint arXiv: 2304.01373}, 2023.

\bibitem[Chen et~al.(2024)Chen, Xiao, Zhang, Luo, Lian, and Liu]{bge-m3}
Chen, J., Xiao, S., Zhang, P., Luo, K., Lian, D., and Liu, Z.
\newblock Bge m3-embedding: Multi-lingual, multi-functionality, multi-granularity text embeddings through self-knowledge distillation, 2024.

\bibitem[Chen et~al.(2021)Chen, Tworek, Jun, Yuan, de~Oliveira~Pinto, Kaplan, Edwards, Burda, Joseph, Brockman, Ray, Puri, Krueger, Petrov, Khlaaf, Sastry, Mishkin, Chan, Gray, Ryder, Pavlov, Power, Kaiser, Bavarian, Winter, Tillet, Such, Cummings, Plappert, Chantzis, Barnes, Herbert-Voss, Guss, Nichol, Paino, Tezak, Tang, Babuschkin, Balaji, Jain, Saunders, Hesse, Carr, Leike, Achiam, Misra, Morikawa, Radford, Knight, Brundage, Murati, Mayer, Welinder, McGrew, Amodei, McCandlish, Sutskever, and Zaremba]{chen2021evaluating}
Chen, M., Tworek, J., Jun, H., Yuan, Q., de~Oliveira~Pinto, H.~P., Kaplan, J., Edwards, H., Burda, Y., Joseph, N., Brockman, G., Ray, A., Puri, R., Krueger, G., Petrov, M., Khlaaf, H., Sastry, G., Mishkin, P., Chan, B., Gray, S., Ryder, N., Pavlov, M., Power, A., Kaiser, L., Bavarian, M., Winter, C., Tillet, P., Such, F.~P., Cummings, D., Plappert, M., Chantzis, F., Barnes, E., Herbert-Voss, A., Guss, W.~H., Nichol, A., Paino, A., Tezak, N., Tang, J., Babuschkin, I., Balaji, S., Jain, S., Saunders, W., Hesse, C., Carr, A.~N., Leike, J., Achiam, J., Misra, V., Morikawa, E., Radford, A., Knight, M., Brundage, M., Murati, M., Mayer, K., Welinder, P., McGrew, B., Amodei, D., McCandlish, S., Sutskever, I., and Zaremba, W.
\newblock Evaluating large language models trained on code.
\newblock \emph{arXiv preprint arXiv: 2107.03374}, 2021.

\bibitem[Chung et~al.(2024)Chung, Hou, Longpre, Zoph, Tay, Fedus, Li, Wang, Dehghani, Brahma, et~al.]{chung2024scaling}
Chung, H.~W., Hou, L., Longpre, S., Zoph, B., Tay, Y., Fedus, W., Li, Y., Wang, X., Dehghani, M., Brahma, S., et~al.
\newblock Scaling instruction-finetuned language models.
\newblock \emph{Journal of Machine Learning Research}, 25\penalty0 (70):\penalty0 1--53, 2024.

\bibitem[Chung et~al.(2023)Chung, Kamar, and Amershi]{chung2023increasing}
Chung, J. J.~Y., Kamar, E., and Amershi, S.
\newblock Increasing diversity while maintaining accuracy: Text data generation with large language models and human interventions.
\newblock \emph{arXiv preprint arXiv: 2306.04140}, 2023.

\bibitem[Clark et~al.(2018)Clark, Cowhey, Etzioni, Khot, Sabharwal, Schoenick, and Tafjord]{clark2018think}
Clark, P., Cowhey, I., Etzioni, O., Khot, T., Sabharwal, A., Schoenick, C., and Tafjord, O.
\newblock Think you have solved question answering? try arc, the ai2 reasoning challenge.
\newblock \emph{arXiv preprint arXiv: 1803.05457}, 2018.

\bibitem[Cobbe et~al.(2021)Cobbe, Kosaraju, Bavarian, Chen, Jun, Kaiser, Plappert, Tworek, Hilton, Nakano, Hesse, and Schulman]{cobbe2021training}
Cobbe, K., Kosaraju, V., Bavarian, M., Chen, M., Jun, H., Kaiser, L., Plappert, M., Tworek, J., Hilton, J., Nakano, R., Hesse, C., and Schulman, J.
\newblock Training verifiers to solve math word problems.
\newblock \emph{arXiv preprint arXiv: 2110.14168}, 2021.

\bibitem[Contributors(2023)]{2023opencompass}
Contributors, O.
\newblock Opencompass: A universal evaluation platform for foundation models.
\newblock \url{https://github.com/open-compass/opencompass}, 2023.

\bibitem[DeepSeek-AI et~al.(2024)DeepSeek-AI, :, Bi, Chen, Chen, Chen, Dai, Deng, Ding, Dong, Du, Fu, Gao, Gao, Gao, Ge, Guan, Guo, Guo, Hao, Hao, He, Hu, Huang, Li, Li, Li, Li, Li, Liang, Lin, Liu, Liu, Liu, Liu, Liu, Liu, Lu, Lu, Luo, Ma, Nie, Pei, Piao, Qiu, Qu, Ren, Ren, Ruan, Sha, Shao, Song, Su, Sun, Sun, Tang, Wang, Wang, Wang, Wang, Wang, Wu, Wu, Xie, Xie, Xie, Xiong, Xu, Xu, Xu, Yang, You, Yu, Yu, Zhang, Zhang, Zhang, Zhang, Zhang, Zhang, Zhang, Zhang, Zhao, Zhao, Zhou, Zhou, Zhu, and Zou]{deepseek-ai2024deepseek}
DeepSeek-AI, :, Bi, X., Chen, D., Chen, G., Chen, S., Dai, D., Deng, C., Ding, H., Dong, K., Du, Q., Fu, Z., Gao, H., Gao, K., Gao, W., Ge, R., Guan, K., Guo, D., Guo, J., Hao, G., Hao, Z., He, Y., Hu, W., Huang, P., Li, E., Li, G., Li, J., Li, Y., Li, Y.~K., Liang, W., Lin, F., Liu, A.~X., Liu, B., Liu, W., Liu, X., Liu, X., Liu, Y., Lu, H., Lu, S., Luo, F., Ma, S., Nie, X., Pei, T., Piao, Y., Qiu, J., Qu, H., Ren, T., Ren, Z., Ruan, C., Sha, Z., Shao, Z., Song, J., Su, X., Sun, J., Sun, Y., Tang, M., Wang, B., Wang, P., Wang, S., Wang, Y., Wang, Y., Wu, T., Wu, Y., Xie, X., Xie, Z., Xie, Z., Xiong, Y., Xu, H., Xu, R.~X., Xu, Y., Yang, D., You, Y., Yu, S., Yu, X., Zhang, B., Zhang, H., Zhang, L., Zhang, L., Zhang, M., Zhang, M., Zhang, W., Zhang, Y., Zhao, C., Zhao, Y., Zhou, S., Zhou, S., Zhu, Q., and Zou, Y.
\newblock Deepseek llm: Scaling open-source language models with longtermism.
\newblock \emph{arXiv preprint arXiv: 2401.02954}, 2024.

\bibitem[Gao et~al.(2024)Gao, Niu, He, Xu, Liu, Liu, Hu, and Wen]{gao2024interpretable}
Gao, Z., Niu, B., He, X., Xu, H., Liu, H., Liu, A., Hu, X., and Wen, L.
\newblock Interpretable contrastive monte carlo tree search reasoning.
\newblock \emph{arXiv preprint arXiv: 2410.01707}, 2024.

\bibitem[Ge et~al.(2024)Ge, Chan, Wang, Yu, Mi, and Yu]{ge2024scaling}
Ge, T., Chan, X., Wang, X., Yu, D., Mi, H., and Yu, D.
\newblock Scaling synthetic data creation with 1,000,000,000 personas.
\newblock \emph{arXiv preprint arXiv: 2406.20094}, 2024.

\bibitem[GLM et~al.(2024)GLM, :, Zeng, Xu, Wang, Zhang, Yin, Zhang, Rojas, Feng, Zhao, Lai, Yu, Wang, Sun, Zhang, Cheng, Gui, Tang, Zhang, Sun, Li, Zhao, Wu, Zhong, Liu, Huang, Zhang, Zheng, Lu, Duan, Zhang, Cao, Yang, Tam, Zhao, Liu, Xia, Zhang, Gu, Lv, Liu, Liu, Yang, Song, Zhang, An, Xu, Niu, Yang, Li, Bai, Dong, Qi, Wang, Yang, Du, Hou, and Wang]{glm2024chatglm}
GLM, T., :, Zeng, A., Xu, B., Wang, B., Zhang, C., Yin, D., Zhang, D., Rojas, D., Feng, G., Zhao, H., Lai, H., Yu, H., Wang, H., Sun, J., Zhang, J., Cheng, J., Gui, J., Tang, J., Zhang, J., Sun, J., Li, J., Zhao, L., Wu, L., Zhong, L., Liu, M., Huang, M., Zhang, P., Zheng, Q., Lu, R., Duan, S., Zhang, S., Cao, S., Yang, S., Tam, W.~L., Zhao, W., Liu, X., Xia, X., Zhang, X., Gu, X., Lv, X., Liu, X., Liu, X., Yang, X., Song, X., Zhang, X., An, Y., Xu, Y., Niu, Y., Yang, Y., Li, Y., Bai, Y., Dong, Y., Qi, Z., Wang, Z., Yang, Z., Du, Z., Hou, Z., and Wang, Z.
\newblock Chatglm: A family of large language models from glm-130b to glm-4 all tools.
\newblock \emph{arXiv preprint arXiv: 2406.12793}, 2024.

\bibitem[Groeneveld et~al.(2024)Groeneveld, Beltagy, Walsh, Bhagia, Kinney, Tafjord, Jha, Ivison, Magnusson, Wang, Arora, Atkinson, Authur, Chandu, Cohan, Dumas, Elazar, Gu, Hessel, Khot, Merrill, Morrison, Muennighoff, Naik, Nam, Peters, Pyatkin, Ravichander, Schwenk, Shah, Smith, Strubell, Subramani, Wortsman, Dasigi, Lambert, Richardson, Zettlemoyer, Dodge, Lo, Soldaini, Smith, and Hajishirzi]{groeneveld2024olmo}
Groeneveld, D., Beltagy, I., Walsh, P., Bhagia, A., Kinney, R., Tafjord, O., Jha, A., Ivison, H., Magnusson, I., Wang, Y., Arora, S., Atkinson, D., Authur, R., Chandu, K.~R., Cohan, A., Dumas, J., Elazar, Y., Gu, Y., Hessel, J., Khot, T., Merrill, W., Morrison, J.~D., Muennighoff, N., Naik, A., Nam, C., Peters, M.~E., Pyatkin, V., Ravichander, A., Schwenk, D., Shah, S., Smith, W., Strubell, E., Subramani, N., Wortsman, M., Dasigi, P., Lambert, N., Richardson, K., Zettlemoyer, L.~S., Dodge, J., Lo, K., Soldaini, L., Smith, N.~A., and Hajishirzi, H.
\newblock Olmo: Accelerating the science of language models.
\newblock \emph{Annual Meeting of the Association for Computational Linguistics}, 2024.
\newblock \doi{10.48550/arXiv.2402.00838}.

\bibitem[Guo et~al.(2024)Guo, Shang, Vazirgiannis, and Clavel]{guo-etal-2024-curious}
Guo, Y., Shang, G., Vazirgiannis, M., and Clavel, C.
\newblock The curious decline of linguistic diversity: Training language models on synthetic text.
\newblock In Duh, K., Gomez, H., and Bethard, S. (eds.), \emph{Findings of the Association for Computational Linguistics: NAACL 2024}, pp.\  3589--3604, Mexico City, Mexico, June 2024. Association for Computational Linguistics.
\newblock \doi{10.18653/v1/2024.findings-naacl.228}.
\newblock URL \url{https://aclanthology.org/2024.findings-naacl.228}.

\bibitem[Hendrycks et~al.(2021)Hendrycks, Burns, Basart, Zou, Mazeika, Song, and Steinhardt]{hendryckstest2021}
Hendrycks, D., Burns, C., Basart, S., Zou, A., Mazeika, M., Song, D., and Steinhardt, J.
\newblock Measuring massive multitask language understanding.
\newblock \emph{Proceedings of the International Conference on Learning Representations (ICLR)}, 2021.

\bibitem[Huang et~al.(2023)Huang, Gu, Hou, Wu, Wang, Yu, and Han]{huang-etal-2023-large}
Huang, J., Gu, S., Hou, L., Wu, Y., Wang, X., Yu, H., and Han, J.
\newblock Large language models can self-improve.
\newblock In Bouamor, H., Pino, J., and Bali, K. (eds.), \emph{Proceedings of the 2023 Conference on Empirical Methods in Natural Language Processing}, pp.\  1051--1068, Singapore, December 2023. Association for Computational Linguistics.
\newblock \doi{10.18653/v1/2023.emnlp-main.67}.
\newblock URL \url{https://aclanthology.org/2023.emnlp-main.67}.

\bibitem[Ivison et~al.(2023)Ivison, Wang, Pyatkin, Lambert, Peters, Dasigi, Jang, Wadden, Smith, Beltagy, and Hajishirzi]{ivison2023camels}
Ivison, H., Wang, Y., Pyatkin, V., Lambert, N., Peters, M., Dasigi, P., Jang, J., Wadden, D., Smith, N.~A., Beltagy, I., and Hajishirzi, H.
\newblock Camels in a changing climate: Enhancing lm adaptation with tulu 2.
\newblock \emph{arXiv preprint arXiv: 2311.10702}, 2023.

\bibitem[Jiang et~al.(2024{\natexlab{a}})Jiang, Sablayrolles, Roux, Mensch, Savary, Bamford, Chaplot, de~las Casas, Hanna, Bressand, Lengyel, Bour, Lample, Lavaud, Saulnier, Lachaux, Stock, Subramanian, Yang, Antoniak, Scao, Gervet, Lavril, Wang, Lacroix, and Sayed]{jiang2024mixtral}
Jiang, A.~Q., Sablayrolles, A., Roux, A., Mensch, A., Savary, B., Bamford, C., Chaplot, D.~S., de~las Casas, D., Hanna, E.~B., Bressand, F., Lengyel, G., Bour, G., Lample, G., Lavaud, L.~R., Saulnier, L., Lachaux, M.-A., Stock, P., Subramanian, S., Yang, S., Antoniak, S., Scao, T.~L., Gervet, T., Lavril, T., Wang, T., Lacroix, T., and Sayed, W.~E.
\newblock Mixtral of experts.
\newblock \emph{arXiv preprint arXiv: 2401.04088}, 2024{\natexlab{a}}.

\bibitem[Jiang et~al.(2024{\natexlab{b}})Jiang, min Chan, Xue, Liu, and Guo]{jiang2024importance}
Jiang, C., min Chan, C., Xue, W., Liu, Q., and Guo, Y.
\newblock Importance weighting can help large language models self-improve.
\newblock \emph{arXiv preprint arXiv: 2408.09849}, 2024{\natexlab{b}}.

\bibitem[Jones et~al.(2023)Jones, Palangi, Simões, Chandrasekaran, Mukherjee, Mitra, Awadallah, and Kamar]{jones2023teaching}
Jones, E., Palangi, H., Simões, C., Chandrasekaran, V., Mukherjee, S., Mitra, A., Awadallah, A., and Kamar, E.
\newblock Teaching language models to hallucinate less with synthetic tasks.
\newblock \emph{arXiv preprint arXiv: 2310.06827}, 2023.

\bibitem[Kang et~al.(2024)Kang, Luo, Zhu, Hansen, Glass, Cox, Ritter, Feris, and Karlinsky]{kang-etal-2024-self}
Kang, J., Luo, H., Zhu, Y., Hansen, J., Glass, J., Cox, D., Ritter, A., Feris, R., and Karlinsky, L.
\newblock Self-specialization: Uncovering latent expertise within large language models.
\newblock In Ku, L.-W., Martins, A., and Srikumar, V. (eds.), \emph{Findings of the Association for Computational Linguistics: ACL 2024}, pp.\  2681--2706, Bangkok, Thailand, aug 2024. Association for Computational Linguistics.
\newblock \doi{10.18653/v1/2024.findings-acl.157}.
\newblock URL \url{https://aclanthology.org/2024.findings-acl.157}.

\bibitem[Khashabi et~al.(2020)Khashabi, Min, Khot, Sabharwal, Tafjord, Clark, and Hajishirzi]{khashabi2020unifiedqa}
Khashabi, D., Min, S., Khot, T., Sabharwal, A., Tafjord, O., Clark, P., and Hajishirzi, H.
\newblock Unifiedqa: Crossing format boundaries with a single qa system.
\newblock \emph{arXiv preprint arXiv: 2005.00700}, 2020.

\bibitem[Kwon et~al.(2023)Kwon, Li, Zhuang, Sheng, Zheng, Yu, Gonzalez, Zhang, and Stoica]{kwon2023efficient}
Kwon, W., Li, Z., Zhuang, S., Sheng, Y., Zheng, L., Yu, C.~H., Gonzalez, J.~E., Zhang, H., and Stoica, I.
\newblock Efficient memory management for large language model serving with pagedattention.
\newblock In \emph{Proceedings of the ACM SIGOPS 29th Symposium on Operating Systems Principles}, 2023.

\bibitem[Köksal et~al.(2023)Köksal, Schick, Korhonen, and Schütze]{kksal2023longform}
Köksal, A., Schick, T., Korhonen, A., and Schütze, H.
\newblock Longform: Effective instruction tuning with reverse instructions.
\newblock \emph{Conference on Empirical Methods in Natural Language Processing}, 2023.
\newblock \doi{10.18653/v1/2024.findings-emnlp.414}.

\bibitem[Lee et~al.(2024)Lee, Wattanawong, Kim, Mangalam, Shen, Anumanchipalli, Mahoney, Keutzer, and Gholami]{lee2024llm2llm}
Lee, N., Wattanawong, T., Kim, S., Mangalam, K., Shen, S., Anumanchipalli, G., Mahoney, M.~W., Keutzer, K., and Gholami, A.
\newblock Llm2llm: Boosting llms with novel iterative data enhancement.
\newblock \emph{arXiv preprint arXiv: 2403.15042}, 2024.

\bibitem[Li et~al.(2024)Li, Chen, Chen, He, Gu, and Zhou]{DBLP:conf/acl/LiCCHGZ24}
Li, M., Chen, L., Chen, J., He, S., Gu, J., and Zhou, T.
\newblock Selective reflection-tuning: Student-selected data recycling for {LLM} instruction-tuning.
\newblock In Ku, L., Martins, A., and Srikumar, V. (eds.), \emph{Findings of the Association for Computational Linguistics, {ACL} 2024, Bangkok, Thailand and virtual meeting, August 11-16, 2024}, pp.\  16189--16211. Association for Computational Linguistics, 2024.
\newblock \doi{10.18653/V1/2024.FINDINGS-ACL.958}.
\newblock URL \url{https://doi.org/10.18653/v1/2024.findings-acl.958}.

\bibitem[Li et~al.(2023)Li, Yu, Zhou, Schick, Levy, Zettlemoyer, Weston, and Lewis]{li2023selfalignment}
Li, X., Yu, P., Zhou, C., Schick, T., Levy, O., Zettlemoyer, L., Weston, J., and Lewis, M.
\newblock Self-alignment with instruction backtranslation.
\newblock \emph{arXiv preprint arXiv: 2308.06259}, 2023.

\bibitem[Lian et~al.(2023)Lian, Goodson, Pentland, Cook, Vong, and "Teknium"]{OpenOrca}
Lian, W., Goodson, B., Pentland, E., Cook, A., Vong, C., and "Teknium".
\newblock Openorca: An open dataset of gpt augmented flan reasoning traces.
\newblock \url{https://https://huggingface.co/Open-Orca/OpenOrca}, 2023.

\bibitem[Liang et~al.(2024)Liang, Zhang, Qu, Zheng, Guo, Du, Yang, Liu, Lin, Ma, Huang, and Zhang]{liang2024isheep}
Liang, Y., Zhang, G., Qu, X., Zheng, T., Guo, J., Du, X., Yang, Z., Liu, J., Lin, C., Ma, L., Huang, W., and Zhang, J.
\newblock I-sheep: Self-alignment of llm from scratch through an iterative self-enhancement paradigm.
\newblock \emph{arXiv preprint arXiv: 2408.08072}, 2024.

\bibitem[Liu et~al.(2024)Liu, Wei, Liu, Si, Zhang, Rao, Zheng, Peng, Yang, Zhou, and Dai]{liu2024best}
Liu, R., Wei, J., Liu, F., Si, C., Zhang, Y., Rao, J., Zheng, S., Peng, D., Yang, D., Zhou, D., and Dai, A.~M.
\newblock Best practices and lessons learned on synthetic data.
\newblock \emph{arXiv preprint arXiv: 2404.07503}, 2024.

\bibitem[Liu et~al.(2023)Liu, Qiao, Neiswanger, Wang, Tan, Tao, Li, Wang, Sun, Pangarkar, Fan, Gu, Miller, Zhuang, He, Li, Koto, Tang, Ranjan, Shen, Ren, Iriondo, Mu, Hu, Schulze, Nakov, Baldwin, and Xing]{liu2023llm360}
Liu, Z., Qiao, A., Neiswanger, W., Wang, H., Tan, B., Tao, T., Li, J., Wang, Y., Sun, S., Pangarkar, O., Fan, R., Gu, Y., Miller, V., Zhuang, Y., He, G., Li, H., Koto, F., Tang, L., Ranjan, N., Shen, Z., Ren, X., Iriondo, R., Mu, C., Hu, Z., Schulze, M., Nakov, P., Baldwin, T., and Xing, E.~P.
\newblock Llm360: Towards fully transparent open-source llms.
\newblock \emph{arXiv preprint arXiv: 2312.06550}, 2023.

\bibitem[Lu et~al.(2023)Lu, Zhong, Huang, Wang, Zhu, Mi, Wang, Wang, Zeng, Shang, Jiang, and Liu]{lu2023self}
Lu, J., Zhong, W., Huang, W., Wang, Y., Zhu, Q., Mi, F., Wang, B., Wang, W., Zeng, X., Shang, L., Jiang, X., and Liu, Q.
\newblock Self: Self-evolution with language feedback.
\newblock \emph{arXiv preprint arXiv: 2310.00533}, 2023.

\bibitem[Lupidi et~al.(2024)Lupidi, Gemmell, Cancedda, Dwivedi-Yu, Weston, Foerster, Raileanu, and Lomeli]{lupidi2024source2synth}
Lupidi, A., Gemmell, C., Cancedda, N., Dwivedi-Yu, J., Weston, J., Foerster, J., Raileanu, R., and Lomeli, M.
\newblock Source2synth: Synthetic data generation and curation grounded in real data sources.
\newblock \emph{arXiv preprint arXiv: 2409.08239}, 2024.

\bibitem[Madaan et~al.(2023)Madaan, Tandon, Gupta, Hallinan, Gao, Wiegreffe, Alon, Dziri, Prabhumoye, Yang, Gupta, Majumder, Hermann, Welleck, Yazdanbakhsh, and Clark]{madaan2023selfrefine}
Madaan, A., Tandon, N., Gupta, P., Hallinan, S., Gao, L., Wiegreffe, S., Alon, U., Dziri, N., Prabhumoye, S., Yang, Y., Gupta, S., Majumder, B.~P., Hermann, K., Welleck, S., Yazdanbakhsh, A., and Clark, P.
\newblock Self-refine: Iterative refinement with self-feedback.
\newblock \emph{arXiv preprint arXiv: 2303.17651}, 2023.

\bibitem[O'Neill et~al.(2023)O'Neill, Ting, Ciuca, Miller, and Bui]{o'neill2023steering}
O'Neill, C., Ting, Y.-S., Ciuca, I., Miller, J., and Bui, T.
\newblock Steering language generation: Harnessing contrastive expert guidance and negative prompting for coherent and diverse synthetic data generation.
\newblock \emph{arXiv preprint arXiv: 2308.07645}, 2023.

\bibitem[OpenAI et~al.(2023)OpenAI, Achiam, Adler, Agarwal, Ahmad, Akkaya, Aleman, Almeida, Altenschmidt, Altman, Anadkat, Avila, Babuschkin, Balaji, Balcom, Baltescu, Bao, Bavarian, Belgum, Bello, Berdine, Bernadett-Shapiro, Berner, Bogdonoff, Boiko, Boyd, Brakman, Brockman, Brooks, Brundage, Button, Cai, Campbell, Cann, Carey, Carlson, Carmichael, Chan, Chang, Chantzis, Chen, Chen, Chen, Chen, Chen, Chess, Cho, Chu, Chung, Cummings, Currier, Dai, Decareaux, Degry, Deutsch, Deville, Dhar, Dohan, Dowling, Dunning, Ecoffet, Eleti, Eloundou, Farhi, Fedus, Felix, Fishman, Forte, Fulford, Gao, Georges, Gibson, Goel, Gogineni, Goh, Gontijo-Lopes, Gordon, Grafstein, Gray, Greene, Gross, Gu, Guo, Hallacy, Han, Harris, He, Heaton, Heidecke, Hesse, Hickey, Hickey, Hoeschele, Houghton, Hsu, Hu, Hu, Huizinga, Jain, Jain, Jang, Jiang, Jiang, Jin, Jin, Jomoto, Jonn, Jun, Kaftan, Łukasz Kaiser, Kamali, Kanitscheider, Keskar, Khan, Kilpatrick, Kim, Kim, Kim, Kirchner, Kiros, Knight, Kokotajlo, Łukasz Kondraciuk, Kondrich,
  Konstantinidis, Kosic, Krueger, Kuo, Lampe, Lan, Lee, Leike, Leung, Levy, Li, Lim, Lin, Lin, Litwin, Lopez, Lowe, Lue, Makanju, Malfacini, Manning, Markov, Markovski, Martin, Mayer, Mayne, McGrew, McKinney, McLeavey, McMillan, McNeil, Medina, Mehta, Menick, Metz, Mishchenko, Mishkin, Monaco, Morikawa, Mossing, Mu, Murati, Murk, Mély, Nair, Nakano, Nayak, Neelakantan, Ngo, Noh, Ouyang, O'Keefe, Pachocki, Paino, Palermo, Pantuliano, Parascandolo, Parish, Parparita, Passos, Pavlov, Peng, Perelman, de~Avila Belbute~Peres, Petrov, de~Oliveira~Pinto, Michael, Pokorny, Pokrass, Pong, Powell, Power, Power, Proehl, Puri, Radford, Rae, Ramesh, Raymond, Real, Rimbach, Ross, Rotsted, Roussez, Ryder, Saltarelli, Sanders, Santurkar, Sastry, Schmidt, Schnurr, Schulman, Selsam, Sheppard, Sherbakov, Shieh, Shoker, Shyam, Sidor, Sigler, Simens, Sitkin, Slama, Sohl, Sokolowsky, Song, Staudacher, Such, Summers, Sutskever, Tang, Tezak, Thompson, Tillet, Tootoonchian, Tseng, Tuggle, Turley, Tworek, Uribe, Vallone, Vijayvergiya,
  Voss, Wainwright, Wang, Wang, Wang, Ward, Wei, Weinmann, Welihinda, Welinder, Weng, Weng, Wiethoff, Willner, Winter, Wolrich, Wong, Workman, Wu, Wu, Wu, Xiao, Xu, Yoo, Yu, Yuan, Zaremba, Zellers, Zhang, Zhang, Zhao, Zheng, Zhuang, Zhuk, and Zoph]{openai2023gpt4}
OpenAI, Achiam, J., Adler, S., Agarwal, S., Ahmad, L., Akkaya, I., Aleman, F.~L., Almeida, D., Altenschmidt, J., Altman, S., Anadkat, S., Avila, R., Babuschkin, I., Balaji, S., Balcom, V., Baltescu, P., Bao, H., Bavarian, M., Belgum, J., Bello, I., Berdine, J., Bernadett-Shapiro, G., Berner, C., Bogdonoff, L., Boiko, O., Boyd, M., Brakman, A.-L., Brockman, G., Brooks, T., Brundage, M., Button, K., Cai, T., Campbell, R., Cann, A., Carey, B., Carlson, C., Carmichael, R., Chan, B., Chang, C., Chantzis, F., Chen, D., Chen, S., Chen, R., Chen, J., Chen, M., Chess, B., Cho, C., Chu, C., Chung, H.~W., Cummings, D., Currier, J., Dai, Y., Decareaux, C., Degry, T., Deutsch, N., Deville, D., Dhar, A., Dohan, D., Dowling, S., Dunning, S., Ecoffet, A., Eleti, A., Eloundou, T., Farhi, D., Fedus, L., Felix, N., Fishman, S.~P., Forte, J., Fulford, I., Gao, L., Georges, E., Gibson, C., Goel, V., Gogineni, T., Goh, G., Gontijo-Lopes, R., Gordon, J., Grafstein, M., Gray, S., Greene, R., Gross, J., Gu, S.~S., Guo, Y., Hallacy,
  C., Han, J., Harris, J., He, Y., Heaton, M., Heidecke, J., Hesse, C., Hickey, A., Hickey, W., Hoeschele, P., Houghton, B., Hsu, K., Hu, S., Hu, X., Huizinga, J., Jain, S., Jain, S., Jang, J., Jiang, A., Jiang, R., Jin, H., Jin, D., Jomoto, S., Jonn, B., Jun, H., Kaftan, T., Łukasz Kaiser, Kamali, A., Kanitscheider, I., Keskar, N.~S., Khan, T., Kilpatrick, L., Kim, J.~W., Kim, C., Kim, Y., Kirchner, J.~H., Kiros, J., Knight, M., Kokotajlo, D., Łukasz Kondraciuk, Kondrich, A., Konstantinidis, A., Kosic, K., Krueger, G., Kuo, V., Lampe, M., Lan, I., Lee, T., Leike, J., Leung, J., Levy, D., Li, C.~M., Lim, R., Lin, M., Lin, S., Litwin, M., Lopez, T., Lowe, R., Lue, P., Makanju, A., Malfacini, K., Manning, S., Markov, T., Markovski, Y., Martin, B., Mayer, K., Mayne, A., McGrew, B., McKinney, S.~M., McLeavey, C., McMillan, P., McNeil, J., Medina, D., Mehta, A., Menick, J., Metz, L., Mishchenko, A., Mishkin, P., Monaco, V., Morikawa, E., Mossing, D., Mu, T., Murati, M., Murk, O., Mély, D., Nair, A., Nakano, R.,
  Nayak, R., Neelakantan, A., Ngo, R., Noh, H., Ouyang, L., O'Keefe, C., Pachocki, J., Paino, A., Palermo, J., Pantuliano, A., Parascandolo, G., Parish, J., Parparita, E., Passos, A., Pavlov, M., Peng, A., Perelman, A., de~Avila Belbute~Peres, F., Petrov, M., de~Oliveira~Pinto, H.~P., Michael, Pokorny, Pokrass, M., Pong, V.~H., Powell, T., Power, A., Power, B., Proehl, E., Puri, R., Radford, A., Rae, J., Ramesh, A., Raymond, C., Real, F., Rimbach, K., Ross, C., Rotsted, B., Roussez, H., Ryder, N., Saltarelli, M., Sanders, T., Santurkar, S., Sastry, G., Schmidt, H., Schnurr, D., Schulman, J., Selsam, D., Sheppard, K., Sherbakov, T., Shieh, J., Shoker, S., Shyam, P., Sidor, S., Sigler, E., Simens, M., Sitkin, J., Slama, K., Sohl, I., Sokolowsky, B., Song, Y., Staudacher, N., Such, F.~P., Summers, N., Sutskever, I., Tang, J., Tezak, N., Thompson, M.~B., Tillet, P., Tootoonchian, A., Tseng, E., Tuggle, P., Turley, N., Tworek, J., Uribe, J. F.~C., Vallone, A., Vijayvergiya, A., Voss, C., Wainwright, C., Wang,
  J.~J., Wang, A., Wang, B., Ward, J., Wei, J., Weinmann, C., Welihinda, A., Welinder, P., Weng, J., Weng, L., Wiethoff, M., Willner, D., Winter, C., Wolrich, S., Wong, H., Workman, L., Wu, S., Wu, J., Wu, M., Xiao, K., Xu, T., Yoo, S., Yu, K., Yuan, Q., Zaremba, W., Zellers, R., Zhang, C., Zhang, M., Zhao, S., Zheng, T., Zhuang, J., Zhuk, W., and Zoph, B.
\newblock Gpt-4 technical report.
\newblock \emph{arXiv preprint arXiv: 2303.08774}, 2023.

\bibitem[Peng et~al.(2023)Peng, Li, He, Galley, and Gao]{peng2023instruction}
Peng, B., Li, C., He, P., Galley, M., and Gao, J.
\newblock Instruction tuning with gpt-4.
\newblock \emph{arXiv preprint arXiv: 2304.03277}, 2023.

\bibitem[Reimers \& Gurevych(2019)Reimers and Gurevych]{reimers-2019-sentence-bert}
Reimers, N. and Gurevych, I.
\newblock Sentence-bert: Sentence embeddings using siamese bert-networks.
\newblock In \emph{Proceedings of the 2019 Conference on Empirical Methods in Natural Language Processing}. Association for Computational Linguistics, 11 2019.
\newblock URL \url{https://arxiv.org/abs/1908.10084}.

\bibitem[Rein et~al.(2023)Rein, Hou, Stickland, Petty, Pang, Dirani, Michael, and Bowman]{rein2023gpqa}
Rein, D., Hou, B.~L., Stickland, A.~C., Petty, J., Pang, R.~Y., Dirani, J., Michael, J., and Bowman, S.~R.
\newblock Gpqa: A graduate-level google-proof q\&a benchmark.
\newblock \emph{arXiv preprint arXiv: 2311.12022}, 2023.

\bibitem[Renze \& Guven(2024)Renze and Guven]{renze2024selfreflection}
Renze, M. and Guven, E.
\newblock Self-reflection in llm agents: Effects on problem-solving performance.
\newblock \emph{arXiv preprint arXiv: 2405.06682}, 2024.

\bibitem[Sanh et~al.(2021)Sanh, Webson, Raffel, Bach, Sutawika, Alyafeai, Chaffin, Stiegler, Scao, Raja, Dey, Bari, Xu, Thakker, Sharma, Szczechla, Kim, Chhablani, Nayak, Datta, Chang, Jiang, Wang, Manica, Shen, Yong, Pandey, Bawden, Wang, Neeraj, Rozen, Sharma, Santilli, Fevry, Fries, Teehan, Bers, Biderman, Gao, Wolf, and Rush]{sanh2021multitask}
Sanh, V., Webson, A., Raffel, C., Bach, S.~H., Sutawika, L., Alyafeai, Z., Chaffin, A., Stiegler, A., Scao, T.~L., Raja, A., Dey, M., Bari, M.~S., Xu, C., Thakker, U., Sharma, S.~S., Szczechla, E., Kim, T., Chhablani, G., Nayak, N., Datta, D., Chang, J., Jiang, M. T.-J., Wang, H., Manica, M., Shen, S., Yong, Z.~X., Pandey, H., Bawden, R., Wang, T., Neeraj, T., Rozen, J., Sharma, A., Santilli, A., Fevry, T., Fries, J.~A., Teehan, R., Bers, T., Biderman, S., Gao, L., Wolf, T., and Rush, A.~M.
\newblock Multitask prompted training enables zero-shot task generalization.
\newblock \emph{arXiv preprint arXiv: 2110.08207}, 2021.

\bibitem[Scheurer et~al.(2022)Scheurer, Campos, Chan, Chen, Cho, and Perez]{scheurer2022training}
Scheurer, J., Campos, J.~A., Chan, J.~S., Chen, A., Cho, K., and Perez, E.
\newblock Training language models with language feedback.
\newblock \emph{arXiv preprint arXiv: 2204.14146}, 2022.

\bibitem[Shumailov et~al.(2023)Shumailov, Shumaylov, Zhao, Gal, Papernot, and Anderson]{shumailov2023curse}
Shumailov, I., Shumaylov, Z., Zhao, Y., Gal, Y., Papernot, N., and Anderson, R.
\newblock The curse of recursion: Training on generated data makes models forget.
\newblock \emph{arXiv preprint arXiv: 2305.17493}, 2023.

\bibitem[Sun et~al.(2023)Sun, Shen, Zhou, Zhang, Chen, Cox, Yang, and Gan]{sun2023principledriven}
Sun, Z., Shen, Y., Zhou, Q., Zhang, H., Chen, Z., Cox, D., Yang, Y., and Gan, C.
\newblock Principle-driven self-alignment of language models from scratch with minimal human supervision.
\newblock \emph{arXiv preprint arXiv: 2305.03047}, 2023.

\bibitem[Taori et~al.(2023)Taori, Gulrajani, Zhang, Dubois, Li, Guestrin, Liang, and Hashimoto]{alpaca}
Taori, R., Gulrajani, I., Zhang, T., Dubois, Y., Li, X., Guestrin, C., Liang, P., and Hashimoto, T.~B.
\newblock Stanford alpaca: An instruction-following llama model.
\newblock \url{https://github.com/tatsu-lab/stanford_alpaca}, 2023.

\bibitem[Team et~al.(2023)Team, Anil, Borgeaud, Alayrac, Yu, Soricut, Schalkwyk, Dai, Hauth, Millican, Silver, Johnson, Antonoglou, Schrittwieser, Glaese, Chen, Pitler, Lillicrap, Lazaridou, Firat, Molloy, Isard, Barham, Hennigan, Lee, Viola, Reynolds, Xu, Doherty, Collins, Meyer, Rutherford, Moreira, Ayoub, Goel, Krawczyk, Du, Chi, Cheng, Ni, Shah, Kane, Chan, Faruqui, Severyn, Lin, Li, Cheng, Ittycheriah, Mahdieh, Chen, Sun, Tran, Bagri, Lakshminarayanan, Liu, Orban, Güra, Zhou, Song, Boffy, Ganapathy, Zheng, Choe, Ágoston Weisz, Zhu, Lu, Gopal, Kahn, Kula, Pitman, Shah, Taropa, Merey, Baeuml, Chen, Shafey, Zhang, Sercinoglu, Tucker, Piqueras, Krikun, Barr, Savinov, Danihelka, Roelofs, White, Andreassen, von Glehn, Yagati, Kazemi, Gonzalez, Khalman, Sygnowski, Frechette, Smith, Culp, Proleev, Luan, Chen, Lottes, Schucher, Lebron, Rrustemi, Clay, Crone, Kocisky, Zhao, Perz, Yu, Howard, Bloniarz, Rae, Lu, Sifre, Maggioni, Alcober, Garrette, Barnes, Thakoor, Austin, Barth-Maron, Wong, Joshi, Chaabouni,
  Fatiha, Ahuja, Tomar, Senter, Chadwick, Kornakov, Attaluri, Iturrate, Liu, Li, Cogan, Chen, Jia, Gu, Zhang, Grimstad, Hartman, Garcia, Pillai, Devlin, Laskin, de~Las~Casas, Valter, Tao, Blanco, Badia, Reitter, Chen, Brennan, Rivera, Brin, Iqbal, Surita, Labanowski, Rao, Winkler, Parisotto, Gu, Olszewska, Addanki, Miech, Louis, Teplyashin, Brown, Catt, Balaguer, Xiang, Wang, Ashwood, Briukhov, Webson, Ganapathy, Sanghavi, Kannan, Chang, Stjerngren, Djolonga, Sun, Bapna, Aitchison, Pejman, Michalewski, Yu, Wang, Love, Ahn, Bloxwich, Han, Humphreys, Sellam, Bradbury, Godbole, Samangooei, Damoc, Kaskasoli, Arnold, Vasudevan, Agrawal, Riesa, Lepikhin, Tanburn, Srinivasan, Lim, Hodkinson, Shyam, Ferret, Hand, Garg, Paine, Li, Li, Giang, Neitz, Abbas, York, Reid, Cole, Chowdhery, Das, Rogozińska, Nikolaev, Sprechmann, Nado, Zilka, Prost, He, Monteiro, Mishra, Welty, Newlan, Jia, Allamanis, Hu, de~Liedekerke, Gilmer, Saroufim, Rijhwani, Hou, Shrivastava, Baddepudi, Goldin, Ozturel, Cassirer, Xu, Sohn, Sachan,
  Amplayo, Swanson, Petrova, Narayan, Guez, Brahma, Landon, Patel, Zhao, Villela, Wang, Jia, Rahtz, Giménez, Yeung, Keeling, Georgiev, Mincu, Wu, Haykal, Saputro, Vodrahalli, Qin, Cankara, Sharma, Fernando, Hawkins, Neyshabur, Kim, Hutter, Agrawal, Castro-Ros, van~den Driessche, Wang, Yang, yiin Chang, Komarek, McIlroy, Lučić, Zhang, Farhan, Sharman, Natsev, Michel, Bansal, Qiao, Cao, Shakeri, Butterfield, Chung, Rubenstein, Agrawal, Mensch, Soparkar, Lenc, Chung, Pope, Maggiore, Kay, Jhakra, Wang, Maynez, Phuong, Tobin, Tacchetti, Trebacz, Robinson, Katariya, Riedel, Bailey, Xiao, Ghelani, Aroyo, Slone, Houlsby, Xiong, Yang, Gribovskaya, Adler, Wirth, Lee, Li, Kagohara, Pavagadhi, Bridgers, Bortsova, Ghemawat, Ahmed, Liu, Powell, Bolina, Iinuma, Zablotskaia, Besley, Chung, Dozat, Comanescu, Si, Greer, Su, Polacek, Kaufman, Tokumine, Hu, Buchatskaya, Miao, Elhawaty, Siddhant, Tomasev, Xing, Greer, Miller, Ashraf, Roy, Zhang, Ma, Filos, Besta, Blevins, Klimenko, Yeh, Changpinyo, Mu, Chang, Pajarskas, Muir,
  Cohen, Lan, Haridasan, Marathe, Hansen, Douglas, Samuel, Wang, Austin, Lan, Jiang, Chiu, Lorenzo, Sjösund, Cevey, Gleicher, Avrahami, Boral, Srinivasan, Selo, May, Aisopos, Hussenot, Soares, Baumli, Chang, Recasens, Caine, Pritzel, Pavetic, Pardo, Gergely, Frye, Ramasesh, Horgan, Badola, Kassner, Roy, Dyer, Campos, Tomala, Tang, Badawy, White, Mustafa, Lang, Jindal, Vikram, Gong, Caelles, Hemsley, Thornton, Feng, Stokowiec, Zheng, Thacker, Çağlar Ünlü, Zhang, Saleh, Svensson, Bileschi, Patil, Anand, Ring, Tsihlas, Vezer, Selvi, Shevlane, Rodriguez, Kwiatkowski, Daruki, Rong, Dafoe, FitzGerald, Gu-Lemberg, Khan, Hendricks, Pellat, Feinberg, Cobon-Kerr, Sainath, Rauh, Hashemi, Ives, Hasson, Noland, Cao, Byrd, Hou, Wang, Sottiaux, Paganini, Lespiau, Moufarek, Hassan, Shivakumar, van Amersfoort, Mandhane, Joshi, Goyal, Tung, Brock, Sheahan, Misra, Li, Rakićević, Dehghani, Liu, Mittal, Oh, Noury, Sezener, Huot, Lamm, Cao, Chen, Mudgal, Stella, Brooks, Vasudevan, Liu, Chain, Melinkeri, Cohen, Wang,
  Seymore, Zubkov, Goel, Yue, Krishnakumaran, Albert, Hurley, Sano, Mohananey, Joughin, Filonov, Kępa, Eldawy, Lim, Rishi, Badiezadegan, Bos, Chang, Jain, Padmanabhan, Puttagunta, Krishna, Baker, Kalb, Bedapudi, Kurzrok, Lei, Yu, Litvin, Zhou, Wu, Sobell, Siciliano, Papir, Neale, Bragagnolo, Toor, Chen, Anklin, Wang, Feng, Gholami, Ling, Liu, Walter, Moghaddam, Kishore, Adamek, Mercado, Mallinson, Wandekar, Cagle, Ofek, Garrido, Lombriser, Mukha, Sun, Mohammad, Matak, Qian, Peswani, Janus, Yuan, Schelin, David, Garg, He, Duzhyi, Älgmyr, Lottaz, Li, Yadav, Xu, Chinien, Shivanna, Chuklin, Li, Spadine, Wolfe, Mohamed, Das, Dai, He, von Dincklage, Upadhyay, Maurya, Chi, Krause, Salama, Rabinovitch, M, Selvan, Dektiarev, Ghiasi, Guven, Gupta, Liu, Sharma, Shtacher, Paul, Akerlund, Aubet, Huang, Zhu, Zhu, Teixeira, Fritze, Bertolini, Marinescu, Bölle, Paulus, Gupta, Latkar, Chang, Sanders, Wilson, Wu, Tan, Thiet, Doshi, Lall, Mishra, Chen, Luong, Benjamin, Lee, Andrejczuk, Rabiej, Ranjan, Styrc, Yin, Simon,
  Harriott, Bansal, Robsky, Bacon, Greene, Mirylenka, Zhou, Sarvana, Goyal, Andermatt, Siegler, Horn, Israel, Pongetti, Chen, Selvatici, Silva, Wang, Tolins, Guu, Yogev, Cai, Agostini, Shah, Nguyen, Donnaile, Pereira, Friso, Stambler, Kurzrok, Kuang, Romanikhin, Geller, Yan, Jang, Lee, Fica, Malmi, Tan, Banica, Balle, Pham, Huang, Avram, Shi, Singh, Hidey, Ahuja, Saxena, Dooley, Potharaju, O'Neill, Gokulchandran, Foley, Zhao, Dusenberry, Liu, Mehta, Kotikalapudi, Safranek-Shrader, Goodman, Kessinger, Globen, Kolhar, Gorgolewski, Ibrahim, Song, Eichenbaum, Brovelli, Potluri, Lahoti, Baetu, Ghorbani, Chen, Crawford, Pal, Sridhar, Gurita, Mujika, Petrovski, Cedoz, Li, Chen, Santo, Goyal, Punjabi, Kappaganthu, Kwak, LV, Velury, Choudhury, Hall, Shah, Figueira, Thomas, Lu, Zhou, Kumar, Jurdi, Chikkerur, Ma, Yu, Kwak, Ähdel, Rajayogam, Choma, Liu, Barua, Ji, Park, Hellendoorn, Bailey, Bilal, Zhou, Khatir, Sutton, Rzadkowski, Macintosh, Shagin, Medina, Liang, Zhou, Shah, Bi, Dankovics, Banga, Lehmann, Bredesen,
  Lin, Hoffmann, Lai, Chung, Yang, Balani, Bražinskas, Sozanschi, Hayes, Alcalde, Makarov, Chen, Stella, Snijders, Mandl, Kärrman, Nowak, Wu, Dyck, Vaidyanathan, R, Mallet, Rudominer, Johnston, Mittal, Udathu, Christensen, Verma, Irving, Santucci, Elsayed, Davoodi, Georgiev, Tenney, Hua, Cideron, Leurent, Alnahlawi, Georgescu, Wei, Zheng, Scandinaro, Jiang, Snoek, Sundararajan, Wang, Ontiveros, Karo, Cole, Rajashekhar, Tumeh, Ben-David, Jain, Uesato, Datta, Bunyan, Wu, Zhang, Stanczyk, Zhang, Steiner, Naskar, Azzam, Johnson, Paszke, Chiu, Elias, Mohiuddin, Muhammad, Miao, Lee, Vieillard, Park, Zhang, Stanway, Garmon, Karmarkar, Dong, Lee, Kumar, Zhou, Evens, Isaac, Irving, Loper, Fink, Arkatkar, Chen, Shafran, Petrychenko, Chen, Jia, Levskaya, Zhu, Grabowski, Mao, Magni, Yao, Snaider, Casagrande, Palmer, Suganthan, Castaño, Giannoumis, Kim, Rybiński, Sreevatsa, Prendki, Soergel, Goedeckemeyer, Gierke, Jafari, Gaba, Wiesner, Wright, Wei, Vashisht, Kulizhskaya, Hoover, Le, Li, Iwuanyanwu, Liu, Ramirez,
  Khorlin, Cui, LIN, Wu, Aguilar, Pallo, Chakladar, Perng, Abellan, Zhang, Dasgupta, Kushman, Penchev, Repina, Wu, van~der Weide, Ponnapalli, Kaplan, Simsa, Li, Dousse, Yang, Piper, Ie, Pasumarthi, Lintz, Vijayakumar, Andor, Valenzuela, Lui, Paduraru, Peng, Lee, Zhang, Greene, Nguyen, Kurylowicz, Hardin, Dixon, Janzer, Choo, Feng, Zhang, Singhal, Du, McKinnon, Antropova, Bolukbasi, Keller, Reid, Finchelstein, Raad, Crocker, Hawkins, Dadashi, Gaffney, Franko, Bulanova, Leblond, Chung, Askham, Cobo, Xu, Fischer, Xu, Sorokin, Alberti, Lin, Evans, Dimitriev, Forbes, Banarse, Tung, Omernick, Bishop, Sterneck, Jain, Xia, Amid, Piccinno, Wang, Banzal, Mankowitz, Polozov, Krakovna, Brown, Bateni, Duan, Firoiu, Thotakuri, Natan, Geist, tan Girgin, Li, Ye, Roval, Tojo, Kwong, Lee-Thorp, Yew, Sinopalnikov, Ramos, Mellor, Sharma, Wu, Miller, Sonnerat, Vnukov, Greig, Beattie, Caveness, Bai, Eisenschlos, Korchemniy, Tsai, Jasarevic, Kong, Dao, Zheng, Liu, Yang, Zhu, Teh, Sanmiya, Gladchenko, Trdin, Toyama, Rosen, Tavakkol,
  Xue, Elkind, Woodman, Carpenter, Papamakarios, Kemp, Kafle, Grunina, Sinha, Talbert, Wu, Owusu-Afriyie, Du, Thornton, Pont-Tuset, Narayana, Li, Fatehi, Wieting, Ajmeri, Uria, Ko, Knight, Héliou, Niu, Gu, Pang, Li, Levine, Stolovich, Santamaria-Fernandez, Goenka, Yustalim, Strudel, Elqursh, Deck, Lee, Li, Levin, Hoffmann, Holtmann-Rice, Bachem, Arora, Koh, Yeganeh, Põder, Tariq, Sun, Ionita, Seyedhosseini, Tafti, Liu, Gulati, Liu, Ye, Chrzaszcz, Wang, Sethi, Li, Brown, Singh, Fan, Parisi, Stanton, Koverkathu, Choquette-Choo, Li, Lu, Ittycheriah, Shroff, Varadarajan, Bahargam, Willoughby, Gaddy, Desjardins, Cornero, Robenek, Mittal, Albrecht, Shenoy, Moiseev, Jacobsson, Ghaffarkhah, Rivière, Walton, Crepy, Parrish, Zhou, Farabet, Radebaugh, Srinivasan, van~der Salm, Fidjeland, Scellato, Latorre-Chimoto, Klimczak-Plucińska, Bridson, de~Cesare, Hudson, Mendolicchio, Walker, Morris, Mauger, Guseynov, Reid, Odoom, Loher, Cotruta, Yenugula, Grewe, Petrushkina, Duerig, Sanchez, Yadlowsky, Shen, Globerson, Webb,
  Dua, Li, Bhupatiraju, Hurt, Qureshi, Agarwal, Shani, Eyal, Khare, Belle, Wang, Tekur, Kale, Wei, Sang, Saeta, Liechty, Sun, Zhao, Lee, Nayak, Fritz, Vuyyuru, Aslanides, Vyas, Wicke, Ma, Eltyshev, Martin, Cate, Manyika, Amiri, Kim, Xiong, Kang, Luisier, Tripuraneni, Madras, Guo, Waters, Wang, Ainslie, Baldridge, Zhang, Pruthi, Bauer, Yang, Mansour, Gelman, Xu, Polovets, Liu, Cai, Chen, Sheng, Xue, Ozair, Angermueller, Li, Sinha, Wang, Wiesinger, Koukoumidis, Tian, Iyer, Gurumurthy, Goldenson, Shah, Blake, Yu, Urbanowicz, Palomaki, Fernando, Durden, Mehta, Momchev, Rahimtoroghi, Georgaki, Raul, Ruder, Redshaw, Lee, Zhou, Jalan, Li, Hechtman, Schuh, Nasr, Milan, Mikulik, Franco, Green, Nguyen, Kelley, Mahendru, Hu, Howland, Vargas, Hui, Bansal, Rao, Ghiya, Wang, Ye, Sarr, Preston, Elish, Li, Kaku, Gupta, Pasupat, Juan, Someswar, M., Chen, Amini, Fabrikant, Chu, Dong, Muthal, Buthpitiya, Jauhari, Hua, Khandelwal, Hitron, Ren, Rinaldi, Drath, Dabush, Jiang, Godhia, Sachs, Chen, Fan, Taitelbaum, Noga, Dai, Wang,
  Liang, Hamer, Ferng, Elkind, Atias, Lee, Listík, Carlen, van~de Kerkhof, Pikus, Zaher, Müller, Zykova, Stefanec, Gatsko, Hirnschall, Sethi, Xu, Ahuja, Tsai, Stefanoiu, Feng, Dhandhania, Katyal, Gupta, Parulekar, Pitta, Zhao, Bhatia, Bhavnani, Alhadlaq, Li, Danenberg, Tu, Pine, Filippova, Ghosh, Limonchik, Urala, Lanka, Clive, Sun, Li, Wu, Hongtongsak, Li, Thakkar, Omarov, Majmundar, Alverson, Kucharski, Patel, Jain, Zabelin, Pelagatti, Kohli, Kumar, Kim, Sankar, Shah, Ramachandruni, Zeng, Bariach, Weidinger, Vu, Andreev, He, Hui, Kashem, Subramanya, Hsiao, Hassabis, Kavukcuoglu, Sadovsky, Le, Strohman, Wu, Petrov, Dean, and Vinyals]{team2023gemini}
Team, G., Anil, R., Borgeaud, S., Alayrac, J.-B., Yu, J., Soricut, R., Schalkwyk, J., Dai, A.~M., Hauth, A., Millican, K., Silver, D., Johnson, M., Antonoglou, I., Schrittwieser, J., Glaese, A., Chen, J., Pitler, E., Lillicrap, T., Lazaridou, A., Firat, O., Molloy, J., Isard, M., Barham, P.~R., Hennigan, T., Lee, B., Viola, F., Reynolds, M., Xu, Y., Doherty, R., Collins, E., Meyer, C., Rutherford, E., Moreira, E., Ayoub, K., Goel, M., Krawczyk, J., Du, C., Chi, E., Cheng, H.-T., Ni, E., Shah, P., Kane, P., Chan, B., Faruqui, M., Severyn, A., Lin, H., Li, Y., Cheng, Y., Ittycheriah, A., Mahdieh, M., Chen, M., Sun, P., Tran, D., Bagri, S., Lakshminarayanan, B., Liu, J., Orban, A., Güra, F., Zhou, H., Song, X., Boffy, A., Ganapathy, H., Zheng, S., Choe, H., Ágoston Weisz, Zhu, T., Lu, Y., Gopal, S., Kahn, J., Kula, M., Pitman, J., Shah, R., Taropa, E., Merey, M.~A., Baeuml, M., Chen, Z., Shafey, L.~E., Zhang, Y., Sercinoglu, O., Tucker, G., Piqueras, E., Krikun, M., Barr, I., Savinov, N., Danihelka, I.,
  Roelofs, B., White, A., Andreassen, A., von Glehn, T., Yagati, L., Kazemi, M., Gonzalez, L., Khalman, M., Sygnowski, J., Frechette, A., Smith, C., Culp, L., Proleev, L., Luan, Y., Chen, X., Lottes, J., Schucher, N., Lebron, F., Rrustemi, A., Clay, N., Crone, P., Kocisky, T., Zhao, J., Perz, B., Yu, D., Howard, H., Bloniarz, A., Rae, J.~W., Lu, H., Sifre, L., Maggioni, M., Alcober, F., Garrette, D., Barnes, M., Thakoor, S., Austin, J., Barth-Maron, G., Wong, W., Joshi, R., Chaabouni, R., Fatiha, D., Ahuja, A., Tomar, G.~S., Senter, E., Chadwick, M., Kornakov, I., Attaluri, N., Iturrate, I., Liu, R., Li, Y., Cogan, S., Chen, J., Jia, C., Gu, C., Zhang, Q., Grimstad, J., Hartman, A.~J., Garcia, X., Pillai, T.~S., Devlin, J., Laskin, M., de~Las~Casas, D., Valter, D., Tao, C., Blanco, L., Badia, A.~P., Reitter, D., Chen, M., Brennan, J., Rivera, C., Brin, S., Iqbal, S., Surita, G., Labanowski, J., Rao, A., Winkler, S., Parisotto, E., Gu, Y., Olszewska, K., Addanki, R., Miech, A., Louis, A., Teplyashin, D.,
  Brown, G., Catt, E., Balaguer, J., Xiang, J., Wang, P., Ashwood, Z., Briukhov, A., Webson, A., Ganapathy, S., Sanghavi, S., Kannan, A., Chang, M.-W., Stjerngren, A., Djolonga, J., Sun, Y., Bapna, A., Aitchison, M., Pejman, P., Michalewski, H., Yu, T., Wang, C., Love, J., Ahn, J., Bloxwich, D., Han, K., Humphreys, P., Sellam, T., Bradbury, J., Godbole, V., Samangooei, S., Damoc, B., Kaskasoli, A., Arnold, S. M.~R., Vasudevan, V., Agrawal, S., Riesa, J., Lepikhin, D., Tanburn, R., Srinivasan, S., Lim, H., Hodkinson, S., Shyam, P., Ferret, J., Hand, S., Garg, A., Paine, T.~L., Li, J., Li, Y., Giang, M., Neitz, A., Abbas, Z., York, S., Reid, M., Cole, E., Chowdhery, A., Das, D., Rogozińska, D., Nikolaev, V., Sprechmann, P., Nado, Z., Zilka, L., Prost, F., He, L., Monteiro, M., Mishra, G., Welty, C., Newlan, J., Jia, D., Allamanis, M., Hu, C.~H., de~Liedekerke, R., Gilmer, J., Saroufim, C., Rijhwani, S., Hou, S., Shrivastava, D., Baddepudi, A., Goldin, A., Ozturel, A., Cassirer, A., Xu, Y., Sohn, D., Sachan,
  D., Amplayo, R.~K., Swanson, C., Petrova, D., Narayan, S., Guez, A., Brahma, S., Landon, J., Patel, M., Zhao, R., Villela, K., Wang, L., Jia, W., Rahtz, M., Giménez, M., Yeung, L., Keeling, J., Georgiev, P., Mincu, D., Wu, B., Haykal, S., Saputro, R., Vodrahalli, K., Qin, J., Cankara, Z., Sharma, A., Fernando, N., Hawkins, W., Neyshabur, B., Kim, S., Hutter, A., Agrawal, P., Castro-Ros, A., van~den Driessche, G., Wang, T., Yang, F., yiin Chang, S., Komarek, P., McIlroy, R., Lučić, M., Zhang, G., Farhan, W., Sharman, M., Natsev, P., Michel, P., Bansal, Y., Qiao, S., Cao, K., Shakeri, S., Butterfield, C., Chung, J., Rubenstein, P.~K., Agrawal, S., Mensch, A., Soparkar, K., Lenc, K., Chung, T., Pope, A., Maggiore, L., Kay, J., Jhakra, P., Wang, S., Maynez, J., Phuong, M., Tobin, T., Tacchetti, A., Trebacz, M., Robinson, K., Katariya, Y., Riedel, S., Bailey, P., Xiao, K., Ghelani, N., Aroyo, L., Slone, A., Houlsby, N., Xiong, X., Yang, Z., Gribovskaya, E., Adler, J., Wirth, M., Lee, L., Li, M., Kagohara, T.,
  Pavagadhi, J., Bridgers, S., Bortsova, A., Ghemawat, S., Ahmed, Z., Liu, T., Powell, R., Bolina, V., Iinuma, M., Zablotskaia, P., Besley, J., Chung, D.-W., Dozat, T., Comanescu, R., Si, X., Greer, J., Su, G., Polacek, M., Kaufman, R.~L., Tokumine, S., Hu, H., Buchatskaya, E., Miao, Y., Elhawaty, M., Siddhant, A., Tomasev, N., Xing, J., Greer, C., Miller, H., Ashraf, S., Roy, A., Zhang, Z., Ma, A., Filos, A., Besta, M., Blevins, R., Klimenko, T., Yeh, C.-K., Changpinyo, S., Mu, J., Chang, O., Pajarskas, M., Muir, C., Cohen, V., Lan, C.~L., Haridasan, K., Marathe, A., Hansen, S., Douglas, S., Samuel, R., Wang, M., Austin, S., Lan, C., Jiang, J., Chiu, J., Lorenzo, J.~A., Sjösund, L.~L., Cevey, S., Gleicher, Z., Avrahami, T., Boral, A., Srinivasan, H., Selo, V., May, R., Aisopos, K., Hussenot, L., Soares, L.~B., Baumli, K., Chang, M.~B., Recasens, A., Caine, B., Pritzel, A., Pavetic, F., Pardo, F., Gergely, A., Frye, J., Ramasesh, V., Horgan, D., Badola, K., Kassner, N., Roy, S., Dyer, E., Campos, V.~C.,
  Tomala, A., Tang, Y., Badawy, D.~E., White, E., Mustafa, B., Lang, O., Jindal, A., Vikram, S., Gong, Z., Caelles, S., Hemsley, R., Thornton, G., Feng, F., Stokowiec, W., Zheng, C., Thacker, P., Çağlar Ünlü, Zhang, Z., Saleh, M., Svensson, J., Bileschi, M., Patil, P., Anand, A., Ring, R., Tsihlas, K., Vezer, A., Selvi, M., Shevlane, T., Rodriguez, M., Kwiatkowski, T., Daruki, S., Rong, K., Dafoe, A., FitzGerald, N., Gu-Lemberg, K., Khan, M., Hendricks, L.~A., Pellat, M., Feinberg, V., Cobon-Kerr, J., Sainath, T., Rauh, M., Hashemi, S.~H., Ives, R., Hasson, Y., Noland, E., Cao, Y., Byrd, N., Hou, L., Wang, Q., Sottiaux, T., Paganini, M., Lespiau, J.-B., Moufarek, A., Hassan, S., Shivakumar, K., van Amersfoort, J., Mandhane, A., Joshi, P., Goyal, A., Tung, M., Brock, A., Sheahan, H., Misra, V., Li, C., Rakićević, N., Dehghani, M., Liu, F., Mittal, S., Oh, J., Noury, S., Sezener, E., Huot, F., Lamm, M., Cao, N.~D., Chen, C., Mudgal, S., Stella, R., Brooks, K., Vasudevan, G., Liu, C., Chain, M., Melinkeri,
  N., Cohen, A., Wang, V., Seymore, K., Zubkov, S., Goel, R., Yue, S., Krishnakumaran, S., Albert, B., Hurley, N., Sano, M., Mohananey, A., Joughin, J., Filonov, E., Kępa, T., Eldawy, Y., Lim, J., Rishi, R., Badiezadegan, S., Bos, T., Chang, J., Jain, S., Padmanabhan, S. G.~S., Puttagunta, S., Krishna, K., Baker, L., Kalb, N., Bedapudi, V., Kurzrok, A., Lei, S., Yu, A., Litvin, O., Zhou, X., Wu, Z., Sobell, S., Siciliano, A., Papir, A., Neale, R., Bragagnolo, J., Toor, T., Chen, T., Anklin, V., Wang, F., Feng, R., Gholami, M., Ling, K., Liu, L., Walter, J., Moghaddam, H., Kishore, A., Adamek, J., Mercado, T., Mallinson, J., Wandekar, S., Cagle, S., Ofek, E., Garrido, G., Lombriser, C., Mukha, M., Sun, B., Mohammad, H.~R., Matak, J., Qian, Y., Peswani, V., Janus, P., Yuan, Q., Schelin, L., David, O., Garg, A., He, Y., Duzhyi, O., Älgmyr, A., Lottaz, T., Li, Q., Yadav, V., Xu, L., Chinien, A., Shivanna, R., Chuklin, A., Li, J., Spadine, C., Wolfe, T., Mohamed, K., Das, S., Dai, Z., He, K., von Dincklage, D.,
  Upadhyay, S., Maurya, A., Chi, L., Krause, S., Salama, K., Rabinovitch, P.~G., M, P. K.~R., Selvan, A., Dektiarev, M., Ghiasi, G., Guven, E., Gupta, H., Liu, B., Sharma, D., Shtacher, I.~H., Paul, S., Akerlund, O., Aubet, F.-X., Huang, T., Zhu, C., Zhu, E., Teixeira, E., Fritze, M., Bertolini, F., Marinescu, L.-E., Bölle, M., Paulus, D., Gupta, K., Latkar, T., Chang, M., Sanders, J., Wilson, R., Wu, X., Tan, Y.-X., Thiet, L.~N., Doshi, T., Lall, S., Mishra, S., Chen, W., Luong, T., Benjamin, S., Lee, J., Andrejczuk, E., Rabiej, D., Ranjan, V., Styrc, K., Yin, P., Simon, J., Harriott, M.~R., Bansal, M., Robsky, A., Bacon, G., Greene, D., Mirylenka, D., Zhou, C., Sarvana, O., Goyal, A., Andermatt, S., Siegler, P., Horn, B., Israel, A., Pongetti, F., Chen, C.-W.~L., Selvatici, M., Silva, P., Wang, K., Tolins, J., Guu, K., Yogev, R., Cai, X., Agostini, A., Shah, M., Nguyen, H., Donnaile, N.~Ã., Pereira, S., Friso, L., Stambler, A., Kurzrok, A., Kuang, C., Romanikhin, Y., Geller, M., Yan, Z., Jang, K., Lee,
  C.-C., Fica, W., Malmi, E., Tan, Q., Banica, D., Balle, D., Pham, R., Huang, Y., Avram, D., Shi, H., Singh, J., Hidey, C., Ahuja, N., Saxena, P., Dooley, D., Potharaju, S.~P., O'Neill, E., Gokulchandran, A., Foley, R., Zhao, K., Dusenberry, M., Liu, Y., Mehta, P., Kotikalapudi, R., Safranek-Shrader, C., Goodman, A., Kessinger, J., Globen, E., Kolhar, P., Gorgolewski, C., Ibrahim, A., Song, Y., Eichenbaum, A., Brovelli, T., Potluri, S., Lahoti, P., Baetu, C., Ghorbani, A., Chen, C., Crawford, A., Pal, S., Sridhar, M., Gurita, P., Mujika, A., Petrovski, I., Cedoz, P.-L., Li, C., Chen, S., Santo, N.~D., Goyal, S., Punjabi, J., Kappaganthu, K., Kwak, C., LV, P., Velury, S., Choudhury, H., Hall, J., Shah, P., Figueira, R., Thomas, M., Lu, M., Zhou, T., Kumar, C., Jurdi, T., Chikkerur, S., Ma, Y., Yu, A., Kwak, S., Ähdel, V., Rajayogam, S., Choma, T., Liu, F., Barua, A., Ji, C., Park, J.~H., Hellendoorn, V., Bailey, A., Bilal, T., Zhou, H., Khatir, M., Sutton, C., Rzadkowski, W., Macintosh, F., Shagin, K.,
  Medina, P., Liang, C., Zhou, J., Shah, P., Bi, Y., Dankovics, A., Banga, S., Lehmann, S., Bredesen, M., Lin, Z., Hoffmann, J.~E., Lai, J., Chung, R., Yang, K., Balani, N., Bražinskas, A., Sozanschi, A., Hayes, M., Alcalde, H.~F., Makarov, P., Chen, W., Stella, A., Snijders, L., Mandl, M., Kärrman, A., Nowak, P., Wu, X., Dyck, A., Vaidyanathan, K., R, R., Mallet, J., Rudominer, M., Johnston, E., Mittal, S., Udathu, A., Christensen, J., Verma, V., Irving, Z., Santucci, A., Elsayed, G., Davoodi, E., Georgiev, M., Tenney, I., Hua, N., Cideron, G., Leurent, E., Alnahlawi, M., Georgescu, I., Wei, N., Zheng, I., Scandinaro, D., Jiang, H., Snoek, J., Sundararajan, M., Wang, X., Ontiveros, Z., Karo, I., Cole, J., Rajashekhar, V., Tumeh, L., Ben-David, E., Jain, R., Uesato, J., Datta, R., Bunyan, O., Wu, S., Zhang, J., Stanczyk, P., Zhang, Y., Steiner, D., Naskar, S., Azzam, M., Johnson, M., Paszke, A., Chiu, C.-C., Elias, J.~S., Mohiuddin, A., Muhammad, F., Miao, J., Lee, A., Vieillard, N., Park, J., Zhang, J.,
  Stanway, J., Garmon, D., Karmarkar, A., Dong, Z., Lee, J., Kumar, A., Zhou, L., Evens, J., Isaac, W., Irving, G., Loper, E., Fink, M., Arkatkar, I., Chen, N., Shafran, I., Petrychenko, I., Chen, Z., Jia, J., Levskaya, A., Zhu, Z., Grabowski, P., Mao, Y., Magni, A., Yao, K., Snaider, J., Casagrande, N., Palmer, E., Suganthan, P., Castaño, A., Giannoumis, I., Kim, W., Rybiński, M., Sreevatsa, A., Prendki, J., Soergel, D., Goedeckemeyer, A., Gierke, W., Jafari, M., Gaba, M., Wiesner, J., Wright, D.~G., Wei, Y., Vashisht, H., Kulizhskaya, Y., Hoover, J., Le, M., Li, L., Iwuanyanwu, C., Liu, L., Ramirez, K., Khorlin, A., Cui, A., LIN, T., Wu, M., Aguilar, R., Pallo, K., Chakladar, A., Perng, G., Abellan, E.~A., Zhang, M., Dasgupta, I., Kushman, N., Penchev, I., Repina, A., Wu, X., van~der Weide, T., Ponnapalli, P., Kaplan, C., Simsa, J., Li, S., Dousse, O., Yang, F., Piper, J., Ie, N., Pasumarthi, R., Lintz, N., Vijayakumar, A., Andor, D., Valenzuela, P., Lui, M., Paduraru, C., Peng, D., Lee, K., Zhang, S.,
  Greene, S., Nguyen, D.~D., Kurylowicz, P., Hardin, C., Dixon, L., Janzer, L., Choo, K., Feng, Z., Zhang, B., Singhal, A., Du, D., McKinnon, D., Antropova, N., Bolukbasi, T., Keller, O., Reid, D., Finchelstein, D., Raad, M.~A., Crocker, R., Hawkins, P., Dadashi, R., Gaffney, C., Franko, K., Bulanova, A., Leblond, R., Chung, S., Askham, H., Cobo, L.~C., Xu, K., Fischer, F., Xu, J., Sorokin, C., Alberti, C., Lin, C.-C., Evans, C., Dimitriev, A., Forbes, H., Banarse, D., Tung, Z., Omernick, M., Bishop, C., Sterneck, R., Jain, R., Xia, J., Amid, E., Piccinno, F., Wang, X., Banzal, P., Mankowitz, D.~J., Polozov, A., Krakovna, V., Brown, S., Bateni, M., Duan, D., Firoiu, V., Thotakuri, M., Natan, T., Geist, M., tan Girgin, S., Li, H., Ye, J., Roval, O., Tojo, R., Kwong, M., Lee-Thorp, J., Yew, C., Sinopalnikov, D., Ramos, S., Mellor, J., Sharma, A., Wu, K., Miller, D., Sonnerat, N., Vnukov, D., Greig, R., Beattie, J., Caveness, E., Bai, L., Eisenschlos, J., Korchemniy, A., Tsai, T., Jasarevic, M., Kong, W., Dao,
  P., Zheng, Z., Liu, F., Yang, F., Zhu, R., Teh, T.~H., Sanmiya, J., Gladchenko, E., Trdin, N., Toyama, D., Rosen, E., Tavakkol, S., Xue, L., Elkind, C., Woodman, O., Carpenter, J., Papamakarios, G., Kemp, R., Kafle, S., Grunina, T., Sinha, R., Talbert, A., Wu, D., Owusu-Afriyie, D., Du, C., Thornton, C., Pont-Tuset, J., Narayana, P., Li, J., Fatehi, S., Wieting, J., Ajmeri, O., Uria, B., Ko, Y., Knight, L., Héliou, A., Niu, N., Gu, S., Pang, C., Li, Y., Levine, N., Stolovich, A., Santamaria-Fernandez, R., Goenka, S., Yustalim, W., Strudel, R., Elqursh, A., Deck, C., Lee, H., Li, Z., Levin, K., Hoffmann, R., Holtmann-Rice, D., Bachem, O., Arora, S., Koh, C., Yeganeh, S.~H., Põder, S., Tariq, M., Sun, Y., Ionita, L., Seyedhosseini, M., Tafti, P., Liu, Z., Gulati, A., Liu, J., Ye, X., Chrzaszcz, B., Wang, L., Sethi, N., Li, T., Brown, B., Singh, S., Fan, W., Parisi, A., Stanton, J., Koverkathu, V., Choquette-Choo, C.~A., Li, Y., Lu, T., Ittycheriah, A., Shroff, P., Varadarajan, M., Bahargam, S., Willoughby,
  R., Gaddy, D., Desjardins, G., Cornero, M., Robenek, B., Mittal, B., Albrecht, B., Shenoy, A., Moiseev, F., Jacobsson, H., Ghaffarkhah, A., Rivière, M., Walton, A., Crepy, C., Parrish, A., Zhou, Z., Farabet, C., Radebaugh, C., Srinivasan, P., van~der Salm, C., Fidjeland, A., Scellato, S., Latorre-Chimoto, E., Klimczak-Plucińska, H., Bridson, D., de~Cesare, D., Hudson, T., Mendolicchio, P., Walker, L., Morris, A., Mauger, M., Guseynov, A., Reid, A., Odoom, S., Loher, L., Cotruta, V., Yenugula, M., Grewe, D., Petrushkina, A., Duerig, T., Sanchez, A., Yadlowsky, S., Shen, A., Globerson, A., Webb, L., Dua, S., Li, D., Bhupatiraju, S., Hurt, D., Qureshi, H., Agarwal, A., Shani, T., Eyal, M., Khare, A., Belle, S.~R., Wang, L., Tekur, C., Kale, M.~S., Wei, J., Sang, R., Saeta, B., Liechty, T., Sun, Y., Zhao, Y., Lee, S., Nayak, P., Fritz, D., Vuyyuru, M.~R., Aslanides, J., Vyas, N., Wicke, M., Ma, X., Eltyshev, E., Martin, N., Cate, H., Manyika, J., Amiri, K., Kim, Y., Xiong, X., Kang, K., Luisier, F.,
  Tripuraneni, N., Madras, D., Guo, M., Waters, A., Wang, O., Ainslie, J., Baldridge, J., Zhang, H., Pruthi, G., Bauer, J., Yang, F., Mansour, R., Gelman, J., Xu, Y., Polovets, G., Liu, J., Cai, H., Chen, W., Sheng, X., Xue, E., Ozair, S., Angermueller, C., Li, X., Sinha, A., Wang, W., Wiesinger, J., Koukoumidis, E., Tian, Y., Iyer, A., Gurumurthy, M., Goldenson, M., Shah, P., Blake, M., Yu, H., Urbanowicz, A., Palomaki, J., Fernando, C., Durden, K., Mehta, H., Momchev, N., Rahimtoroghi, E., Georgaki, M., Raul, A., Ruder, S., Redshaw, M., Lee, J., Zhou, D., Jalan, K., Li, D., Hechtman, B., Schuh, P., Nasr, M., Milan, K., Mikulik, V., Franco, J., Green, T., Nguyen, N., Kelley, J., Mahendru, A., Hu, A., Howland, J., Vargas, B., Hui, J., Bansal, K., Rao, V., Ghiya, R., Wang, E., Ye, K., Sarr, J.~M., Preston, M.~M., Elish, M., Li, S., Kaku, A., Gupta, J., Pasupat, I., Juan, D.-C., Someswar, M., M., T., Chen, X., Amini, A., Fabrikant, A., Chu, E., Dong, X., Muthal, A., Buthpitiya, S., Jauhari, S., Hua, N.,
  Khandelwal, U., Hitron, A., Ren, J., Rinaldi, L., Drath, S., Dabush, A., Jiang, N.-J., Godhia, H., Sachs, U., Chen, A., Fan, Y., Taitelbaum, H., Noga, H., Dai, Z., Wang, J., Liang, C., Hamer, J., Ferng, C.-S., Elkind, C., Atias, A., Lee, P., Listík, V., Carlen, M., van~de Kerkhof, J., Pikus, M., Zaher, K., Müller, P., Zykova, S., Stefanec, R., Gatsko, V., Hirnschall, C., Sethi, A., Xu, X.~F., Ahuja, C., Tsai, B., Stefanoiu, A., Feng, B., Dhandhania, K., Katyal, M., Gupta, A., Parulekar, A., Pitta, D., Zhao, J., Bhatia, V., Bhavnani, Y., Alhadlaq, O., Li, X., Danenberg, P., Tu, D., Pine, A., Filippova, V., Ghosh, A., Limonchik, B., Urala, B., Lanka, C.~K., Clive, D., Sun, Y., Li, E., Wu, H., Hongtongsak, K., Li, I., Thakkar, K., Omarov, K., Majmundar, K., Alverson, M., Kucharski, M., Patel, M., Jain, M., Zabelin, M., Pelagatti, P., Kohli, R., Kumar, S., Kim, J., Sankar, S., Shah, V., Ramachandruni, L., Zeng, X., Bariach, B., Weidinger, L., Vu, T., Andreev, A., He, A., Hui, K., Kashem, S., Subramanya, A.,
  Hsiao, S., Hassabis, D., Kavukcuoglu, K., Sadovsky, A., Le, Q., Strohman, T., Wu, Y., Petrov, S., Dean, J., and Vinyals, O.
\newblock Gemini: A family of highly capable multimodal models.
\newblock \emph{arXiv preprint arXiv: 2312.11805}, 2023.

\bibitem[Team(2024)]{qwen2.5}
Team, Q.
\newblock Qwen2.5: A party of foundation models, September 2024.
\newblock URL \url{https://qwenlm.github.io/blog/qwen2.5/}.

\bibitem[Teknium(2023)]{OpenHermes25}
Teknium.
\newblock Openhermes 2.5: An open dataset of synthetic data for generalist llm assistants, 2023.
\newblock URL \url{https://huggingface.co/datasets/teknium/OpenHermes-2.5}.

\bibitem[Touvron et~al.(2023)Touvron, Lavril, Izacard, Martinet, Lachaux, Lacroix, Rozière, Goyal, Hambro, Azhar, Rodriguez, Joulin, Grave, and Lample]{touvron2023llama}
Touvron, H., Lavril, T., Izacard, G., Martinet, X., Lachaux, M.-A., Lacroix, T., Rozière, B., Goyal, N., Hambro, E., Azhar, F., Rodriguez, A., Joulin, A., Grave, E., and Lample, G.
\newblock Llama: Open and efficient foundation language models.
\newblock \emph{arXiv preprint arXiv: 2302.13971}, 2023.

\bibitem[Vitter(1985)]{10.1145/3147.3165}
Vitter, J.~S.
\newblock Random sampling with a reservoir.
\newblock \emph{ACM Trans. Math. Softw.}, 11\penalty0 (1):\penalty0 37–57, March 1985.
\newblock ISSN 0098-3500.
\newblock \doi{10.1145/3147.3165}.
\newblock URL \url{https://doi.org/10.1145/3147.3165}.

\bibitem[Wang et~al.(2022{\natexlab{a}})Wang, Kordi, Mishra, Liu, Smith, Khashabi, and Hajishirzi]{wang2022selfinstruct}
Wang, Y., Kordi, Y., Mishra, S., Liu, A., Smith, N.~A., Khashabi, D., and Hajishirzi, H.
\newblock Self-instruct: Aligning language models with self-generated instructions.
\newblock \emph{arXiv preprint arXiv: 2212.10560}, 2022{\natexlab{a}}.

\bibitem[Wang et~al.(2022{\natexlab{b}})Wang, Mishra, Alipoormolabashi, Kordi, Mirzaei, Arunkumar, Ashok, Dhanasekaran, Naik, Stap, et~al.]{supernaturalinstructions}
Wang, Y., Mishra, S., Alipoormolabashi, P., Kordi, Y., Mirzaei, A., Arunkumar, A., Ashok, A., Dhanasekaran, A.~S., Naik, A., Stap, D., et~al.
\newblock Super-naturalinstructions:generalization via declarative instructions on 1600+ tasks.
\newblock In \emph{EMNLP}, 2022{\natexlab{b}}.

\bibitem[Wei et~al.(2021)Wei, Bosma, Zhao, Guu, Yu, Lester, Du, Dai, and Le]{wei2021finetuned}
Wei, J., Bosma, M., Zhao, V., Guu, K., Yu, A.~W., Lester, B., Du, N., Dai, A.~M., and Le, Q.~V.
\newblock Finetuned language models are zero-shot learners.
\newblock \emph{International Conference on Learning Representations}, 2021.

\bibitem[Xie et~al.(2024)Xie, Goyal, Zheng, Kan, Lillicrap, Kawaguchi, and Shieh]{xie2024monte}
Xie, Y., Goyal, A., Zheng, W., Kan, M.-Y., Lillicrap, T.~P., Kawaguchi, K., and Shieh, M.
\newblock Monte carlo tree search boosts reasoning via iterative preference learning.
\newblock \emph{arXiv preprint arXiv: 2405.00451}, 2024.

\bibitem[Yang et~al.(2024{\natexlab{a}})Yang, Yang, Hui, Zheng, Yu, Zhou, Li, Li, Liu, Huang, Dong, Wei, Lin, Tang, Wang, Yang, Tu, Zhang, Ma, Yang, Xu, Zhou, Bai, He, Lin, Dang, Lu, Chen, Yang, Li, Xue, Ni, Zhang, Wang, Peng, Men, Gao, Lin, Wang, Bai, Tan, Zhu, Li, Liu, Ge, Deng, Zhou, Ren, Zhang, Wei, Ren, Liu, Fan, Yao, Zhang, Wan, Chu, Liu, Cui, Zhang, Guo, and Fan]{yang2024qwen2}
Yang, A., Yang, B., Hui, B., Zheng, B., Yu, B., Zhou, C., Li, C., Li, C., Liu, D., Huang, F., Dong, G., Wei, H., Lin, H., Tang, J., Wang, J., Yang, J., Tu, J., Zhang, J., Ma, J., Yang, J., Xu, J., Zhou, J., Bai, J., He, J., Lin, J., Dang, K., Lu, K., Chen, K., Yang, K., Li, M., Xue, M., Ni, N., Zhang, P., Wang, P., Peng, R., Men, R., Gao, R., Lin, R., Wang, S., Bai, S., Tan, S., Zhu, T., Li, T., Liu, T., Ge, W., Deng, X., Zhou, X., Ren, X., Zhang, X., Wei, X., Ren, X., Liu, X., Fan, Y., Yao, Y., Zhang, Y., Wan, Y., Chu, Y., Liu, Y., Cui, Z., Zhang, Z., Guo, Z., and Fan, Z.
\newblock Qwen2 technical report.
\newblock \emph{arXiv preprint arXiv: 2407.10671}, 2024{\natexlab{a}}.

\bibitem[Yang et~al.(2024{\natexlab{b}})Yang, Pang, Feng, Wang, Chen, Zhu, and Liu]{yang2024selfdistillation}
Yang, Z., Pang, T., Feng, H., Wang, H., Chen, W., Zhu, M., and Liu, Q.
\newblock Self-distillation bridges distribution gap in language model fine-tuning.
\newblock \emph{arXiv preprint arXiv: 2402.13669}, 2024{\natexlab{b}}.

\bibitem[Yu et~al.(2023)Yu, Jiang, Shi, Yu, Liu, Zhang, Kwok, Li, Weller, and Liu]{yu2023metamath}
Yu, L., Jiang, W., Shi, H., Yu, J., Liu, Z., Zhang, Y., Kwok, J.~T., Li, Z., Weller, A., and Liu, W.
\newblock Metamath: Bootstrap your own mathematical questions for large language models.
\newblock \emph{International Conference on Learning Representations}, 2023.
\newblock \doi{10.48550/arXiv.2309.12284}.

\bibitem[Yuan et~al.(2024)Yuan, Pang, Cho, Li, Sukhbaatar, Xu, and Weston]{yuan2024selfrewarding}
Yuan, W., Pang, R.~Y., Cho, K., Li, X., Sukhbaatar, S., Xu, J., and Weston, J.
\newblock Self-rewarding language models.
\newblock \emph{arXiv preprint arXiv: 2401.10020}, 2024.

\bibitem[Zellers et~al.(2019)Zellers, Holtzman, Bisk, Farhadi, and Choi]{zellers2019hellaswag}
Zellers, R., Holtzman, A., Bisk, Y., Farhadi, A., and Choi, Y.
\newblock Hellaswag: Can a machine really finish your sentence?
\newblock \emph{Annual Meeting of the Association for Computational Linguistics}, 2019.
\newblock \doi{10.18653/v1/P19-1472}.

\bibitem[Zhang et~al.(2024{\natexlab{a}})Zhang, Qu, Liu, Zhang, Lin, Yu, Pan, Cheng, Liu, Lin, Yuan, Zheng, Pang, Du, Liang, Ma, Li, Ma, Lin, Benetos, Yang, Zhou, Ma, Liu, Niu, Wang, Que, Liu, Liu, Guo, Gao, Zhou, Zhang, Zhou, Wang, Bai, Zhang, Zhang, Wang, Yang, Zhao, Zhang, Ouyang, Huang, and Chen]{zhang2024mapneo}
Zhang, G., Qu, S., Liu, J., Zhang, C., Lin, C., Yu, C.~L., Pan, D., Cheng, E., Liu, J., Lin, Q., Yuan, R., Zheng, T., Pang, W., Du, X., Liang, Y., Ma, Y., Li, Y., Ma, Z., Lin, B., Benetos, E., Yang, H., Zhou, J., Ma, K., Liu, M., Niu, M., Wang, N., Que, Q., Liu, R., Liu, S., Guo, S., Gao, S., Zhou, W., Zhang, X., Zhou, Y., Wang, Y., Bai, Y., Zhang, Y., Zhang, Y., Wang, Z., Yang, Z., Zhao, Z., Zhang, J., Ouyang, W., Huang, W., and Chen, W.
\newblock Map-neo: Highly capable and transparent bilingual large language model series.
\newblock \emph{arXiv preprint arXiv: 2405.19327}, 2024{\natexlab{a}}.

\bibitem[Zhang et~al.(2024{\natexlab{b}})Zhang, Juan, Rashtchian, Ferng, Jiang, and Chen]{zhang2024sled}
Zhang, J., Juan, D.-C., Rashtchian, C., Ferng, C.-S., Jiang, H., and Chen, Y.
\newblock Sled: Self logits evolution decoding for improving factuality in large language models.
\newblock \emph{arXiv preprint arXiv: 2411.02433}, 2024{\natexlab{b}}.

\bibitem[Zheng et~al.(2024)Zheng, Guo, Qu, Guo, Du, Jia, Lin, Huang, Fu, and Zhang]{zheng2024kun}
Zheng, T., Guo, S., Qu, X., Guo, J., Du, X., Jia, Q., Lin, C., Huang, W., Fu, J., and Zhang, G.
\newblock Kun: Answer polishment for chinese self-alignment with instruction back-translation.
\newblock \emph{arXiv preprint arXiv: 2401.06477}, 2024.

\bibitem[Zhou et~al.(2023{\natexlab{a}})Zhou, Liu, Xu, Iyer, Sun, Mao, Ma, Efrat, Yu, Yu, Zhang, Ghosh, Lewis, Zettlemoyer, and Levy]{zhou2023lima}
Zhou, C., Liu, P., Xu, P., Iyer, S., Sun, J., Mao, Y., Ma, X., Efrat, A., Yu, P., Yu, L., Zhang, S., Ghosh, G., Lewis, M., Zettlemoyer, L., and Levy, O.
\newblock Lima: Less is more for alignment.
\newblock \emph{arXiv preprint arXiv: 2305.11206}, 2023{\natexlab{a}}.

\bibitem[Zhou et~al.(2023{\natexlab{b}})Zhou, Lu, Mishra, Brahma, Basu, Luan, Zhou, and Hou]{zhou2023instructionfollowing}
Zhou, J., Lu, T., Mishra, S., Brahma, S., Basu, S., Luan, Y., Zhou, D., and Hou, L.
\newblock Instruction-following evaluation for large language models.
\newblock \emph{arXiv preprint arXiv: 2311.07911}, 2023{\natexlab{b}}.

\end{thebibliography}
